%% file: 0_main.tex
\documentclass[10pt,twocolumn,letterpaper]{article}

\usepackage{cvpr}              % To produce the CAMERA-READY version
\usepackage{graphicx}
\usepackage{amsmath}
\usepackage{amssymb}
\usepackage{booktabs}
\usepackage[accsupp]{axessibility}

\usepackage[pagebackref,breaklinks,colorlinks]{hyperref}

\usepackage[capitalize]{cleveref}
\crefname{section}{Sec.}{Secs.}
\Crefname{section}{Section}{Sections}
\Crefname{table}{Table}{Tables}
\crefname{table}{Tab.}{Tabs.}

\usepackage[labelsep=period]{caption}
\usepackage{xcolor}
\usepackage{enumitem}

\renewcommand{\paragraph}[1]{\vspace{0.2em}\noindent \textbf{#1 \hspace{0.2em}}}

\definecolor{MyDarkRed}{rgb}{0.66, 0.16, 0.16}
\definecolor{MyDarkBlue}{rgb}{0.16, 0.16, 0.66}
\usepackage{lipsum} % Just for generating dummy text, can be removed

\def\cvprPaperID{1281} % *** Enter the CVPR Paper ID here
\def\confName{CVPR}
\def\confYear{2023}

\begin{document}

\title{SCoDA: Domain Adaptive Shape Completion for Real Scans}

\author{
Yushuang Wu$^{1,2,3}$\quad 
Zizheng Yan$^{1,2,3}$\quad 
Ce Chen$^{1,2}$\quad
Lai Wei$^{4}$\quad
Xiao Li$^{5}$\\ 
Guanbin Li$^{6,7}$\quad
Yihao Li$^{1,2}$\quad 
Shuguang Cui$^{2,1}$\quad
Xiaoguang Han$^{2,1\#}$\\
$^1$FNii, CUHKSZ\quad $^2$SSE, CUHKSZ\quad $^3$SRIBD\quad $^4$SDS, CUHKSZ\quad $^5$Microsoft Research Asia \\$^6$Sun Yat-sen University \quad $^7$Research Institute, Sun Yat-sen University, Shenzhen\\
{\tt\small \{yushuangwu, zizhengyan, cechen, laiwei1\}@link.cuhk.edu.cn\quad luoheliyihao@mail.ustc.edu.cn}  \\ {\tt\small \{shuguangcui, hanxiaoguang\}@cuhk.edu.cn\quad liguanbin@mail.sysu.edu.cn\quad li.xiao@microsoft.com}
}

% \maketitle
\twocolumn[{%
\renewcommand\twocolumn[1][]{#1}%
\maketitle
\begin{center}
    \centering
    \captionsetup{type=figure}
    \vspace{-6mm}
    \includegraphics[width=0.85\linewidth]{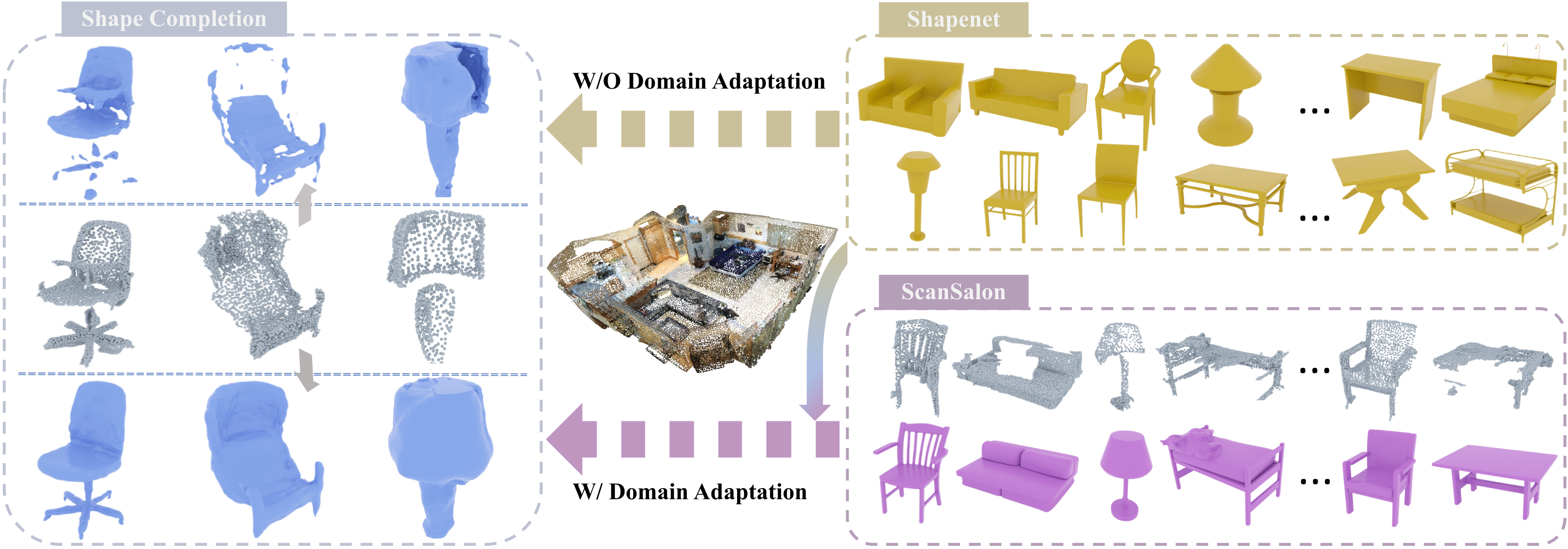}
    % \vspace{-2mm}
    \caption{The proposed task \textbf{SCoDA} aims to transfer the knowledge in the synthetic domain to the reconstruction of noisy and incomplete real scans. A dataset, \textbf{ScanSalon}, with paired real scans and 3D models is contributed. Project page: \href{https://yushuang-wu.github.io/SCoDA/}{yushuang-wu.github.io/SCoDA}.} \label{fig:teaser}
\end{center}%
}]

\input{1_abstract}

\input{2_intro}

\input{3_relatedwork}

\input{4_method}

\input{5_dataset}

\input{6_experiments}
\input{7_conclusions}

\clearpage
%%%%%%%%% REFERENCES
{\small
\bibliographystyle{ieee_fullname}
\bibliography{ref}
}

\end{document}

%%%%%%%%% TITLE
\title{Supplementary Material for \\ SCoDA: Domain Adaptive Shape Completion for Real Scans}

% \author{ABC}
\maketitle

% {
%     \hypersetup{linkcolor=black}
%     \tableofcontents
% }

% \vspace{2cm}
\section{More ScanSalon Details}
% \paragraph{Video Visualization} For more examples in the proposed ScanSalon dataset, we poduce videos that shows (i) the diversity and (ii) the high quality (well recover the details and greatly match the paired real scans) of 3D models in ScanSalon. Please refer to our project page \href{yushuang-wu.github.io/SCoDA}{https://yushuang-wu.github.io/SCoDA/}. 
We provide a comparison between the number of synthetic and real scans in Tab.~\ref{tab:dataset}. The synthetics scans are from the ShapeNet dataset~\cite{chang2015shapenet}. Besides the class ``Bed'', there are more samples of synthetic scans than real scans. 

\begin{table}[tb]\centering
    \caption{Statistics of the proposed dataset ScanSalon.}
    \label{tab:dataset}
    \resizebox{0.475\textwidth}{!}{
    \begin{tabular}{l|cccccc|c}
        \toprule
                   & Chair & Desk & Sofa & Bed & Lamp & Car & Total\\
        \midrule
        Synthetic Scans   & 6,579 & 8,071 & 3,091 & 233 & 2,318 & 3,514 & 23,806  \\
        Real Scans    & 4,651 & 1,630 & 428 & 365 & 133 & 437 & 7,644  \\
        Paired Models   & 497 & 161 & 43 & 36 & 20 & 43 & 800   \\
        \bottomrule
    \end{tabular}
    }
    \vspace{-5mm}
\end{table}

% \begin{itemize}
%     \item You can choose to use two-column or single-column layout.
% {\small
% \begin{verbatim}
% \documentclass[10pt,letterpaper]{article} % Single column
% \documentclass[10pt,letterpaper,twocolum]{article} % Two column
% \end{verbatim}
%         }

%     \item Please update the ConfName, ConfYear, and PaperID accordingly.
% {\small
% \begin{verbatim}
% \def\confName{CVPR\xspace} % *** Enter the Conference Name
% \def\confYear{2022\xspace} % *** Enter the conference Year
% \def\PaperID{8888} % *** Enter the CVPR Paper ID here
% \end{verbatim}
%         }    
%     \item Switch between ``review'', ``camera ready'', and ``arxiv'' versions.
% {\small
% \begin{verbatim}
% \usepackage[review]{cvpr}      % To produce the REVIEW version
% %\usepackage{cvpr}              % To produce the CAMERA-READY version
% %\usepackage[pagenumbers]{cvpr} % To force page numbers, e.g. for an arXiv version
% \end{verbatim}
%         }
% \end{itemize}

% \clearpage

\section{More Implementation Details}
\paragraph{Synthetic Scan Generation}
To get point clouds from the ShapeNet models, we adopt a popular simulation toolbox, BlenSor~\cite{gschwandtner2011blensor}, which supports scanning simulation with different sensors (\textit{e.g.} Velodyne, Kinect, and Time of Flight camera) and parameters. To simulate the sparsity in real scans from the ScanNet dataset, we conduct a random down-sampling with ratio 13\%, which is computed according to the average point number of scans in ScanNet and simulated point clouds from ShapeNet. We also add Gaussian noise with a max scale 0.01 to simulate the noise in scanning. The incompleteness is also introduced in the simulated scanning by self-occlusion. Besides, we adopt the unsupervised clustering-based way introduced in the main paper to partition the point clouds and randomly drop 1$\sim$4 clusters in the training process. Some simulation results (before dropping some clusters in training) are presented in the second row of Fig.~\ref{fig:pcd_comp}. 
As shown in Fig.~\ref{fig:pcd_comp}, in spite of careful simulation, the generated synthetic scans are still different from real scans (the first row of Fig.~\ref{fig:pcd_comp}) in (i) sparsity: dependent of the object materials and object-scanner distance, the sparsity of real scans has a larger variance; (ii) noise: both the scanning intrinsic error the segmentation from background can introduce complex noise; and (iii) incompleteness: the incompleteness of real scans results from the occlusion and surroundings, which is more complex. Thus, there still exists a domain gap between two kinds of point clouds, which can hardly be handled by simulation. % All experiments are implemented on 4$\times$RTX3090, using PyTorch-1.13.0. 

\input{figures/Supp_pcd_comp.tex}

\paragraph{Implementation of Baselines}
The implemented baselines include: (i) IF-Net: it consists of a 6-layer 3DCNN with a 4-layer MLP, and the network structure is used in all baselines; (ii) SelfSup: a mean-square-error (MSE) loss is used to minimize the Euclidean distance between the normalized top-layer feature vectors generated from the two views, which are created in the same way as in our method; (iii) PtComp: two 3DCNN-based UNets (5 layers in the encoder/decoder) are used for the encoding-decoding of voxelized point clouds from the real and synthetic domains for point completion, respectively. An adversarial loss is used to encourage the domain invariance of the codes output by the encoders following \cite{chen2019unpaired}, and for supervised samples, an additional MSE loss is used to minimize the distance between the generations and the ground truths. The completion results share the same resolution with the input, so are then fed into a standard IF-Net for reconstruction; (iv) Adversarial: based on a standard IF-Net-based reconstruction framework, an adversarial loss is adopted to minimize the domain discriminativeness of the top-layer feature vectors generated by the 3DCNN, where a three-layer MLP (the latent dimensions are 256 and 512) is used as the discriminator, and the adversarial loss is rescaled by a ratio of 0.01 to be integrated. 

% \begin{figure*}[h] \centering
%     \includegraphics[width=\textwidth,height=0.4\textwidth]{example-image}
%     \caption{Image 1.} \label{fig:}
% \end{figure*}

% \begin{table}[h]\centering
%     \caption{A simple table with a header row.}
%     \label{tab:table1}
%     \resizebox{0.48\textwidth}{!}{
%     \large
%     \begin{tabular}{*{10}{c}}
%         \toprule
%        Data & Size &  2-Exp & 3-Exp &  4-Exp & 5-Exp &  6-Exp &  7-Exp \\
%         \midrule
%         A & $1280\times 720$ & 1 & 2 & 3 & 4 & 5 & 4 \\
%         B & $1280\times 720$ & 1 & 2 & 3 & 4 & 5 & 4 \\
%         Ours & $4096\times 2168$ & 2 & 3 & 4 & 6 & 5 & 4 \\
%         \bottomrule
%     \end{tabular}
%     }
% \end{table}

\section{More Qualitative Results}
\input{figures/Supp_qualitative_results.tex}
We present the qualitative comparison between our method and all baselines on the 3\% label setting, and here we give the visualized results on the 5\% label setting in Fig.~\ref{fig:qualitative_results}. It can be seen that the proposed method is superior in reconstruction also on the 5\% setting.

\section{More Ablative Results}
\paragraph{Single Module of CDFF and VCST}
In the main paper, we list the results of using only CDFF or VCST module in 3 classes for the space limit. We provide the results on all 6 classes on the 3\% label setting in Tab.~\ref{tab:ablation_6class}. 
\vspace{-3mm}
% \subsection{Variants of CDFF and VCST}
% For qualitative comparison of the ablative experiments, we provide the visualization of 3 chair samples generated by different variants of the CDFF and VCST module in Fig.~\ref{fig:abltab4} and \ref{fig:abltab5}. 
\paragraph{CDFF variants} For qualitative comparison of the ablative experiments, we provide the visualization of 3 chair samples generated by different variants of the CDFF module in Fig.~\ref{fig:abltab4}. Among all variants of CDFF, the completion results are relatively worse without feature fusion (columns (4) and (5)). When only using knowledge from the source domain (column (4)), the reconstruction results tend to rely more on the input scan but create less completion. A potential reason is the training data in the source domain are cleaner and with supervision, which introduces the domain bias. Differently, when  only using knowledge from the target domain (column (5)), the reconstructions tend to produce more completion but have a worse global shape (\textit{e.g.} the second row of column (5)). 

\vspace{-3mm}
\paragraph{VCST variants} The reconstruction results by different variants of the VCST module are visualized in Fig.~\ref{fig:abltab5}. We can observe that (i) when using random down-sampling only, the network tends to give worse completion for the missing object components, which proves the necessity of employing surface-aware augmentation. (ii) when integrating the volume-aware augmentation (columns (6), (8), and (9)), the reconstructions tend to over-complete the shapes (the first two rows of columns (6), (8), and (9)), of which a potential reason is that our clustering-based surface-aware augmentation implies some object-specific information, which creates incompleteness that is more close to the ones in real scans. These results further validate the superiority of surface-aware augmentation. 

\clearpage

\begin{table*}[htbp] \centering
    \newcommand{\Frst}[1]{\textcolor{red}{\textbf{#1}}}
    \newcommand{\Scnd}[1]{\textcolor{blue}{\textbf{#1}}}
    \caption{Experiment results on the 3\% setting of the SCoDA task (the complete results on 6 classes of Tab.~3 in the main paper). The units of CD and mIoU value are $1\times 10^{-3}$ and \%, respectively. \Frst{Red} text indicates the best result.}
    \label{tab:ablation_6class}
    \input{tables/5_supp_ablation1.tex}
    % \vspace{-5mm}
\end{table*}
\input{figures/Supp_ablation_fig.tex}

\clearpage

\section{Failure Case Analysis}
\input{figures/Supp_failure_case.tex}
From abundant reconstruction results generated by our method, we conclude 3 kinds of cases that easily lead to failure, which are also shared by other baselines. The 3 kinds of cases are: (i) much incompleteness makes completion harder, which is also the most common reason that leads to poor reconstruction, \textit{e.g.} the first two rows in Fig.~\ref{fig:failure_case}; (ii) the distribution bias makes completion fail in some parts, \textit{e.g.} the recovery failure of bed and chair legs in the third row of Fig.~\ref{fig:failure_case}; (iii) the strong noise in the input scans misleads the reconstruction, \textit{e.g.} the poor reconstruction quality of lamps in the last row of Fig.~\ref{fig:failure_case}.

\newpage

\pagebreak
{\small
\bibliographystyle{ieee_fullname}
\bibliography{ref}
}

%% file: 1_abstract.tex
%%%%%%%%% ABSTRACT
\begin{abstract}
    3D shape completion from point clouds is a challenging task, especially from scans of real-world objects. Considering the paucity of 3D shape ground truths for real scans, existing works mainly focus on benchmarking this task on synthetic data, e.g. 3D computer-aided design models. %, from which point clouds can be achieved via simulation and the 3D shapes are leveraged as the ground truth for reconstruction learning. 
    However, the domain gap between synthetic and real data limits the generalizability of these methods. Thus, we propose a new task, SCoDA, for the domain adaptation of real scan shape completion from synthetic data. A new dataset, ScanSalon, is contributed with a bunch of elaborate 3D models created by skillful artists according to scans.  
    % However, the domain gap between synthetic and real data limits the generalizability of these methods. We propose a semi-supervised domain adaptive reconstruction framework to sufficiently utilize the rich knowledge in the reconstruction of synthetic data and transfer them to the real domain with only a few labels. 
    To address this new task, we propose a novel cross-domain feature fusion method for knowledge transfer and a novel volume-consistent self-training framework for robust learning from real data. Extensive experiments prove our method is effective to bring an improvement of 6\%$\sim$7\% mIoU.
    % Besides, a new benchmark is constructed on 6 categories, where a bunch of elaborate 3D models is provided for evaluation or semi-supervised learning, 
\end{abstract}

%%% for open review %%%
% 3D shape completion from point clouds is a challenging task, especially from scans of real-world objects. Considering the paucity of 3D shape ground truths for real scans, existing works mainly focus on benchmarking this task on synthetic data, e.g. 3D computer-aided design models. However, the domain gap between synthetic and real data limits the generalizability of these methods. Thus, we propose a new task, SCoDA, for the domain adaptation of real scan shape completion from synthetic data. A new dataset, ScanSalon, is contributed with a bunch of elaborate 3D models created by skillful artists according to scans. To address this new task, we propose a novel cross-domain feature fusion method for knowledge transfer and a novel volume-consistent self-training framework for robust learning from real data. Extensive experiments prove our method is effective to bring an improvement of 6%~7% mIoU.

%% file: 2_intro.tex
\section{Introduction}
\label{sec:intro}
Shape completion and reconstruction from scans is a practical 3D digitization task that is of great significance in applications of virtual and augmented reality. It takes a scanned point cloud as input and aims to recover the 3D shape of the target object. The completion of real scans is challenging for the poor quality of point clouds and the deficiency of 3D shape ground truths. Existing methods exploit synthetic data, \textit{e.g.} 3D computer-aided design (CAD) models to alleviate the demand for real object shapes. For example, authors of \cite{erler2020points2surf, chibane2020implicit, hanocka2020point2mesh} simulate the scanning process to obtain point clouds from CAD models with paired ground truth shapes to train learning-based reconstruction models.
%, and another work \cite{avetisyan2019scan2cad} provides approximated 3D shapes for real scans by retrieving and deforming similar CAD models. 
However, there still exist distinctions between the simulated and real scan because the latter is scanned from a real object with complex scanning noise and occlusion, which limits the generalization quality. % in sparsity, noise, and incompleteness that are highly dependent on the used scanner, object occlusion, and scene layout.

Considering the underexploration in this field, we propose a new task, \textbf{SCoDA}, \textbf{\underline{D}}omain \textbf{\underline{A}}daptive \textbf{\underline{S}}hape \textbf{\underline{Co}}mpletion, that aims to transfer the knowledge from the synthetic domain with rich clean shapes (source domain) into the shape completion of real scans (target domain), as illustrated in Fig.~\ref{fig:teaser}. % Such a knowledge transfer needs to handle two main challenges in this task: (i) input distinction (input domain gap), (ii) no supervision, learn from data and/or few labels. 
To this end, we, for the first time, build an object-centric dataset, \textbf{ScanSalon}, which consists of real \underline{\textbf{Scan}}s with \underline{\textbf{S}}h\underline{\textbf{a}}pe manua\underline{\textbf{l}} annotati\underline{\textbf{on}}s. Our ScanSalon contains a bunch of 3D models that are paired with real scans of objects in six categories: chair, desk/table, sofa, bed, lamp, and car. The 3D models are manually created of high quality by skillful artists for around 10\% of real scans, which can serve as the evaluation ground truths of shape completion or as the few labels for semi-supervised domain adaptation. See Fig.~\ref{fig:teaser} for some examples, and details of \textbf{ScanSalon} are exposed in Sec.~\ref{sec:dataset}.  

The main challenge of the proposed SCoDA task lies in the domain gap between synthetic and real point clouds. Due to the intrinsic complexity of the scanning process, \textit{e.g.}, the scanner parameters and object materials, it is difficult to simulate the scans in terms of sparsity, noisy extent, \etc. More importantly, real scans are usually incomplete resulting from the scene layout and object occlusion during scanning, which can hardly be simulated. 
Thus, we propose a novel domain adaptive shape completion approach to transfer the rich knowledge from the synthetic domain to the real one. At first, the reconstruction module in our approach is based on an implicit function for continuous shape completion~\cite{mescheder2019occupancy, park2019deepsdf, chibane2020implicit}. 
For an effective transfer, we observe that although the local patterns of real scans (\eg, noise,  sparsity, and incompleteness) are distinct from the simulated synthetic ones, the global topology or structure in a same category is usually similar between the synthetic and real data, for example, a chair from either the synthetic or real domain usually consists of a seat, a backrest, legs, etc. (see Fig.~\ref{fig:teaser}). 
% Accordingly, we propose a novel domain adaptive shape completion framework based on global-local feature fusion, which leverages the global representations learned in the synthetic domain, adaptively combined with the local representations learned in the real domain, to help the implicit function recover both fine details and global structures. Besides, a self-supervised implicit learning based on consistency training is also developed for sufficient learning from target data. Specifically, a partial scan is sampled from the original scan to create two views, from which consistent implicit predictions are encouraged to enable the model more robust to the complex sparsity, noise, and incompleteness in real scans. 
% Accordingly, we propose a novel domain adaptive shape completion approach. First, the reconstruction module in our approach is based on an implicit function for continuous shape completion~\cite{mescheder2019occupancy, park2019deepsdf, chibane2020implicit}.
In other words, the global topology is more likely to be domain invariant, while the local patterns are more likely to be domain specific.
Accordingly, we propose a cross-domain feature fusion (CDFF) module to combine the global features and local features learned in the synthetic and real domain, respectively, which helps recover both fine details and global structures in the implicit reconstruction stage. Moreover, a novel volume-consistent self-training (VCST) framework is developed to encourage self-supervised learning from the target data. % Specifically, an unsupervised clustering is performed first to partition the object scan into several components.%, and two views are created with different extent of completeness by dropping different number of components. Then the model is forced to give consistent implicit predictions from these two views at each spatial volume, which aims to improve the model's robustness to the incompleteness of real scans. 
Specifically, we create two views of real scans by dropping different clusters of points to produce incompleteness of different extent, and the model is forced to make consistent implicit predictions at each spatial volume, which encourages the model's robustness to the incompleteness of real scans. 

To construct a benchmark on the proposed new task, we implement some existing solutions to related tasks as baselines, and develop extensive experiments for the baselines and our method on \textbf{ScanSalon}. These experiments also demonstrate the effectiveness of the proposed method.

\vspace{3pt}
In summary, our key contributions are four-fold:  
\begin{itemize}[itemsep=0pt,parsep=0pt,topsep=2bp]
    \item We propose a new task, \textbf{SCoDA}, namely domain adaptive shape completion for real scans; A dataset \textbf{ScanSalon} that contains 800 elaborate 3D models paired with real scans in 6 categories is contributed.
    \item A novel cross-domain feature fusion module is designed to combine the knowledge of global shapes and local patterns learned in the synthetic and real domain, respectively. Such a feature fusion manner may also inspire the works in the 2D domain adaption field. 
    \item A volume-consistent self-training framework is proposed to improve the robustness of shape completion to the complex incompleteness of real scans. 
    \item A benchmark with multiple methods evaluated is constructed for the task of \textbf{SCoDA} based on \textbf{ScanSalon}; Extensive experiments also demonstrate the superiority of the proposed method.
\end{itemize}

%% file: 3_relatedwork.tex
%-------------------------------------------------------------------------
\section{Related Work}
\label{sec:related_works}

\paragraph{Shape Completion} Shape completion or reconstruction from point clouds is a branch of 3D reconstruction task~\cite{choy20163d, tatarchenko2019single, xu2019disn, saito2019pifu, mescheder2019occupancy, xie2019pix2vox, lin2020sdf}. The recovered shape can be represented as dense points~\cite{lin2018learning, peng2021shape, li2021sp} (\textit{i.e.}, point cloud completion or consolidation), polygonal meshes~\cite{litany2018deformable, hanocka2020point2mesh, gao2020learning, wei2021deep}, manifold atlases~\cite{groueix2018papier, williams2019deep, deprelle2019learning, badki2020meshlet, gadelha2021deep}, voxel grids~\cite{choy20163d, hane2017hierarchical, wu2017marrnet}, or implicit representations~\cite{peng2020convolutional, sitzmann2020metasdf, chibane2020implicit, jiang2020local, erler2020points2surf, williams2022neural}. 
Among them, the task of point cloud completion is related to our work, which aims to recover a complete, dense, and clean point cloud for the 3D shape, from a partial, sparse, and noisy input scanned by sensors. The pioneering work, PCN~\cite{pcn18}, proposes a coarse-to-fine completion framework with an encoder-decoder architecture, which encodes the partial points into global features and recovers the fined results in the decoding stage. 
The following works explore further to better exploit local or multi-scale features for higher-quality completion \cite{pcn18, topnet19, grnet20, skip-att20, morphing20, softpoolnet20, pfnet20, cascaded20, wen2021pmp, yu2021pointr, vrnet21, snowflakenet21, zhang2022point, fbnet22}.    
% Most of above approaches exploit synthetic data to training networks in a fully-supervised manner, which usually acquire incomplete point clouds by sampling from the surface of CAD models. Considering their difficulty of generalizing into real-world scans, another line of works propose to learn from the real data in a unsupervised or weakly-supervised manner \cite{chen2019unpaired, wen2021cycle4completion, wu2020multimodal, wu20213d, zhang2021unsupervised, cai2022learning, gu2020weakly}. 
In addition, the implicit representation based approaches now attract increasing attention because of its superior property that enables continuous shape recovery for objects with arbitrary topologies~\cite{mescheder2019occupancy, park2019deepsdf, chen2019learning, chibane2020implicit}. 
Although numerous efforts have been contributed, the existing approaches are usually performed on point clouds scanned from synthetic data via simulation, and a straightforward application on real scans results in an unsatisfactory performance due to the domain gap between synthetic and real scans~\cite{chibane2020implicit, erler2020points2surf, sulzer2022Deep}. For the task of point completion, some works also adopt unsupervised or weakly-supervised learning techniques to learn from unlabeled data~\cite{chen2019unpaired, wen2021cycle4completion, wu2020multimodal, wu20213d, zhang2021unsupervised, cai2022learning, gu2020weakly}. Another line of works retrieve suitable models from 3D CAD repositories and deform them as the estimated shape ground truths of object scans~\cite{avetisyan2019scan2cad, uy2021joint}. Differently, our work first proposes the task of domain adaptive shape completion for real scans, which aims to sufficiently transfer the knowledge from the label-rich synthetic domain into real scan shape completion. 

\paragraph{2D Domain Adaptation} 
Domain adaptation has been a popular research topic in various 2D vision tasks, including image recognition~\cite{chen2020homm,damodaran2018deepjdot,saito2018mcd,cui2020gvb,chen2019transferability,gao2021gradient,long2017deep,berthelot2021adamatch}, semantic segmentation~\cite{melas2021pixmatch,chen2019crdoco,mei2020instance,hoyer2022daformer}, and object detection~\cite{khodabandeh2019robust,chen2020harmonizing,munir2021ssal,rezaeianaran2021seeking}. Existing works can broadly be categorized as (i) adversarial-based methods~\cite{saito2018mcd,cui2020gvb,chen2019transferability,chen2022reusing,chen2019crdoco,chen2020harmonizing} that train a discriminator to discriminate the source and the target domain, (ii) alignment-based methods~\cite{gao2021gradient,wei2021deep,chen2020homm,damodaran2018deepjdot,rezaeianaran2021seeking} that use pre-defined metrics to align the source and the target domain, and (iii) self-training-based methods~\cite{melas2021pixmatch,mei2020instance,hoyer2022daformer,munir2021ssal,berthelot2021adamatch} that exploit the pseudo labels of the target domain to train the model.
Recently, consistency-based self-training methods have become the most popular approach due to their impressive performance. One key factor in the success of consistency training is the augmentations applied to the inputs~\cite{xie2020unsupervised, sohn2020fixmatch, ghosh2020data}, \eg, stronger augmentations have better performance in domain adaptive image recognition~\cite{berthelot2021adamatch}. To this end, we design a novel consistency training framework specialized for the shape completion task. Furthermore, we propose to conduct knowledge transfer via a cross-domain feature fusion manner, which has not been explored by existing works, to our best knowledge.

\paragraph{3D Domain Adaptation} 
% Domain adaptation techniques act on transferring knowledge from a label-rich source domain into the target domain to alleviate the reliance on supervision, which is of great significance in practical applications. 
In the 3D vision field, existing works on domain adaptation are mainly committed to the task of point cloud understanding, including classification, part segmentation, and semantic segmentation. For object-centric point clouds, PointDAN~\cite{qin2019pointdan} constructs a domain adaptive classification benchmark by collecting point clouds from ShapeNet~\cite{chang2015shapenet}, ModelNet~\cite{wu20153d}, and ScanNet~\cite{dai2017scannet} datasets to compose 3 different domains. It also proposes an adversarial representation learning approach with feature alignment at both local and global levels. The following works improve the classification precision by adopting techniques of consistency training~\cite{li2021self}, self-training~\cite{cardace2021refrec, zou2021geometry}, multimodal learning~\cite{afham2022crosspoint}, or learning from self-supervised tasks~\cite{luo2021learnable, achituve2021self, shen2022domain, huang2022generation}. 
Another line of works are committed to scene-centric point cloud analysis, \textit{i.e.}, point cloud semantic segmentation \cite{wu2019squeezesegv2, jaritz2020xmuda, peng2021sparse, yi2021complete, jiang2021lidarnet, zhao2021epointda, bian2022unsupervised1, xiao2022transfer, rochan2022unsupervised, bian2022unsupervised2}. Considering the exhaustive annotation process for 3D point semantics, they propose to transfer the segmentation knowledge in synthetic scenes into real-world point clouds \cite{wu2019squeezesegv2, jiang2021lidarnet, zhao2021epointda, xiao2022transfer}. % Another work, Scan2CAD~\cite{avetisyan2019scan2cad}, provides estimated 3D shape models for real object scans by retrieving from a 3D CAD model database. 
% Besides, some works leverage domain adaptation techniques for point completion mainly based on adversarial training to learn domain-invariant features~\cite{chen2019unpaired, wu20213d, wen2021cycle4completion, zhang2021unsupervised, cai2022learning}. 
Besides, there are a few attempts contributed to domain adaptive single-view reconstruction (SVR) \cite{pinheiro2019domain, yin20213dstylenet, leung2021domain}, which is highly related to our work. Except for the difference in tasks (SVR v.s. shape completion), these works usually apply existing domain adaptation techniques to the target task, while we propose a novel domain adaptation method  based on cross-domain feature fusion, which may also inspire others working on domain adaption. 

%-------------------------------------------------------------------------

%% file: 4_method.tex
\begin{figure*}[tb] \centering
    \includegraphics[width=0.86\textwidth]{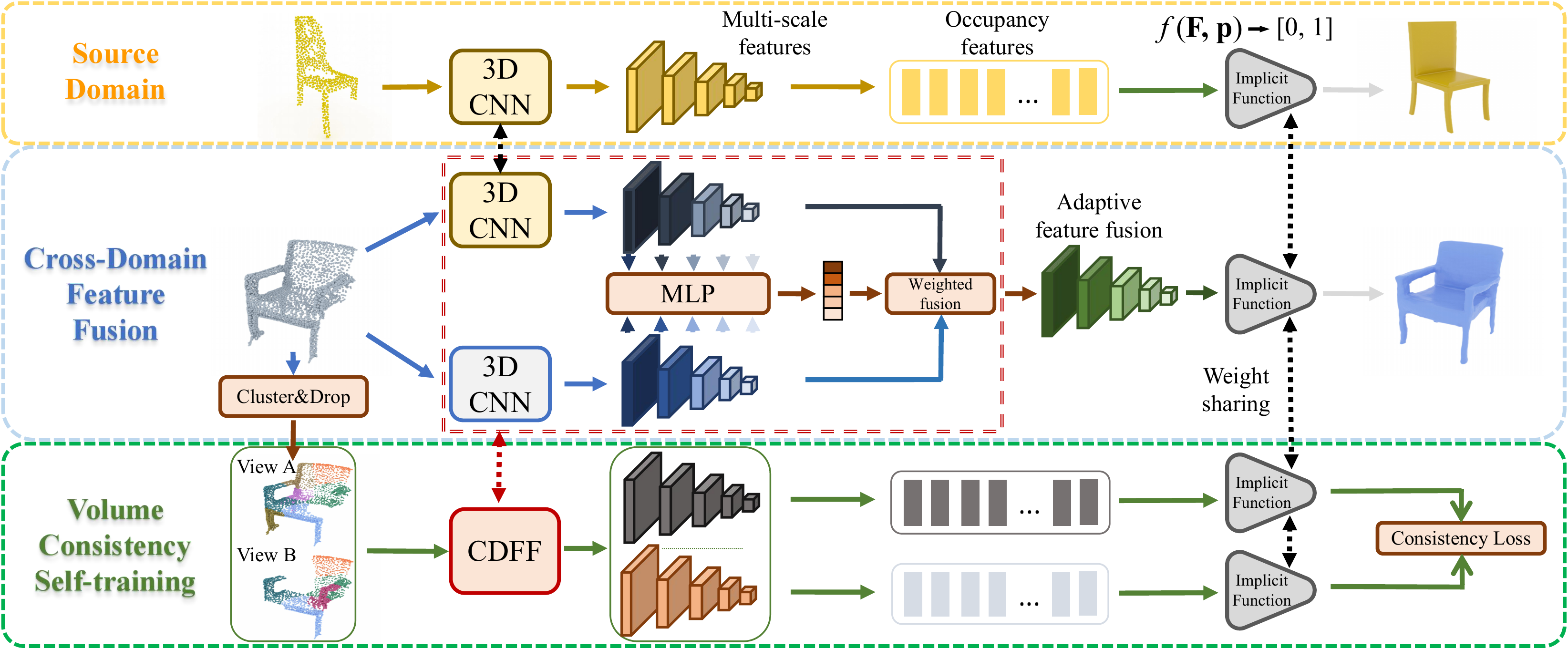}
    \caption{Overview of the proposed method. Two IF-Net encoders are used for the source and the target domain, respectively, and they share an implicit function decoder. The cross-domain feature fusion~(CDFF) works by adaptively combining the global-level and local-level knowledge learned from the source and target domain, respectively. The volume-consistency self-training (VCST) works by enforcing the prediction consistency between two different augmented views to learn the local details.} \label{fig:method}
    \vspace{-5mm}
\end{figure*}

\section{Method}%
\label{sec:method}
% The implicit function is now a popular technique for a continuous reconstruction~\cite{peng2020convolutional, sitzmann2020metasdf, chibane2020implicit, jiang2020local, erler2020points2surf, williams2022neural}. 
IF-Nets are a promising reconstruction method for their impressive performance~\cite{chibane2020implicit}, which is also successfully used in many applications~\cite{chibane2020implicit2, ouasfi2022few}. Constructed on IF-Nets, our approach first improves the representation learning by proposing a novel cross-domain feature fusion module, which aims to transfer the knowledge on the global-level object shape learned in the label-rich synthetic domain into the real domain. %, where local information are captured in a target-specific manner and combine with global features acquired by another extractor learned in the source domain. 
Secondly, to exhaust the domain-specific information in the real data, a novel volume-consistent self-training method is proposed for the specific shape completion task, which also leads to a robust implicit prediction of IF-Nets. An illustration is presented in Fig.~\ref{fig:method}.   

\subsection{Implicit Feature Networks}%
We first introduce IF-Nets~\cite{chibane2020implicit} as our reconstruction framework. An IF-Net is composed of a 3D convolution neural network encoder $g(\cdot)$ for multi-scale feature extraction and a multi-layer perceptron (MLP) $f(\cdot)$ for implicit shape decoding. 

% Given a point cloud sample $P$, it is first converted into a discrete voxel representation $\mathbf{X} \in \mathbb{R}^{N\times N\times N}$, where $N \in \mathbb{N}$ is the resolution of the input space. 
% $\mathbf{X}$ is then fed into the $L$-layer encoder $g(\cdot)$ to generate multi-scale features $\mathbf{F} = \text{concat}(\{\mathbf{F}_l\}), l\in \{1,2,\cdots, L\}$, where $\mathbf{F} \in \mathbb{R}^{d\times  N\times  N\times N}$ is obtained via concatenating features $\{\mathbf{F}_l \}$ in the channel dimension, formulated as:
% $$
% g(\mathbf{X}) = \mathbf{F}, \quad \mathbf{X} \in \mathbb{R}^{N\times N\times N}.
% $$
% Note that $\mathbf{F}$ has a 3D structure aligned with the input $\mathbf{X}$. 
% Given a point query $\mathbf{p} \in \mathbb{R}^3$, continuous features $\mathbf{F}(\mathbf{p})$ at this point can be extracted from $\mathbf{F}$ using trilinear interpolation . 
Given a point cloud sample $P$, it is first converted into a discrete voxel representation $\mathbf{X} \in \mathbb{R}^{N\times N\times N}$, where $N \in \mathbb{N}$ is the resolution of the input space. 
$\mathbf{X}$ is then fed into the $L$-layer encoder $g(\cdot)$ to generate
a set of multi-scale features $\{\mathbf{F}_1, ..., \mathbf{F}_L\}$, then they are upsampled to the same spatial dimension and been concatenated along the channel dimension to create the final feature $\mathbf{F} = \text{concat}(\{\text{upsample}(\mathbf{F}_1), ..., \text{upsample}(\mathbf{F}_L)\})$. Formally:
$$
g(\mathbf{X}) = \mathbf{F}, \quad \mathbf{X} \in \mathbb{R}^{N\times N\times N}, \hspace{0.5em} \mathbf{F} \in \mathbb{R}^{d \times N\times N\times N},
$$
where $d$ is the feature channel number that equals the summation of the channel numbers of $\mathbf{F}_i$.
Note that $\mathbf{F}$ has a 3D structure aligned with the input $\mathbf{X}$. 
Given a point query $\mathbf{p} \in \mathbb{R}^3$, continuous features $\mathbf{F}(\mathbf{p})$ at this point can be extracted from $\mathbf{F}$ using trilinear interpolation. 

Then, the encoding at the point $\mathbf{p}$ is fed into a point-wise decoder $f(\cdot)$ to give a binary prediction indicating if the point lies inside or outside the shape:
$$
f(\mathbf{F}, \mathbf{p}) = f(\mathbf{F}(\mathbf{p})) \mapsto \{0, 1\}.
$$
% where $d$ is the feature channel number.
Given the occupancy value $o(\mathbf{p}) \in \{0, 1\}$ at each position $\mathbf{p}$ pre-computed according to the ground truth shape mesh, a binary cross-entropy loss is used to train $g$ and $f$:
\begin{align*}
\text{min}~L_{IF} = \text{BCE}\big(f(\mathbf{F}, \mathbf{p}), ~~o(\mathbf{p})\big).
\end{align*}
Note that the prediction values can also be continuous signed distance values as in \cite{park2019deepsdf}. 

\subsection{Cross-Domain Feature Fusion}%
The training of a naive IF-Net demands ground truth shapes for generating numerous occupancy labels to supervise the output of $f(\cdot)$, yet most of the real scans are unpaired with shape ground truths. So we develop a cross-domain feature fusion module to well transfer the knowledge in the label-rich source domain. Our idea of exploiting the source knowledge stems from an important \textbf{observation}: although synthetic data have distinct local patterns from real scans, the global-level knowledge (\eg, the common structure and coarse shape) of a certain category can be shared in both domains. Besides, the target data with few labels may be enough for learning rich local-level information. 
% On the other hand, while there are few shape references in the target domain, it is expected to learn rich local information from the target data.
On the other hand, IF-Nets conduct shape completion via exploiting local representations, so it is important to extract high-quality local-level features without introducing bias from the source domain.

To this end, we develop a cross-domain feature fusion module (CDFF) for knowledge transfer. First, two shape encoders, $g_s(\cdot)$ and $g_t(\cdot)$, are used for feature extraction of source and target data, generating $\mathbf{F}_s$ and $\mathbf{F}_t$, respectively. As explained above, $\mathbf{F}_s$ contains rich global information, and $\mathbf{F}_t$ is more reliable in providing domain-specific local-level representations. A simple linear combination is adopted to fuse them into $\mathbf{F}$:
$$
\mathbf{F} = \textbf{w} \cdot \mathbf{F}_s + (1-\textbf{w})\cdot \mathbf{F}_t,
$$
where $\textbf{w} \in [0,1]^d$ is a channel-wise weight vector. To better combine the advantages of $\mathbf{F}_s$ and $\mathbf{F}_t$,
the computation of $\textbf{w}$ takes two aspects into account: (i) exploiting the global-level features in $\mathbf{F}_s$ while believing $\mathbf{F}_t$ for local ones (based on our observation), and (ii) learning $\textbf{w}$ adaptively to maintain the flexibility. The adaptive weight vector $\textbf{w}$ is computed as:
$$
\textbf{w} = \alpha \cdot h(\mathbf{F}_s \odot \mathbf{F}_t) + \textbf{w}^0,
$$
where $\alpha \in \mathbb{R}^+$ is the ratio of adaptiveness, $h(\cdot)\mapsto [0, 1]^d$ is a two-layer MLP with a sigmoid activation at the output layer, and $\odot$ indicates the operation of element-wise multiplication plus a global pooling on the spatial dimension (returns a vector of dimension $d$). $\textbf{w}^0 \in [0,1]^d$ is a constant weight vector that implies a prior derived from our observation, of which each value $w^0_i$ is simply defined by a linear mapping as follows:
$$
w^0_i = \frac{l_i}{L+1}, i \in \{1,2,\cdots, d\},
$$
where $l_i \in \{1,2, \cdots, L\}$ indicates which layer the $i$-th channel belongs to.  In this way, the deeper layer one channel is from, the more the computation of $\mathbf{F}$ in this channel relies on $\mathbf{F}_s$, as the deeper layers of IF-Nets capture more global information.
In addition, a clipping operation is applied to $\textbf{w}$ to limit all values to $[0, 1]$.

\subsection{Volume-Consistent Self-training}%
% To mine rich local information from the target data with few shape references, we develop a consistency training framework to encourage the model to learn from data itself.
%The training of a naive IF-Net demands ground-truth shape to generate numerous occupancy labels to supervise the output of $f(\cdot)$, yet most of real scans are unpaired with shape references. So we develop a consistency training framework to encourage the model to learn from data itself. 
% Consistency training is a self-supervised learning approach that works by creating two views of a single data sample via augmentations and forcing the network's prediction consistency on these two views \cite{sohn2020fixmatch}. 
With few shape ground truths in the target domain, it is hard to learn rich information in a supervised manner. A typical way is to adopt self-supervised learning to learn from the data itself. In the task of SCoDA, the \textbf{incompleteness} is a significant characteristic of real scans that hampers the shape completion quality. Thus, we create two views of scans with different levels of incompleteness, and encourage the model to make consistent implicit predictions on their volume occupancy. Such a volume-consistent self-training (VCST) encourages the model to keep robust to various incompleteness and ``imagine'' the missing parts for shape completion.    

% In the classical consistency training for the task of image classification, augmentations are usually in the form of image-level or pixel-level perturbations to enable the model's local robustness. 
% However, in addition to local patterns like noise and sparsity,  in the task of shape completion from real scans, the \textbf{incompleteness} is a significant characteristic of real scans that hampers the shape completion quality.  
% Thus, we design a partition-based consistency training  (PartCT) framework with a novel augmentation strategy that creates views with different incompleteness patterns. 
Specifically, we first adopt an unsupervised clustering algorithm (\eg, $k$-means clustering) to partition the point cloud into $K$ different parts (see Fig.~\ref{fig:method} for examples). The clustering is based only on the spatial positions of points and thus coarsely splits an object scan into multiple components. With the pre-computed partitions, two views $\mathbf{X}^A$ and $\mathbf{X}^B$ can be generated by randomly dropping different clusters of points from the original real scan, with $K_A$ and $K_B$ clusters left, respectively, where $K \ge K_A > K_B$. Besides, two random downsamplings are also conducted on the two views to create different sparsity. Note that the clustering-based augmentation is actually a surface-aware augmentation. Compared with volume-based ones that randomly mask some spatial volumes, our augmentation strategy implies more object-part knowledge to some extent. 

Given the two views of input, their features are extracted by $g(\cdot)$ and then taken by $f(\cdot)$ to give implicit predictions:
\begin{align*}
g&(\mathbf{X}^v) = \mathbf{F}^v, v = A, B, \\
f&(\mathbf{F}^v, \mathbf{p}) \mapsto \{0, 1\}.
\end{align*}
Note that $\mathbf{F}^v$ can also be generated via our CDFF module. Here we simplify the formulation to better present our VCST. The consistency constraint is imposed on the implicit predictions of $f(\mathbf{F}^A, \mathbf{p})$ and $f(\mathbf{F}^B, \mathbf{p})$ with the same point query $\mathbf{p}$:
\begin{align*}
\text{min}~L_{CT} = \text{BCE}\big(f(\mathbf{F}^B, \mathbf{p}), ~~f(\mathbf{F}^A, \mathbf{p})\big),
\end{align*}
where $\text{BCE}(\cdot, \cdot)$ indicates the binary cross-entropy loss. The consistency objective can be viewed as using the prediction from view $A$ as the pseudo label of view $B$, because $\mathbf{X}^B$ has a poorer completeness than $\mathbf{X}^A$ with $K_B < K_A$.
As there will be noises in the implicit predictions due to the limited shape ground truths of target data, using the noisy prediction as the pseudo label will mislead the training. Therefore, a thresholding operation is performed to mask the predictions with low confidence in $f(\mathbf{F}^A, \mathbf{p})$. 
% Following \cite{sohn2020fixmatch}, a threshholding is performed to mask the predictions with low confidence in $f(\mathbf{F}^A, \mathbf{p})$. 
The overall loss function is a summation of $L_{IF}$ and $L_{CT}$.

%% file: 5_dataset.tex
\section{Dataset}%
\label{sec:dataset}

\begin{figure}[ht!] \centering
\includegraphics[width=1.0\linewidth]{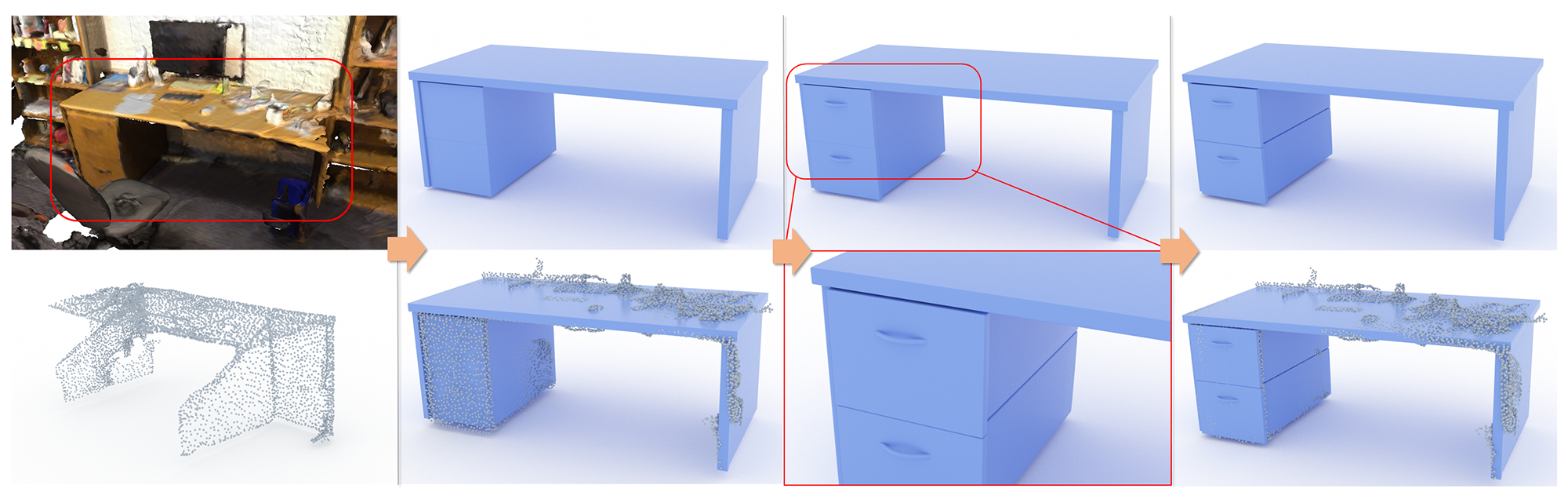}
    \caption{The procedure of 3D model crafting. From left to right: (i) extracting the object scan from a scene; (ii) constructing a coarse mesh frame to fit the scan; (iii) adding fine details; (iv) adjusting the defects detected by inspectors.} 
    \label{fig:dataset_procd}
    \vspace{-3mm}
\end{figure}

\begin{figure*}[ht!] \centering
\includegraphics[width=0.9\linewidth]{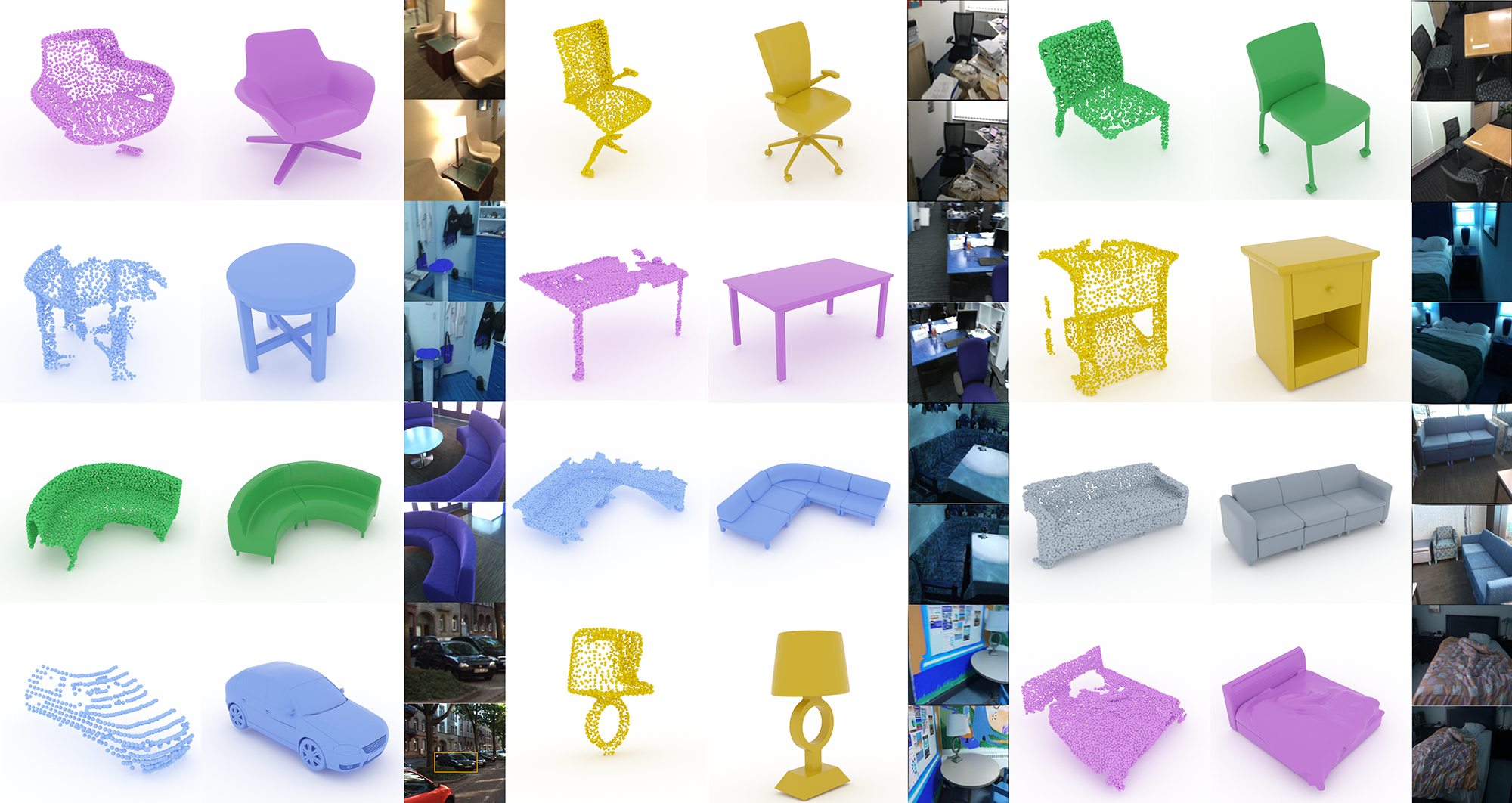}
    \caption{ScanSalon data visualization. Each tuple includes: point cloud (left), created mesh (middle), and photo references (right).} 
    \label{fig:dataset_exh}
    \vspace{-2mm}
\end{figure*}

\paragraph{Overview} % number, classes
The proposed ScanSalon dataset collects 7,644 real scans, of which 800 objects are equipped with 3D shape (artificial) ground truths. All models or scans are from 6 categories of objects, with 5 common indoor objects: chair, desk (or table), sofa, bed, and lamp, and 1 outdoor object: car. Detailed statistics are listed in Tab.~\ref{tab:dataset}. In Fig.~\ref{fig:dataset_exh}, we exhibit some samples of real scans and corresponding 3D shapes in ScanSalon. More visualizations could be seen in the supplementary materials. 
\begin{table}[tb]\centering
    \caption{Statistics of the proposed dataset ScanSalon.}
    \label{tab:dataset}
    \input{tables/0_dataset.tex}
    \vspace{-4mm}
\end{table}

% \input{figures/dataset_exh.tex}
%% I directly comment the original figure, and added ours here. Seems ours should occupy whole width.
%% Three resolutions: figures/DatasetVis_2k.png and figures/DatasetVis_3k.png and figures/DatasetVis_4k.png

\paragraph{Data Collection} % from what dataset
% The synthetic data in our dataset is collected from the ShapeNet dataset~\cite{chang2015shapenet}, which is a large-scale repository of 3D CAD models. The synthetic scans are obtained by using a popular simulation toolbox, BlenSor~\cite{gschwandtner2011blensor}, which supports scanning simulation with different sensors (\textit{e.g.} Velodyne, Kinect, and Time of Flight camera) and parameters, see Fig.~\ref{fig:teaser} for examples. % We also create incompleteness using the method in \cite{bergh2012seeds} for the synthetic data. 
The real scans are collected from two datasets, the ScanNet~\cite{dai2017scannet} and KITTI~\cite{geiger2012we} dataset for indoor and outdoor objects, respectively. The object scans are extracted from a scene point cloud according to the point-level instance segmentation annotations provided by the two datasets, and then rotated and normalized into the same pose and scale, with the aid of pose annotations provided by the Scan2CAD~\cite{avetisyan2019scan2cad} and KITTI~\cite{geiger2012we} dataset. These data are also aligned with models in the ShapeNet database~\cite{chang2015shapenet}, which provides a bunch of synthetic models as the source domain.

% \paragraph{3D Model Creation} % the process of crafting
\paragraph{Shape Annotation for Real Scans}
Two skillful artists participate in the annotation procedure. In addition to the real scans, we provide several photos for each scan from different viewpoints (selected from the videos contained in the ScanNet and KITTI datasets), which provide 2D references to improve the model quality. A brief creation pipeline is shown in Fig.~\ref{fig:dataset_procd}, where the artists are required to create the shapes according to both scans and photo references (see Fig.~\ref{fig:dataset_exh}), overcome the distractions of scan noise, and recover the incompleteness by his(her) rich experience when the photo references are poor or inadequate. They use professional 3D software, Maya\footnote{https://www.autodesk.com/products/maya} to create models. On average, it takes 0.5$\sim$1.0 person-hours to create a 3D model and the creation of all 800 models takes 2.5 months in total. For each created model, another 8 inspectors are invited to verify their (i) recovery extent compared with the real objects in photos and (ii) matching extent with the given scan, and the defects will be fed back to artists for further adjustments until no defects are detected by any inspector. Considering the inaccessibility of the objects in scans from ScanNet and KITTI, we have made our greatest efforts to create \textbf{artificial ground truth} shapes for these scans.   

%% figure is at figures/DatasetMakeProcedure2k.png
%% description:
%%Procedures to make our dataset. First, points are extracted and normalized from ScanNet. Second, the approximated mesh frame is constructed. Third, Mesh surfaces are fit to the point cloud. Some details are added. Last, Surfaces are smoothed by TurboSmooth.

%% file: tables/0_dataset.tex
\resizebox{0.49\textwidth}{!}{
\begin{tabular}{l|cccccc|c}
    \toprule
               & Chair & Desk & Sofa & Bed & Lamp & Car & Total\\
    \midrule
    % Synthetic Scans   & 6,579 & 8,071 & 3,091 & 233 & 2,318 & 3,514 & 23,806  \\
    Real Scans    & 4,651 & 1,630 & 428 & 365 & 133 & 437 & 7,644  \\
    Paired Models   & 497 & 161 & 43 & 36 & 20 & 43 & 800   \\
    \bottomrule
\end{tabular}
}

%% file: 6_experiments.tex
\begin{table*}[t] \centering
    \newcommand{\Frst}[1]{\textcolor{red}{\textbf{#1}}}
    \newcommand{\Scnd}[1]{\textcolor{blue}{\textbf{#1}}}
    \caption{Experiment results on the 3\% and 5\% labels setting of the SCoDA task. The units of CD and mIoU value are $1\times 10^{-3}$ and \%, respectively. \Frst{Red} text indicates the best and \Scnd{blue} text indicates the second best result, respectively (similarly hereinafter).}
    \label{tab:results_6class}
    \input{tables/1_results_6class.tex}
    \vspace{-5mm}
\end{table*}

\section{Experiments}%
\label{sec:Experiments}
\subsection{Settings}
\paragraph{Benchmark setting} We define two settings for the proposed task, which give only 3\% and 5\%  shape ground truths (\textit{e.g.} 47 and 140 chair models randomly sampled from all 4,651 chair scans, respectively) of real scans in the training set, and the rest of samples with corresponding 3D models compose the test set. Note that the 3\% labeled samples are covered in the 5\% labels for observing a cross-setting comparison when given more labels. For evaluation, we consider two metrics to measure the shape completion quality, mIoU: mean volumetric intersection over union (higher is better), and CD: $l_2$ chamfer distance (lower is better). The former measures the matching extent of the defined volumes and the latter measures the surface accuracy. 

\paragraph{Baselines}
We implement four baseline methods to construct the benchmark of the new task. (i) IF-Net: a naive IF-Net~\cite{chibane2020implicit} is adopted without any other designs and is trained on all supervised samples (including synthetic and real labels); (ii) SelfSup: a self-supervised IF-Net that conducts feature-level consistency (by minimizing the distance between two domain feature vectors) on unsupervised samples~\cite{long2017deep}; (iii) PtComp: a domain adaptive point completion method~\cite{chen2019unpaired} is re-implemented to align the real scan inputs with the synthetic ones, incorporated with an extra IF-Net for implicit shape reconstruction. Note the encoder and decoder used in \cite{chen2019unpaired} are revised into a voxel-based 3D CNN to incorporate with IF-Net; (iv) Adversarial: a domain adaptive IF-Net is exploited with adversarial feature learning~\cite{tzeng2017adversarial, wu2020multimodal, leung2021domain} to learn domain-invariant features. The detailed implementation details of each baseline are included in the supplementary materials for the space limit. 

\paragraph{Implementation details} We adopt a 6-layer 3DCNN as the encoder with a 4-layer MLP as the decoder to implement the IF-Nets, using the official codes released by \cite{chibane2020implicit}. Following \cite{chibane2020implicit}, the channel dimension of multi-scale features is 2,583, and a resolution $N$=32 is used for the training of reconstruction. For evaluation, we use a higher resolution with $N$=128. In the CDFF module, $\alpha$ is set to 0.2 to better combine the advantages of $\mathbf{F}_s$ and $\mathbf{F}_t$. For the augmentation strategy in VCST, $K_A$ is randomly selected from $\{7,8\}$, and $K_B$ is randomly selected from $\{5,6,7\}$ for each data sample. The downsampling randomly samples 1k points from each scan (no sampling when has $<$1k points). %Besides, we also try adding noise as another augmentation but has no effect on the performance because of the voxelization operation in IF-Nets. 
A thresholding operation is adopted to filter out predictions with a confidence in $(0.1, 0.9)$ (viewed as unconfident predictions) from view $A$ to guarantee the quality of consistency training~\cite{sohn2020fixmatch}. All models are trained with an Adam optimizer and a learning rate $1\times10^{-4}$, with a mini-batch size 4. $L_{CT}$ only works on unsupervised data. More implementation details of our method are introduced in the supplementary materials. 

\subsection{Results} 
\paragraph{Quantitative comparison}
The performance of baseline methods and ours are listed in Tab.~\ref{tab:results_6class} on the 3\% and 5\% labels settings. It can be seen that our method is superior to all baselines. On the 3\% labels setting, our method achieves $1.32\times10^{-3}$ on CD and 67.32\% on mIoU on average, which is higher than the second best ``Adversarial" by  $\sim$0.3$\times10^{-3}$ CD and $\sim$2\% mIoU. Besides, our method achieves good performances consistently over all 6 categories, most of which get a top rank over all 5 methods. The adversarial domain adaption method achieves the top mIoU among all baseline methods. Besides, our method also achieves the best performance on the 5\% setting for both average CD and mIoU. In addition, it can be observed that an extra 2\% labels bring an improvement of around 5\% mIoU to all methods, and the improvements are especially significant for the categories ``bed'' and ``lamp'', which shows the promising practical value to provide a few more real-data labels to boost the shape completion quality from real scans.  

\begin{figure*}[ht!] \centering
\includegraphics[width=0.81\linewidth]{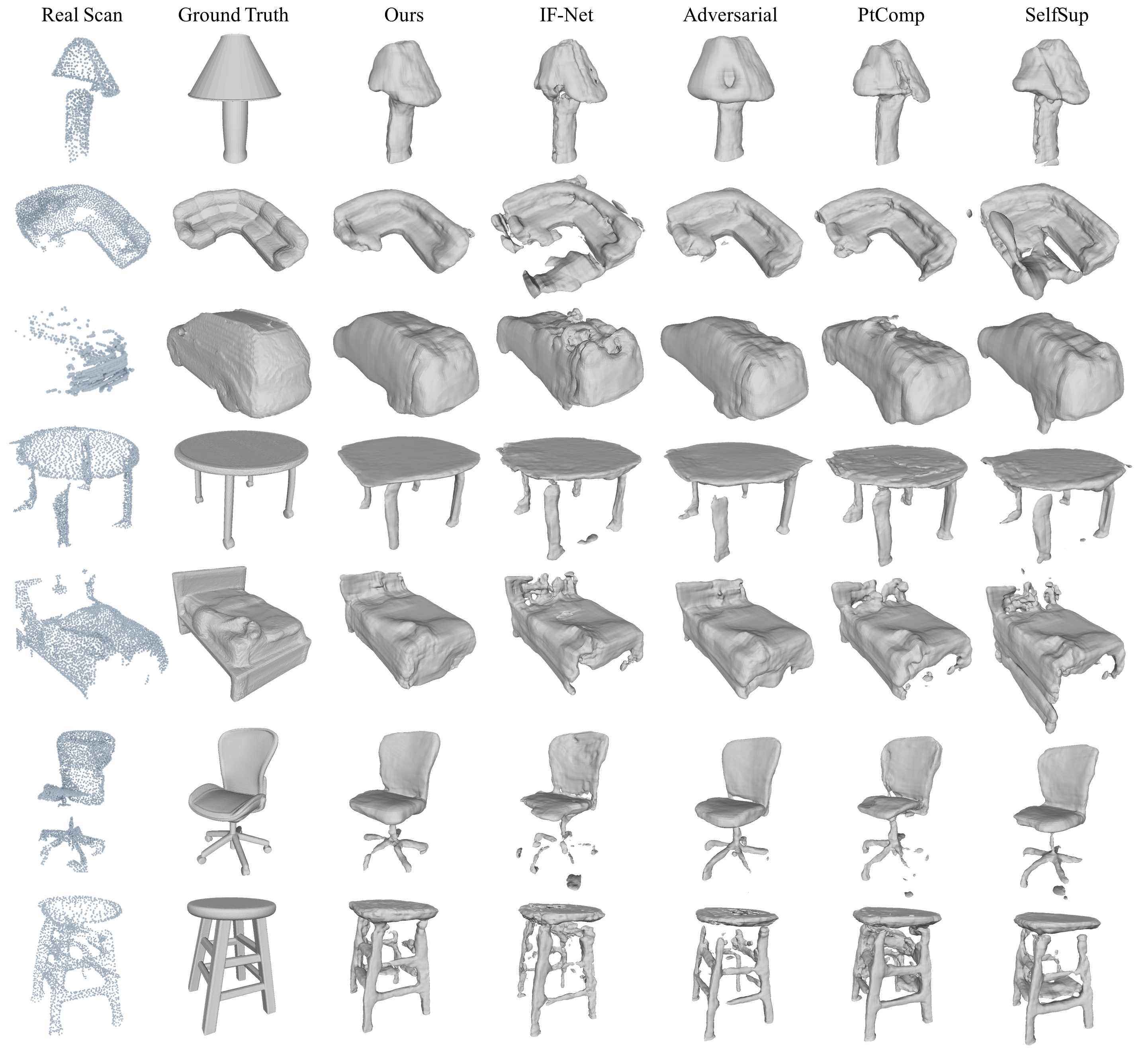}
    \caption{Qualitative comparison between different methods on shape completion with only 3\% labels for training.} 
    \label{fig:qualitative_results}
    \vspace{-6mm}
\end{figure*}

\paragraph{Qualitative comparison}
We visualize the shape completion results of different methods in Fig.~\ref{fig:qualitative_results} to qualitatively compare baselines and our method. With the implicit prediction at each grid, a threshold of 0.5 is used to justify whether an occupancy is predicted here. The resulting occupancy grids with resolution 128 are then transformed into a mesh using the marching cubes algorithm~\cite{cubes1987high}. The visualizations present the superiority of the proposed method, which achieves better completion for missing parts and generates fewer defects almost in all samples of Fig.~\ref{fig:qualitative_results}.

\begin{table}[tb] \centering
    \newcommand{\Frst}[1]{\textcolor{red}{\textbf{#1}}}
    \newcommand{\Scnd}[1]{\textcolor{blue}{\textbf{#1}}}
    \caption{Ablation on using the single module of CDFF or VCST.}
    \label{tab:ablation1}
    \input{tables/2_ablation1.tex}
    \vspace{-3mm}
\end{table}

\begin{table}[tb] \centering
    \newcommand{\Frst}[1]{\textcolor{red}{\textbf{#1}}}
    \newcommand{\Scnd}[1]{\textcolor{blue}{\textbf{#1}}}
    \caption{Analysis of the design of the CDFF module.}
    \label{tab:ablation2}
    \input{tables/3_ablation2.tex}

    \vspace{-5mm}
\end{table}

\begin{table}[tb] \centering
    \newcommand{\Frst}[1]{\textcolor{red}{\textbf{#1}}}
    \newcommand{\Scnd}[1]{\textcolor{blue}{\textbf{#1}}}
    \caption{Analysis of the design of the VCST module.}
    \label{tab:ablation3}
    \input{tables/4_ablation3.tex}
    \vspace{-3mm}
\end{table}

\subsection{Ablation}
\paragraph{Single module}
We first conduct ablative studies on using the CDFF or VCST module only. The results on 3 categories of the 3\% label setting confirm the effectiveness of either of them, see Tab.~\ref{tab:ablation1}. Besides, integrating both two methods brings an extra performance gain, which shows the complementary functions of these two modules. 

\paragraph{CDFF variants}
Next, we implement ablations on different variants of the proposed two modules, and all experiments are conducted on the ``chair'' category under the 3\% label setting. For the CDFF module, we experiment such variants: (i) using the whole features generated by $g_s$ rather than via feature fusion ($\mathbf{F} = \mathbf{F}_s$); (ii) without fusion but using the whole features generated by $g_t$ ($\mathbf{F} = \mathbf{F}_t$); (iii) conducting feature fusion but with a non-adaptive manner, specifically, $w_i$ is set to 0 for the bottom 3 layers and 1 for the top 3 layers (\textbf{Non-Adap.}); (iv) conducting adaptive feature fusion but in a manner contradictory to our observation (\textbf{Contradict.}), where we compute $\mathbf{w}_0$ via $w^0_i = 1-l_i/(L+1)$ instead. The results in Tab.~\ref{tab:ablation2} show that (i) fully using $\mathbf{F}_s$ as $\mathbf{F}$ is equivalent to using a naive IF-Net that has a lower performance; (ii) using $\mathbf{F}_t$ as $\mathbf{F}$ results in a significant performance drop because $g_t$ can hardly learn high-quality global representations with a few labels; (iii) using a non-adaptive feature fusion that complies with our observation also leads to a performance drop but performs better than those without fusion; (iv) a fusion contradictory to our observation brings a $\sim$2\% mIoU drop, which further supports the reasonableness of our observation.   

\paragraph{VCST variants}
Related to our VCST design, we try different combinations to achieve the best augmentation strategy. Alternative strategies include random downsampling, surface-aware sampling (clustering-based), volume-aware sampling, and adding noise. Among them, adding noise has no effect on the performance whatever other augmentations are used, which may be caused by the voxelization operation in IF-Nets. Thus we try the following variants with the combination of the other three augmentation strategies. Note that we control the dropping ratio to guarantee the expected number of dropped points (if only using dropping) are the same for all settings. The results in Tab.~\ref{tab:ablation3} show that: (1) the consistency training with random downsampling only crashed (the 2nd row), using volume-aware or surface-aware augmentation only can still work, but the surface-aware one performs better (the 3rd and 4th row), and the 4th row is the version used in our standard VCST; (3) the volume-aware augmentation can bring minor improvements on mIoU with or without random downsampling (the 3rd and 5th row), which is worse than our surface-aware one, and using both two of them even makes a negative effect (the last row). The reason for the failure of using random downsampling only is that the consistency training fails when two views are very similar in a global fashion. An intuitive explanation for (3) is that compared with our surface-aware augmentation, the incompleteness created by the volume-aware one has a larger gap with real scans. The qualitative comparisons between different variants are included in the supplementary materials.

%% file: tables/1_results_6class.tex
\newcommand{\cfd}{CD$\downarrow$\xspace}
\newcommand{\iou}{mIoU$\uparrow$\xspace}
\newcommand{\chair}{Chair\xspace}
\newcommand{\desk}{Desk\xspace}
\newcommand{\sofa}{Sofa\xspace}
\newcommand{\bed}{Bed\xspace}
\newcommand{\lamp}{Lamp\xspace}
\newcommand{\mobile}{Car\xspace}
\newcommand{\avg}{Average\xspace}

\makebox[\textwidth]{\small (a) Results on the 3\% labels setting.} 
\resizebox{\textwidth}{!}{
    \small
\begin{tabular}{l||*{2}{c}|*{2}{c}|*{2}{c}|*{2}{c}|*{2}{c}|*{2}{c}||*{2}{c}}
    \toprule
    & \multicolumn{2}{c|}{\chair} & \multicolumn{2}{c|}{\desk} & \multicolumn{2}{c|}{\sofa} & \multicolumn{2}{c|}{\bed} & \multicolumn{2}{c|}{\lamp} & \multicolumn{2}{c||}{\mobile} & \multicolumn{2}{c}{\avg} \\
    Method & \cfd & \iou & \cfd & \iou & \cfd & \iou & \cfd & \iou & \cfd & \iou & \cfd & \iou & \cfd & \iou  \\
    \midrule
    IF-Net & \Scnd{1.57} & 56.10 & 2.44 & 43.04 & 0.65 & 79.03 & 1.64 
                  & 67.30 & \Scnd{1.67} & 39.89 & 0.74 & 74.77 & 1.45 & 60.02 \\
    SelfSup& \Frst{1.49} & \Scnd{58.55} & 3.49 & 42.97 & 0.55 & 81.16 & 1.59 
                  & 68.58 & 2.41 & 51.42 & 0.62 & 78.75 & 1.69 & 63.57 \\
    PtComp  & 1.61 & 57.33 & \Frst{2.16} & 44.26 & 0.51 & 79.90 & \Scnd{1.52} & 68.23
                  & 1.95 & 46.97 & \Scnd{0.59} & \Scnd{80.35} & \Scnd{1.39} & 62.84  \\
    Adversarial  & 1.74 & 58.54 & 2.99 & \Scnd{46.02} & \Scnd{0.46} & \Scnd{81.42} & \Frst{1.37} & \Scnd{71.32} 
                  & 2.43 & \Scnd{56.39} & 0.67 & 78.91 & 1.61 & \Scnd{65.43}  \\
    \textbf{Ours}  & 1.58 & \Frst{60.77} & \Scnd{2.36} & \Frst{48.62} & \Frst{0.42} & \Frst{82.00} & 1.57
                  & \Frst{73.05} & \Frst{1.62} & \Frst{58.57} & \Frst{0.41} & \Frst{80.96} & \Frst{1.32} & \Frst{67.32}  \\
    \bottomrule
\end{tabular}
}

\vspace{0.5em}
\makebox[\textwidth]{\small (b) Results on the 5\% labels setting.} 
\resizebox{\textwidth}{!}{
    \small
\begin{tabular}{l||*{2}{c}|*{2}{c}|*{2}{c}|*{2}{c}|*{2}{c}|*{2}{c}||*{2}{c}}
    \toprule
    & \multicolumn{2}{c|}{\chair} & \multicolumn{2}{c|}{\desk} & \multicolumn{2}{c|}{\sofa} & \multicolumn{2}{c|}{\bed} & \multicolumn{2}{c|}{\lamp} & \multicolumn{2}{c||}{\mobile} & \multicolumn{2}{c}{\avg} \\
    Method & \cfd & \iou & \cfd & \iou & \cfd & \iou & \cfd & \iou & \cfd & \iou & \cfd & \iou & \cfd & \iou  \\
    \midrule
    IF-Net & 1.88 & 56.98 & 2.14 & 44.87 & 0.50 & 82.04 & 0.66
                  & 76.05 & 1.72 & 51.33 & 0.52 & 80.13 & 1.24 & 65.23  \\
    SelfSup      & 2.08 & 59.42 & 2.73 & 46.39 & 0.51 & 82.25 & 0.61
                  & 77.22 & 1.46 & 62.02 & 0.43 & 81.89 & 1.09 & 68.06  \\
    PtComp  & \Frst{1.34} & 57.98 & \Frst{1.83} & 46.20 & \Scnd{0.32} & 82.66 & 0.61
                  & 79.07 & \Scnd{1.44} & 61.61 & \Scnd{0.43} & \Scnd{81.89} & \Frst{1.00} & 68.24 \\
    Adversarial  & 1.71 & \Scnd{60.58} & 2.13 & \Scnd{48.46} & 0.41 & \Frst{83.54} & \Scnd{0.51}
                  & \Scnd{80.81} & \Frst{1.33} & \Scnd{64.22} & \Frst{0.41} & 81.86 & 1.08 & \Scnd{69.91} \\
    \textbf{Ours}  & \Scnd{1.37} & \Frst{61.48} & \Scnd{2.09} & \Frst{50.93} & \Frst{0.31} & \Scnd{82.71} & \Frst{0.41}
                  & \Frst{82.27} & 1.57 & \Frst{67.80} & 0.46 & \Frst{83.12} & \Scnd{1.04} & \Frst{71.39}  \\
    \bottomrule
\end{tabular}
}

%% file: tables/2_ablation1.tex
\newcommand{\cfd}{CD$\downarrow$\xspace}
\newcommand{\iou}{mIoU$\uparrow$\xspace}

\newcommand{\abla}{CDFF+VCST\xspace}
\newcommand{\ablba}{CDFF only\xspace}
\newcommand{\ablbb}{VCST only\xspace}

\newcommand{\chair}{Chair\xspace}
\newcommand{\desk}{Bed\xspace}
\newcommand{\sofa}{Lamp\xspace}

\resizebox{0.48\textwidth}{!}{
\small
\begin{tabular}{l||*{2}{c}|*{2}{c}|*{2}{c}}
    \toprule
    & \multicolumn{2}{c|}{\chair} & \multicolumn{2}{c|}{\desk} & \multicolumn{2}{c}{\sofa} \\
    &  \cfd &  \iou & \cfd & \iou &  \cfd & \iou \\
    \midrule
    \abla        & 1.58 & \Frst{60.77} &   \Frst{1.57} & \Frst{73.05}   & \Frst{1.62} & \Frst{58.57} \\
    
    \ablba       & \Frst{1.49} & 58.55 &   1.91 & 71.19   & 1.73 & 53.18 \\
    \ablbb       & 2.08 & 59.42 &   1.63 & 72.62   & 1.85 & 50.18 \\
    \bottomrule
\end{tabular}
}

%% file: tables/3_ablation2.tex
\newcommand{\cfd}{CD$\downarrow$\xspace}
\newcommand{\iou}{mIoU$\uparrow$\xspace}

\newcommand{\abla}{\textbf{Ours}\xspace}
\newcommand{\ablaa}{$\mathbf{F} = \mathbf{F}_s$\xspace}
\newcommand{\ablba}{$\mathbf{F} = \mathbf{F}_t$\xspace}
\newcommand{\ablbb}{Non-Adap.\xspace}
\newcommand{\ablbc}{Contrad.\xspace}

\newcommand{\chair}{Chair\xspace}
\newcommand{\desk}{Bed\xspace}
\newcommand{\sofa}{Lamp\xspace}

\resizebox{0.48\textwidth}{!}{
\small
\begin{tabular}{*{1}{l} ||  *{3}{c} || *{2}{c}}
    \toprule
    & Fusion &  Observ. &  Adaptive & \cfd & \iou  \\
    \midrule
    \abla & \checkmark & \checkmark & \checkmark & \Frst{1.49} & \Frst{58.55} \\
    \ablaa &  &  &  & 1.57~(+0.08) & 56.10~(-2.45) \\
    \ablba &  &  &  & 2.22~(+0.73) & 54.98~(-3.57) \\
    \ablbb & \checkmark & \checkmark &  & 1.92~(+0.43) & 57.77~(-0.78) \\
    \ablbc & \checkmark & & \checkmark & 1.96~(+0.47) & 56.62~(-1.93) \\
    \bottomrule
\end{tabular}
}

%% file: tables/4_ablation3.tex
\newcommand{\cfd}{CD$\downarrow$\xspace}
\newcommand{\iou}{mIoU$\uparrow$\xspace}
\newcommand{\compa}{Random}
\newcommand{\compb}{Volume}
\newcommand{\compc}{Surface}
\newcommand{\compd}{Consist.}

\resizebox{0.4\textwidth}{!}{
\small
\begin{tabular}{*{3}{c} || *{2}{c}}
    \toprule
    \compa &  \compb &  \compc  & \cfd & \iou  \\
    \midrule
    \multicolumn{3}{c||}{w/o consistency training}  & \Frst{1.57} & 56.10 \\
    % \hline
    % \checkmark & & \checkmark & \Frst{1.57}~(-\textbf{0.00}) & 56.25~(+0.15) \\
    \checkmark & & &  3.65~(+2.08) & 47.60~(-8.50) \\
    & \checkmark & &  2.03~(+0.46) & 57.55~(+1.45) \\
    & & \checkmark &  2.37~(+0.80) & 59.03~(+2.93) \\
    
    \checkmark & & \checkmark &  2.08~(+0.51) & \Frst{59.42}~(+3.32) \\
    \checkmark & \checkmark & &  2.37~(+0.80) & 58.13~(+2.03) \\
    \checkmark & \checkmark & \checkmark & 2.30~(+0.73) & 58.55~(+2.45) \\
    \bottomrule
\end{tabular}
}

%% file: 7_conclusions.tex
\section{Conclusion}%
\label{sec:Conclusion}
In this paper, we propose a new task, domain adaptive shape completion of real scans, which aims to transfer the knowledge in the label-rich synthetic domain into the more challenging real domain. We construct a new dataset that contains elaborate 3D models created by professional artists for 10\% of real scans in 6 categories, which can serve for the learning of deep reconstruction models and evaluation. Besides, we develop a new domain adaptive shape completion framework with two novel modules, cross-domain feature fusion and volume-consistent self-training, to better exploit both the synthetic and real data. Extensive experiments validate the superiority and effectiveness of our method and a benchmark is constructed on the proposed task. 

% \paragraph{Limitations} Given a few shape labels of real scans, the current baseline methods and ours can not achieve satisfying results, which may result from the great incompleteness of reals scans and the huge domain gap between the synthetic and real data. Our future work is to develop more strong methods to handle the domain gap and achieve more impressive reconstruction results. Despite this, the proposed task is of promising application value that is worth further exploration. 
\small \paragraph{Acknowledge} The work was supported in part by NSFC-62172348, the Basic Research Project No.~HZQB-KCZYZ-2021067 of Hetao Shenzhen-HK S\&T Cooperation Zone. It was also partially supported by NSFC-62172348, Outstanding Yound Fund of Guangdong Province with No.~2023B1515020055, the National Key R\&D Program of China with grant No.~2018YFB1800800, by Shenzhen Outstanding Talents Training Fund 202002, by Guangdong Research Projects No.~2017ZT07X152 and No.~2019CX01X104, by the Guangdong Provincial Key Laboratory of Future Networks of Intelligence (Grant No.~2022B1212010001), and by Shenzhen Key Laboratory of Big Data and Artificial Intelligence (Grant No.~ZDSYS201707251409055).

%% file: figures/Supp_pcd_comp.tex
\begin{figure}[htbp] \centering
    \includegraphics[width=0.48\textwidth]{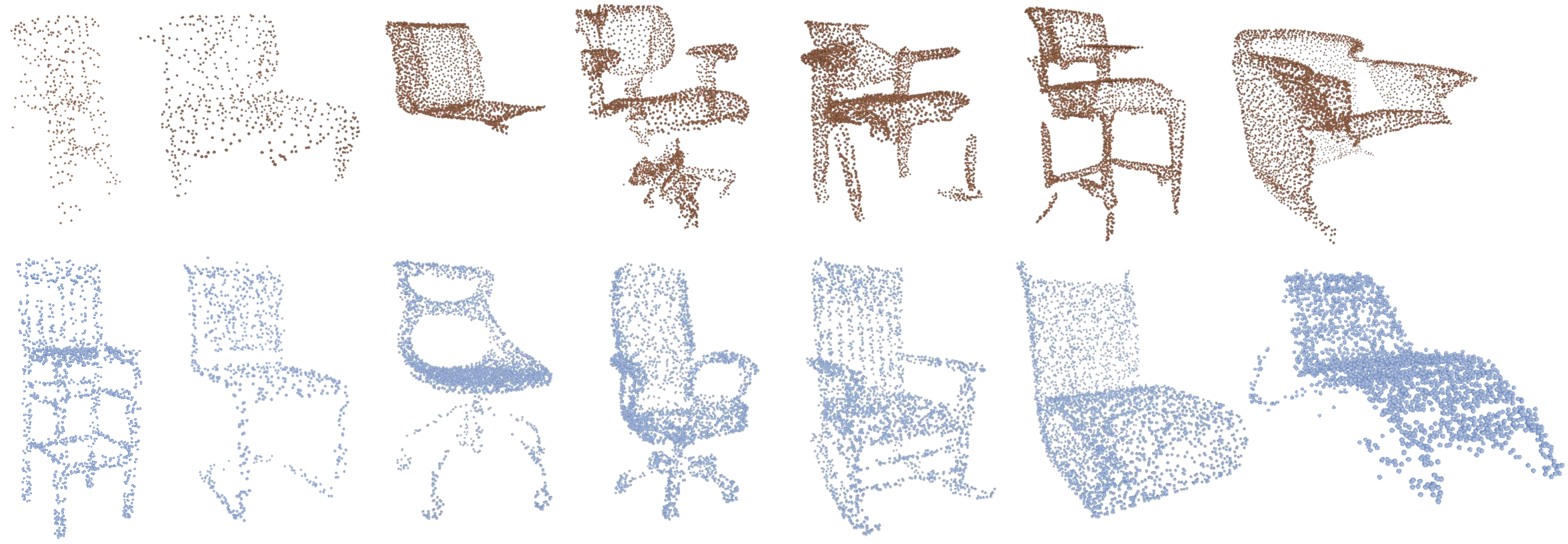}
    \caption{Comparison between real scans (the first row, from the ScanNet dataset) and synthetic scans (the second row, simulated from models in the ShapeNet dataset).} \label{fig:pcd_comp}
    \vspace{-5mm}
\end{figure}

%% file: figures/Supp_qualitative_results.tex
\begin{figure*}[ht!] \centering
\includegraphics[width=0.97\linewidth]{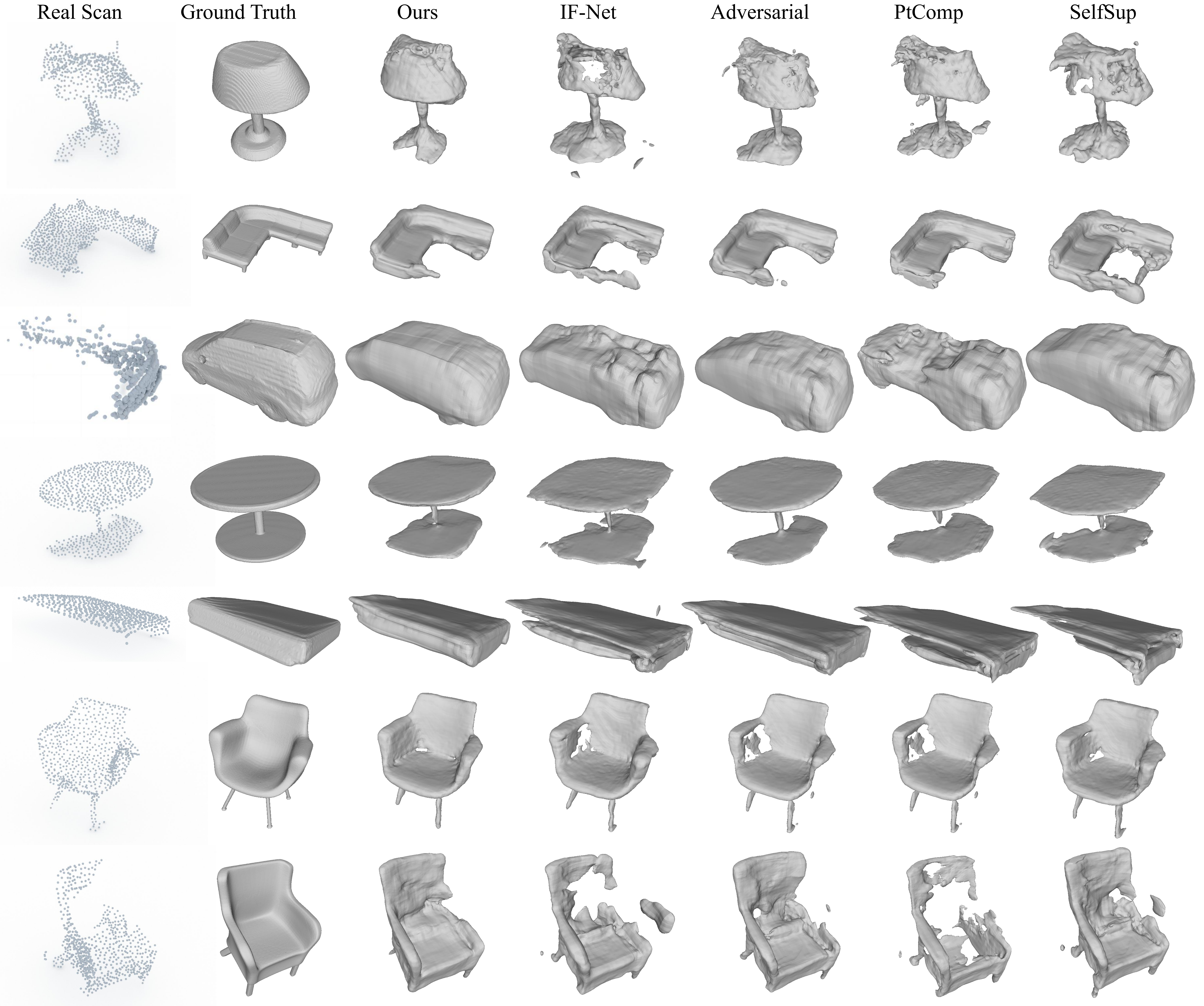}
    \caption{Qualitative comparison between different methods on shape completion with only 5\% labels for training.} 
    \label{fig:qualitative_results}
    \vspace{-3mm}
\end{figure*}

%% file: tables/5_supp_ablation1.tex
\newcommand{\cfd}{CD$\downarrow$\xspace}
\newcommand{\iou}{mIoU$\uparrow$\xspace}
\newcommand{\chair}{Chair\xspace}
\newcommand{\desk}{Desk\xspace}
\newcommand{\sofa}{Sofa\xspace}
\newcommand{\bed}{Bed\xspace}
\newcommand{\lamp}{Lamp\xspace}
\newcommand{\mobile}{Car\xspace}
\newcommand{\avg}{Average\xspace}

\newcommand{\abla}{CDFF+VCST\xspace}
\newcommand{\ablba}{CDFF only\xspace}
\newcommand{\ablbb}{VCST only\xspace}

\resizebox{\textwidth}{!}{
    \small
\begin{tabular}{l||*{2}{c}|*{2}{c}|*{2}{c}|*{2}{c}|*{2}{c}|*{2}{c}||*{2}{c}}
    \toprule
    & \multicolumn{2}{c|}{\chair} & \multicolumn{2}{c|}{\desk} & \multicolumn{2}{c|}{\sofa} & \multicolumn{2}{c|}{\bed} & \multicolumn{2}{c|}{\lamp} & \multicolumn{2}{c||}{\mobile} & \multicolumn{2}{c}{\avg} \\
    Method & \cfd & \iou & \cfd & \iou & \cfd & \iou & \cfd & \iou & \cfd & \iou & \cfd & \iou & \cfd & \iou  \\
    \midrule
    \abla        & 1.58 & \Frst{60.77} & \Frst{2.36} & \Frst{48.62} & \Frst{0.42} & \Frst{82.00} & \Frst{1.57} & \Frst{73.05}   & \Frst{1.62} & \Frst{58.57} & \Frst{0.41} & \Frst{80.96} & \Frst{1.33} & \Frst{67.33} \\
    
    \ablba       & \Frst{1.49} & 58.55 & 2.84 & 48.20 & 0.53 & 79.42 & 1.91 & 71.19   & 1.73 & 53.18 & 0.57 & 79.17 & 1.51 & 64.95 \\
    \ablbb       & 2.08 & 59.42  & 2.89 & 46.86 & 0.43 & 81.60 &   1.63 & 72.62   & 1.85 & 50.18 & 0.62 & 78.62 & 1.58 & 64.88\\
    \bottomrule
\end{tabular}
}

%% file: figures/Supp_ablation_fig.tex
\def\inserta#1{\includegraphics[trim={0 3cm 0 10cm},clip,width=\linewidth]{#1}}
\def\insertb#1{\includegraphics[trim={0 6cm 0 7cm},clip,width=0.8\linewidth]{#1}}
\def\insertc#1{\includegraphics[trim={0 8cm 0 7cm},clip,width=\linewidth]{#1}}
\def\insertd#1{\includegraphics[trim={0 5cm 0 9cm},clip,width=\linewidth]{#1}}
\def\inserte#1{\includegraphics[trim={0 4cm 0 12cm},clip,width=\linewidth]{#1}}
\def\insertf#1{\includegraphics[trim={0 5cm 0 11cm},clip,width=\linewidth]{#1}}
\newcommand{\Frst}[1]{\textcolor{red}{\textbf{#1}}}

\begin{figure*}[htbp]
	\centering
	\begin{minipage}{0.135\linewidth}
		\centering
		\inserta{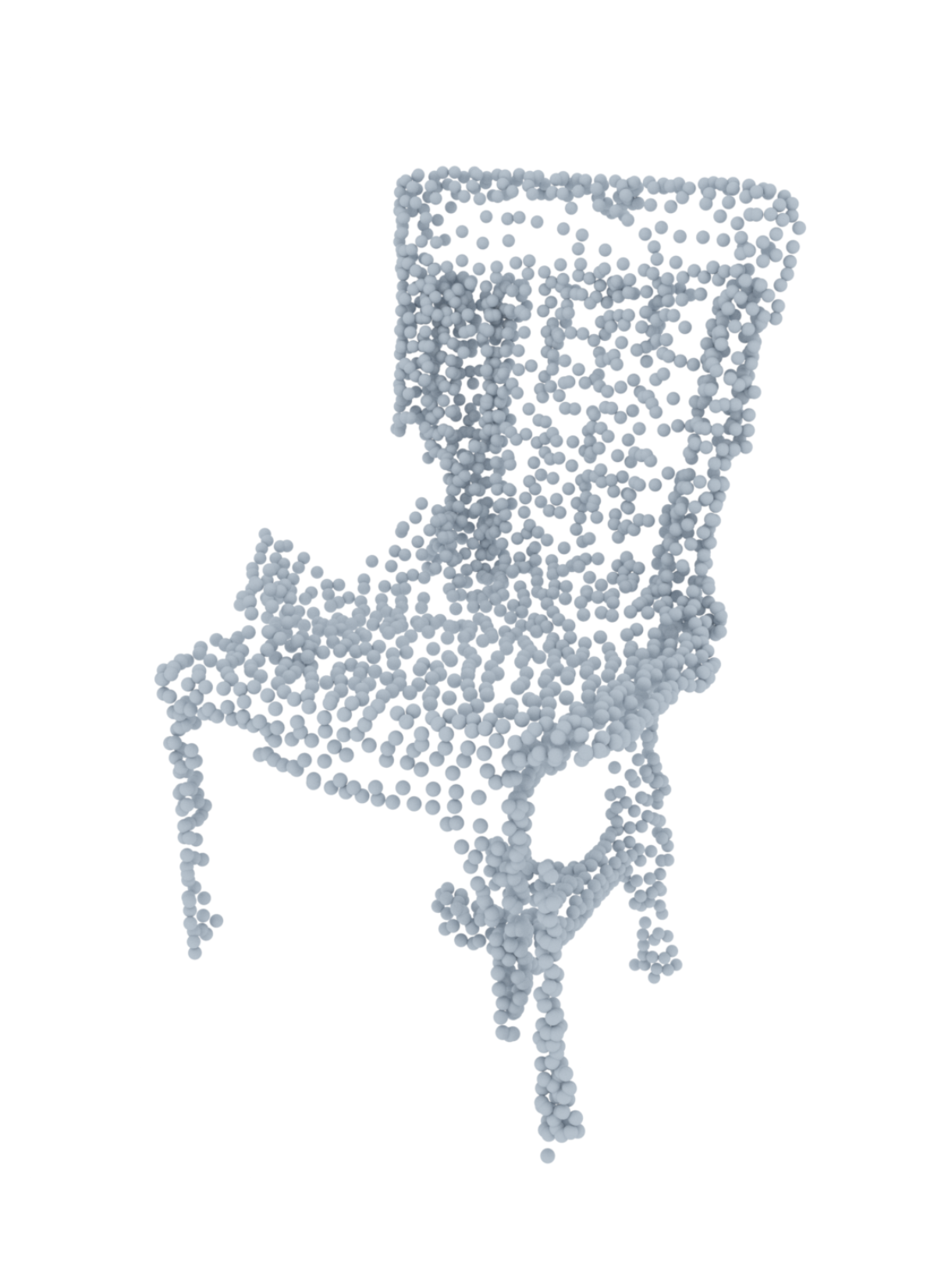} \\
		\insertb{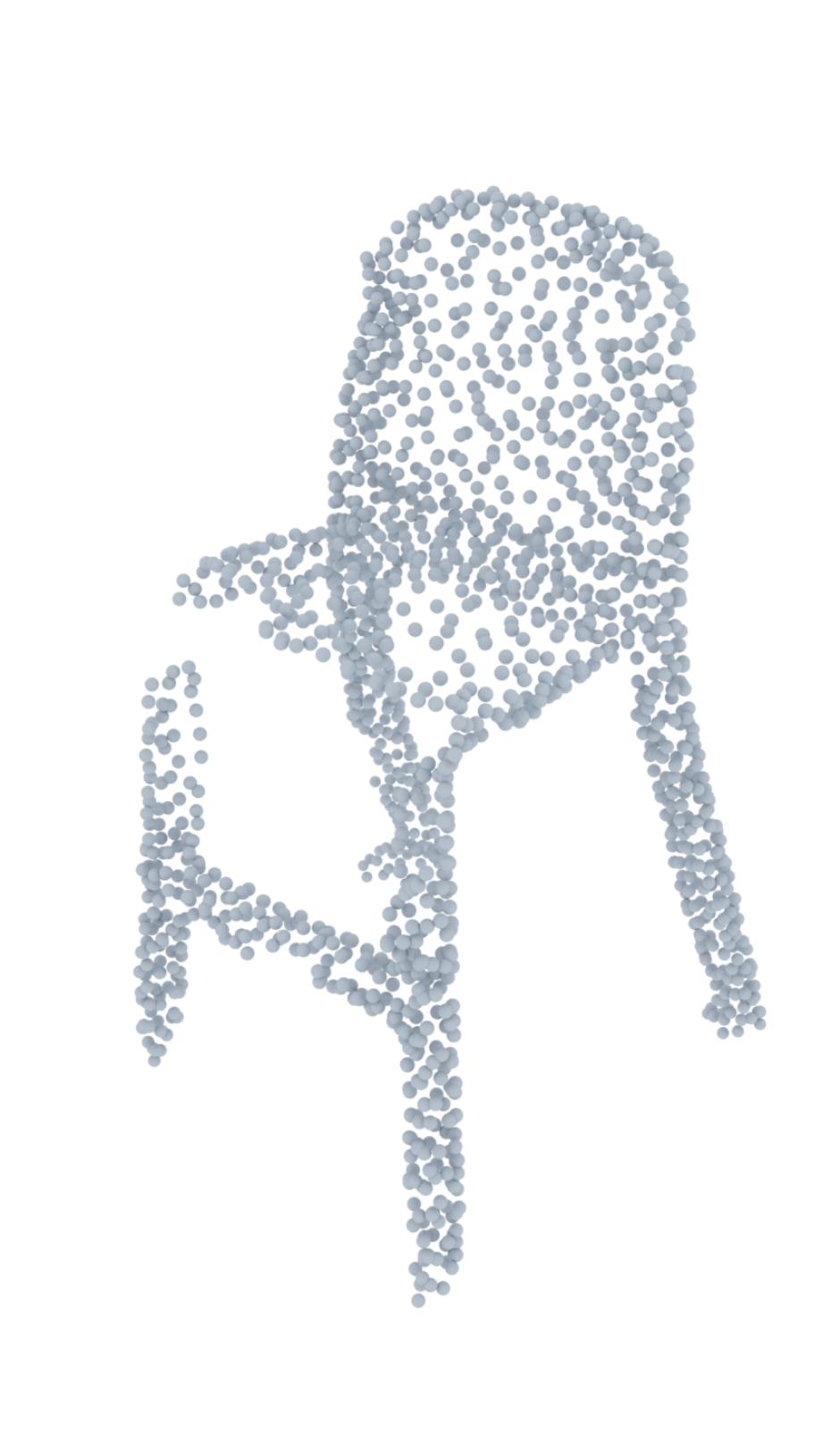} \\
		\insertc{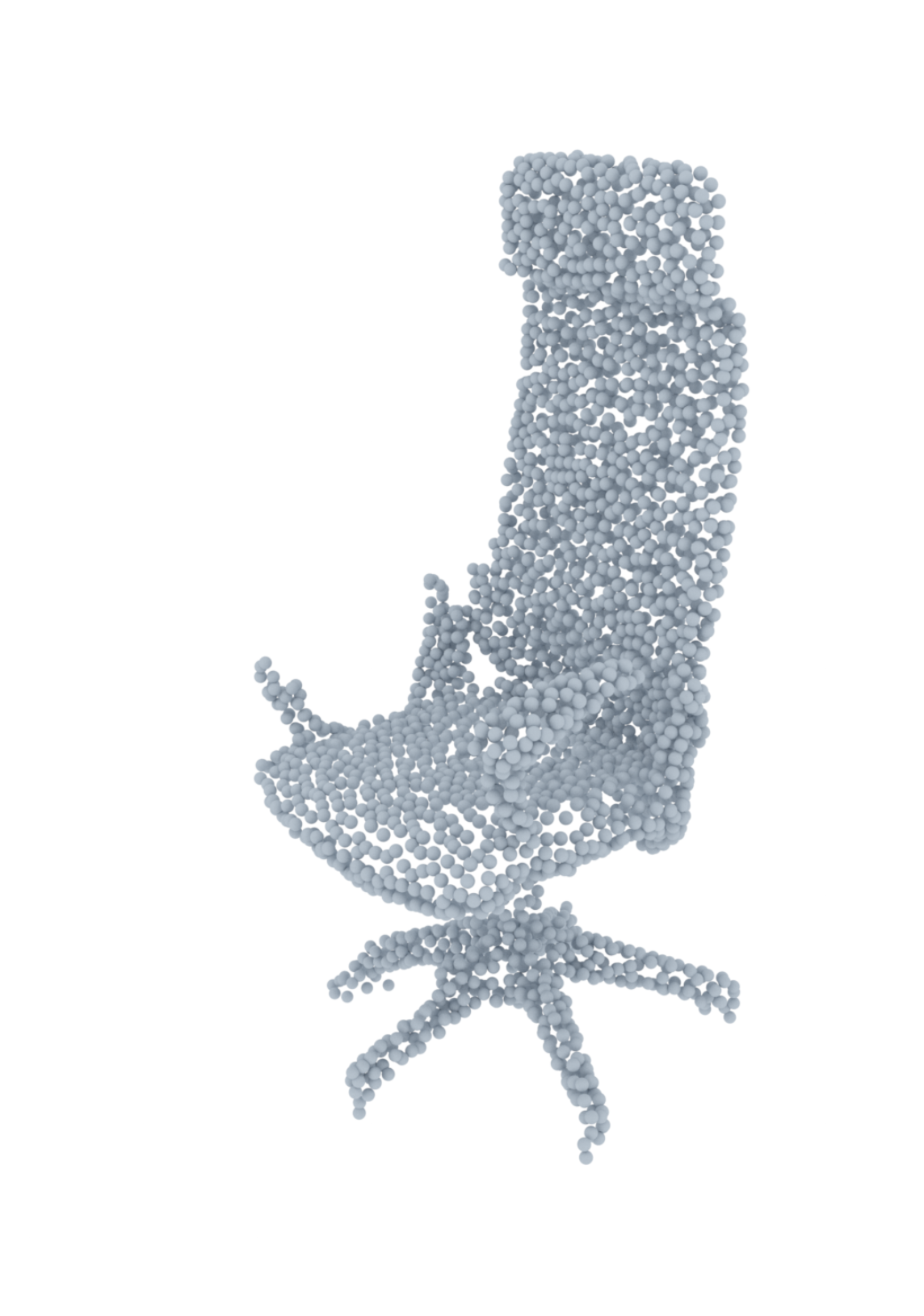} \\
		(1)
	\end{minipage}
	\begin{minipage}{0.135\linewidth}
		\centering
		\inserta{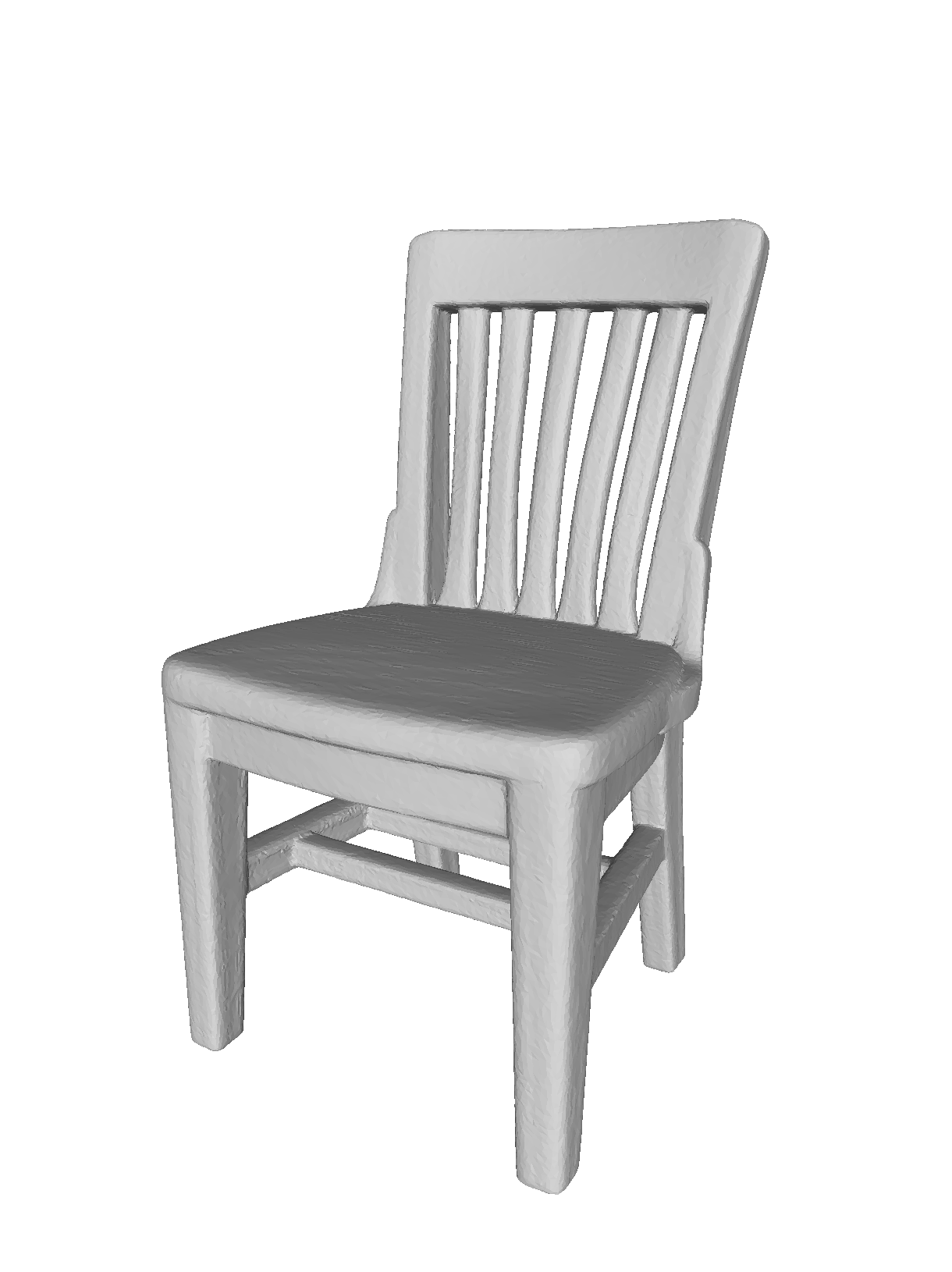} \\
		\insertb{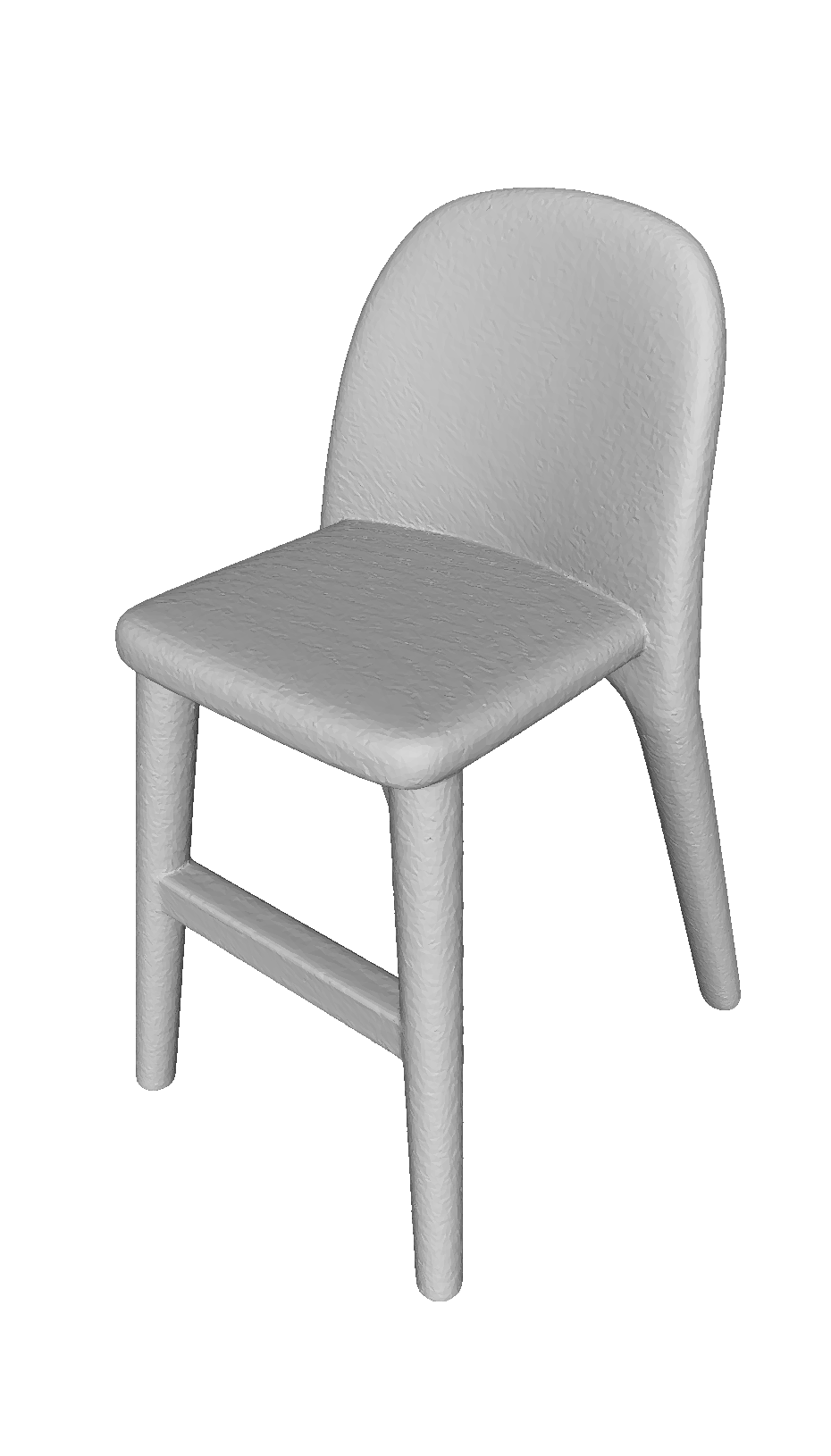} \\
		\insertc{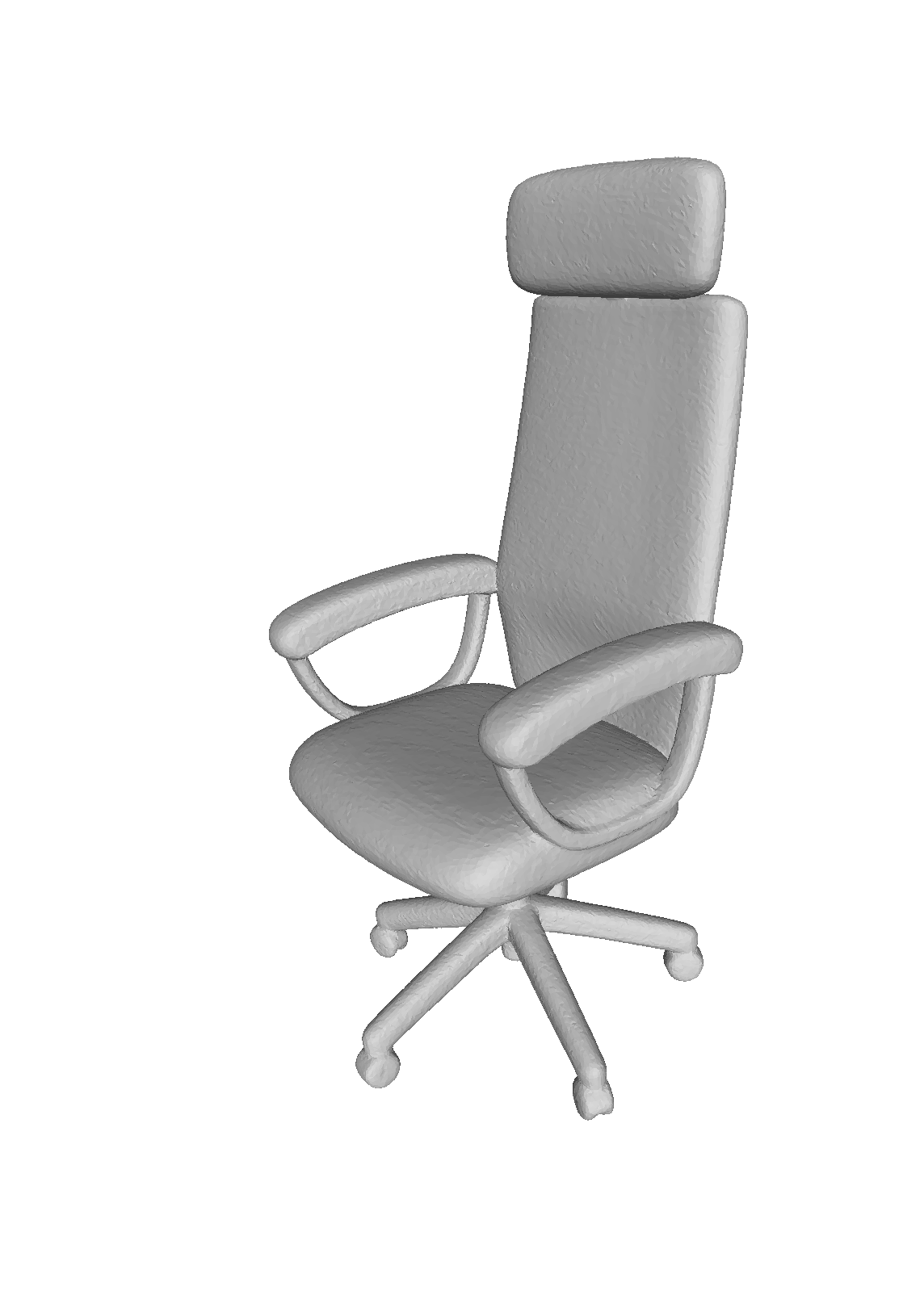} \\
		(2)
	\end{minipage}
	\begin{minipage}{0.135\linewidth}
		\centering
		\inserta{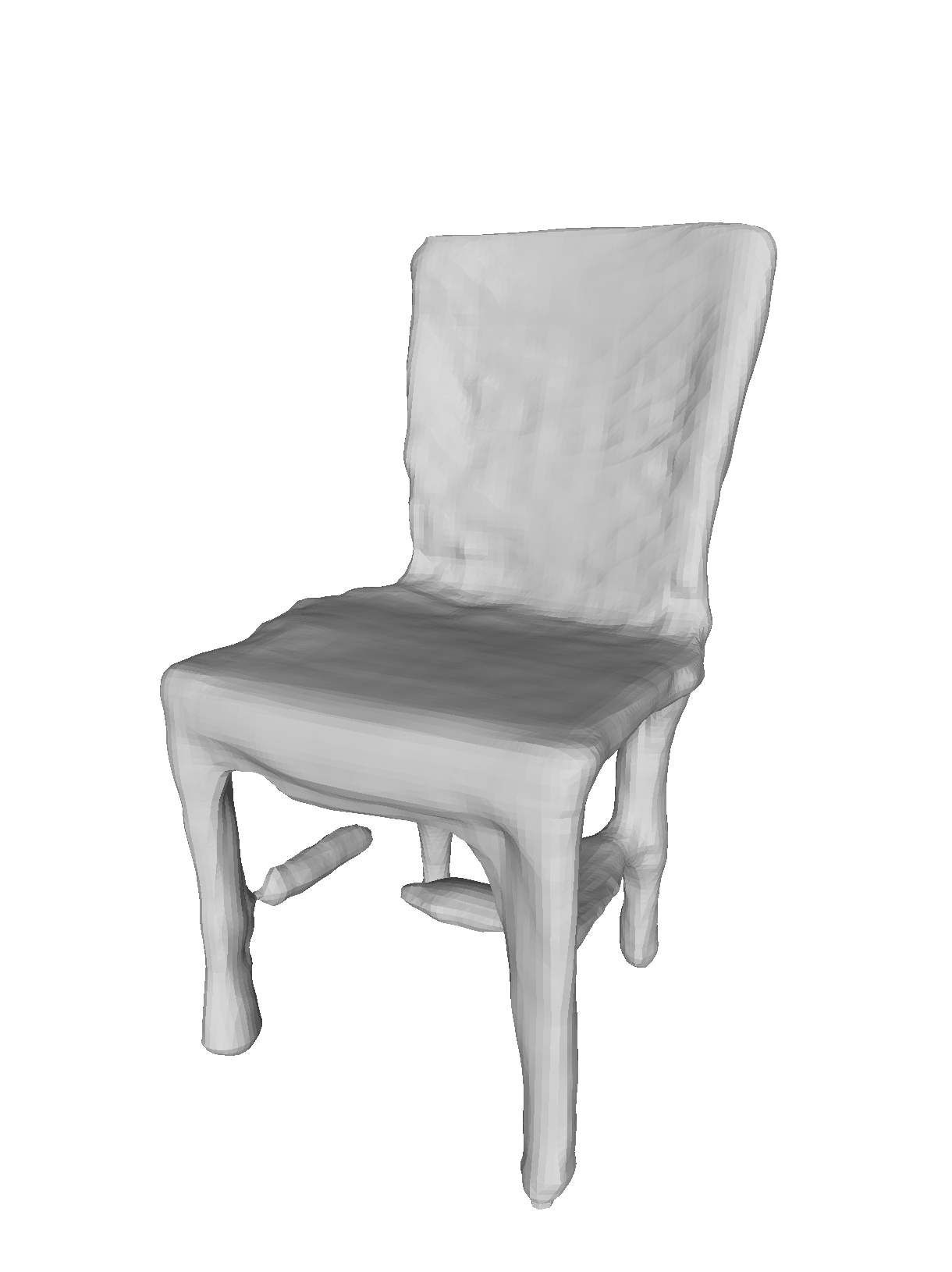} \\
		\insertb{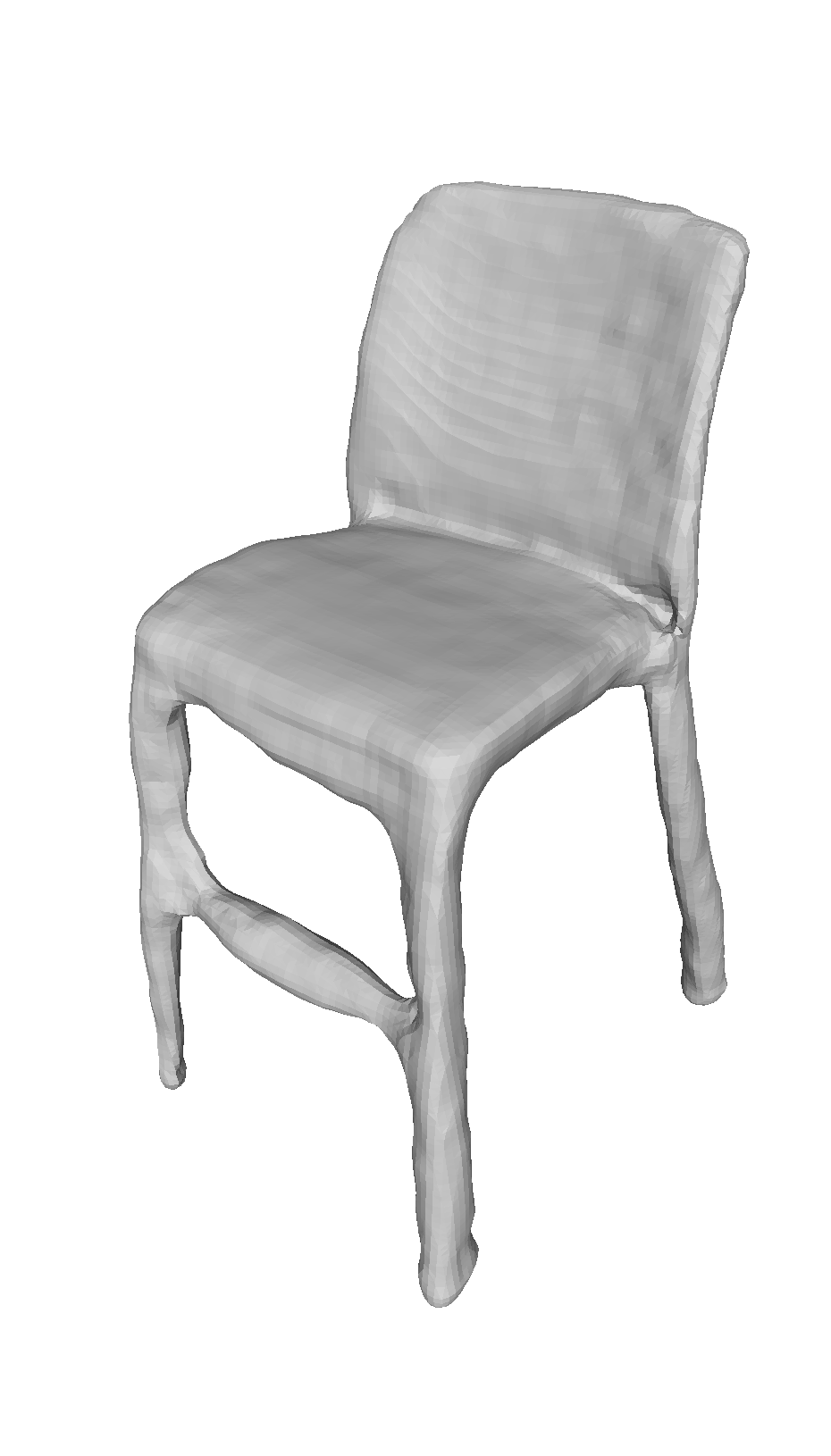} \\
		\insertc{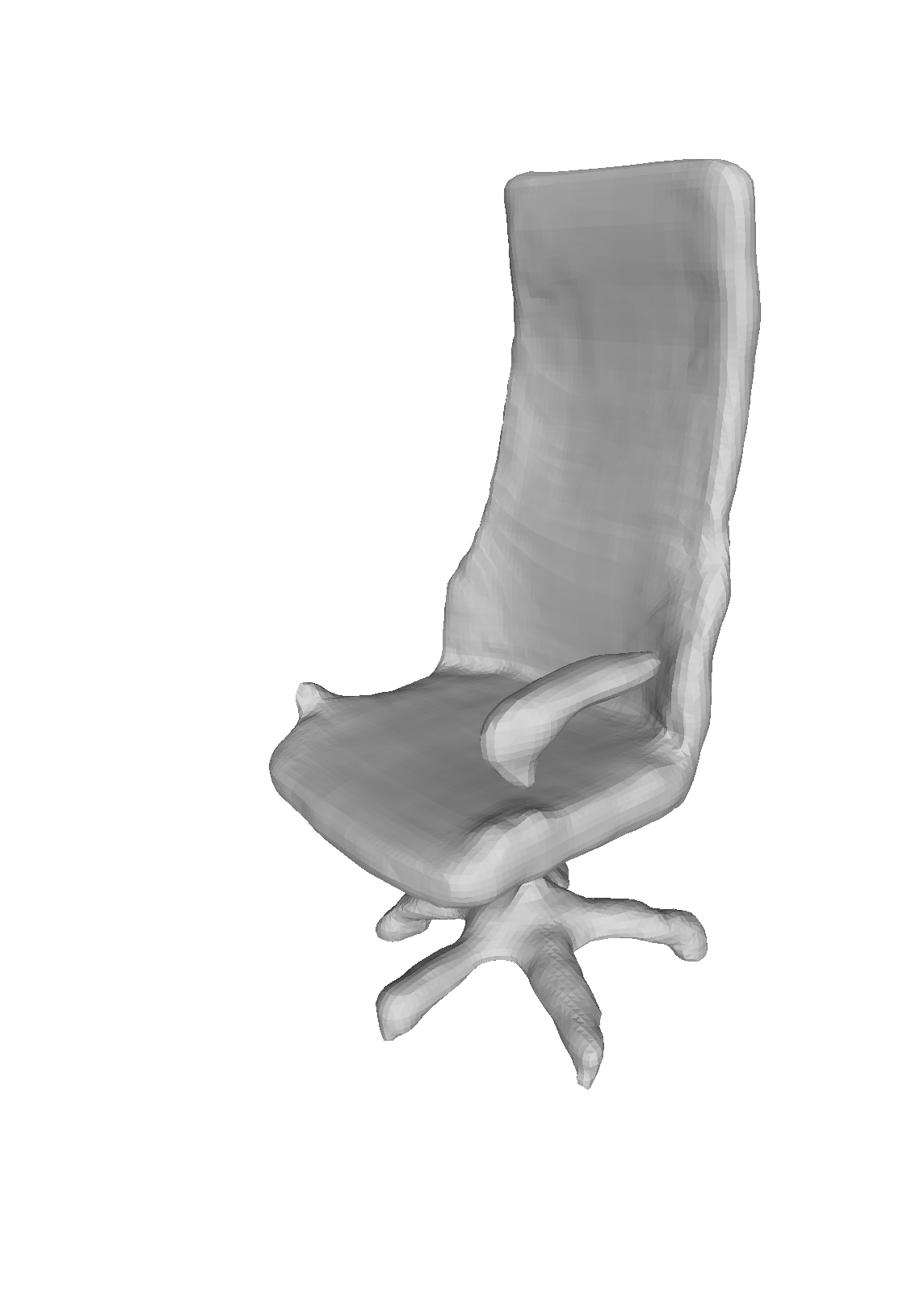} \\
		\Frst{(3)}
	\end{minipage}
	\begin{minipage}{0.135\linewidth}
		\centering
		\inserta{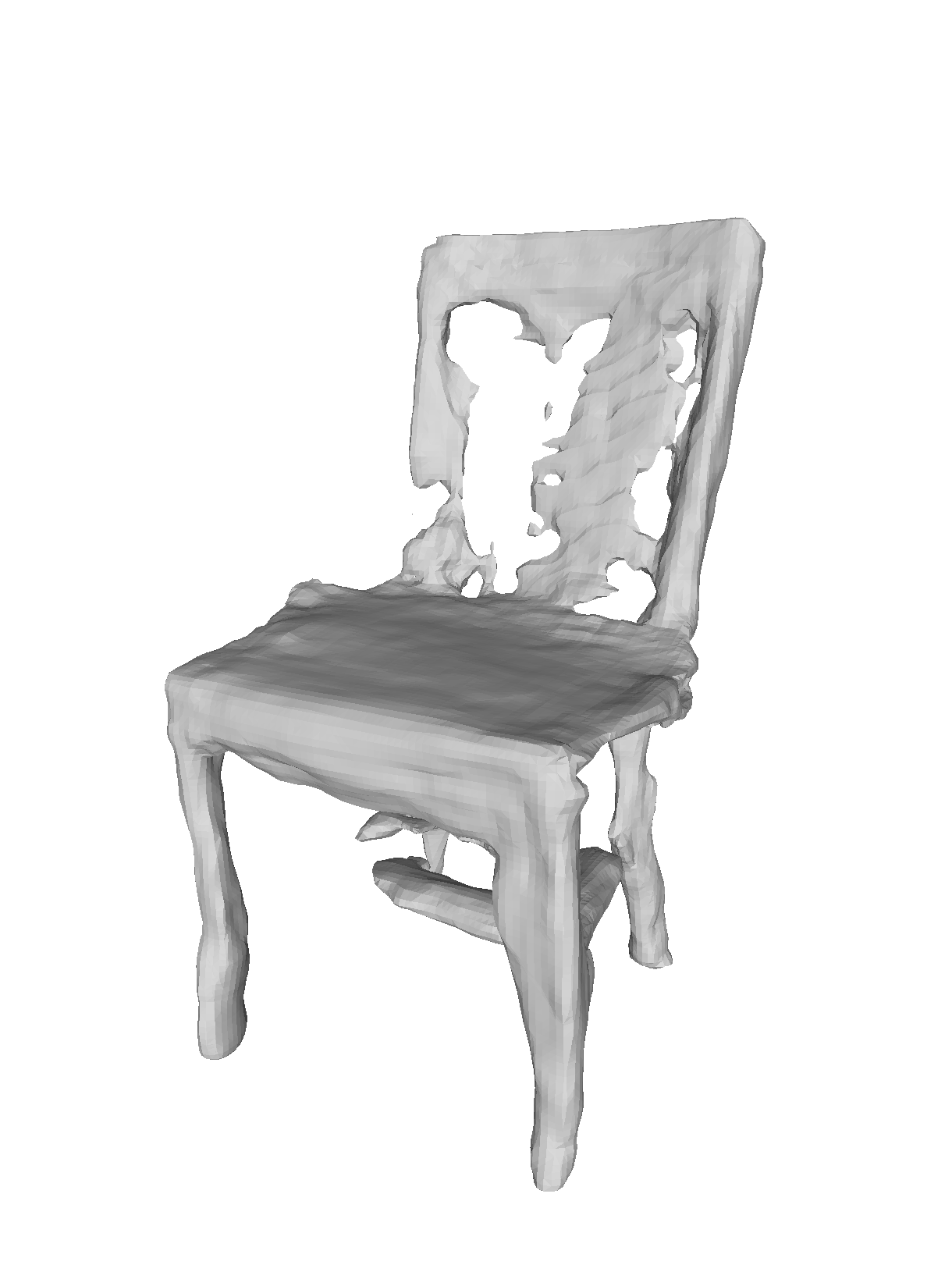} \\
		\insertb{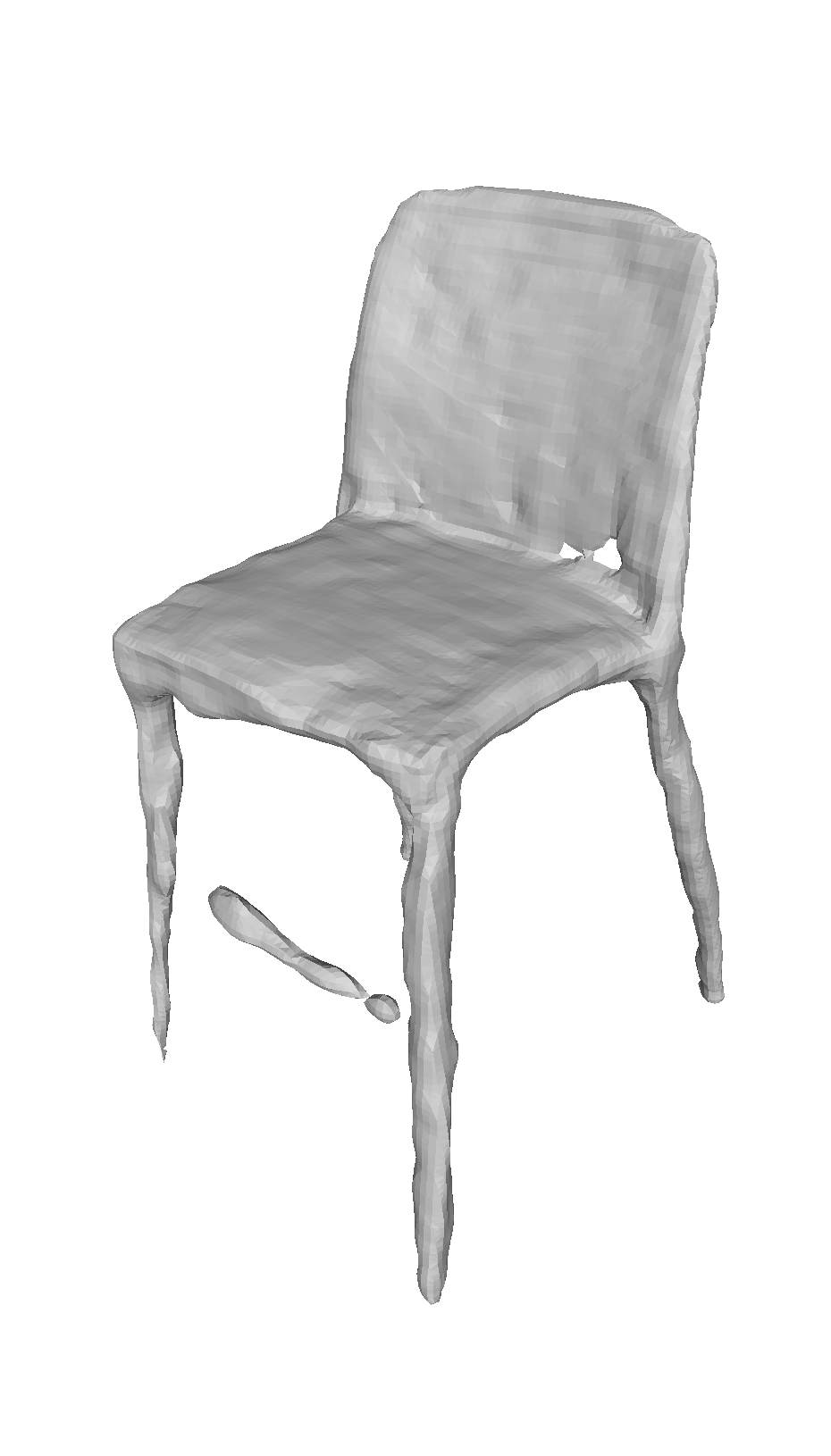} \\
		\insertc{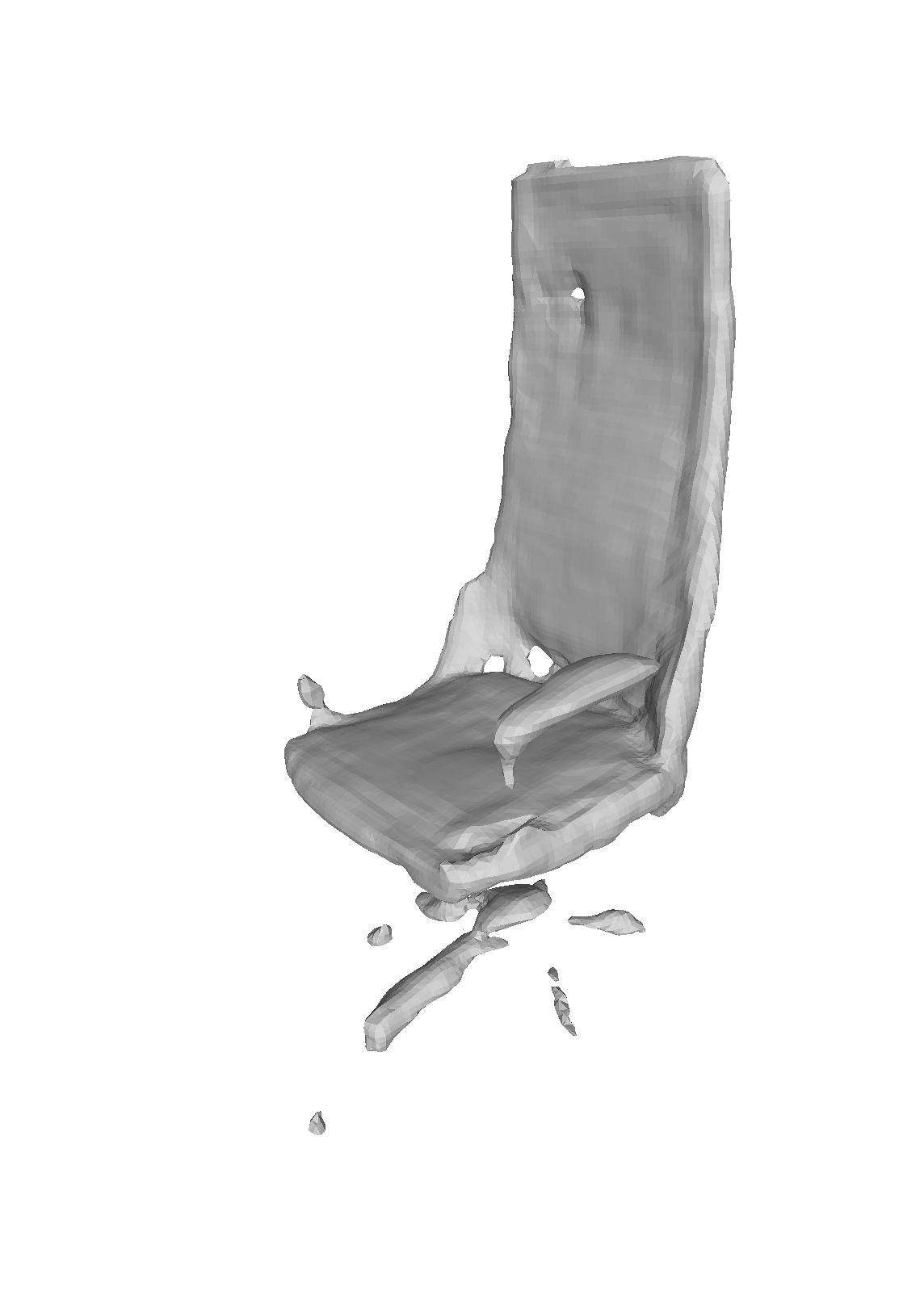} \\
		(4)
	\end{minipage}
	\begin{minipage}{0.135\linewidth}
		\centering
		\inserta{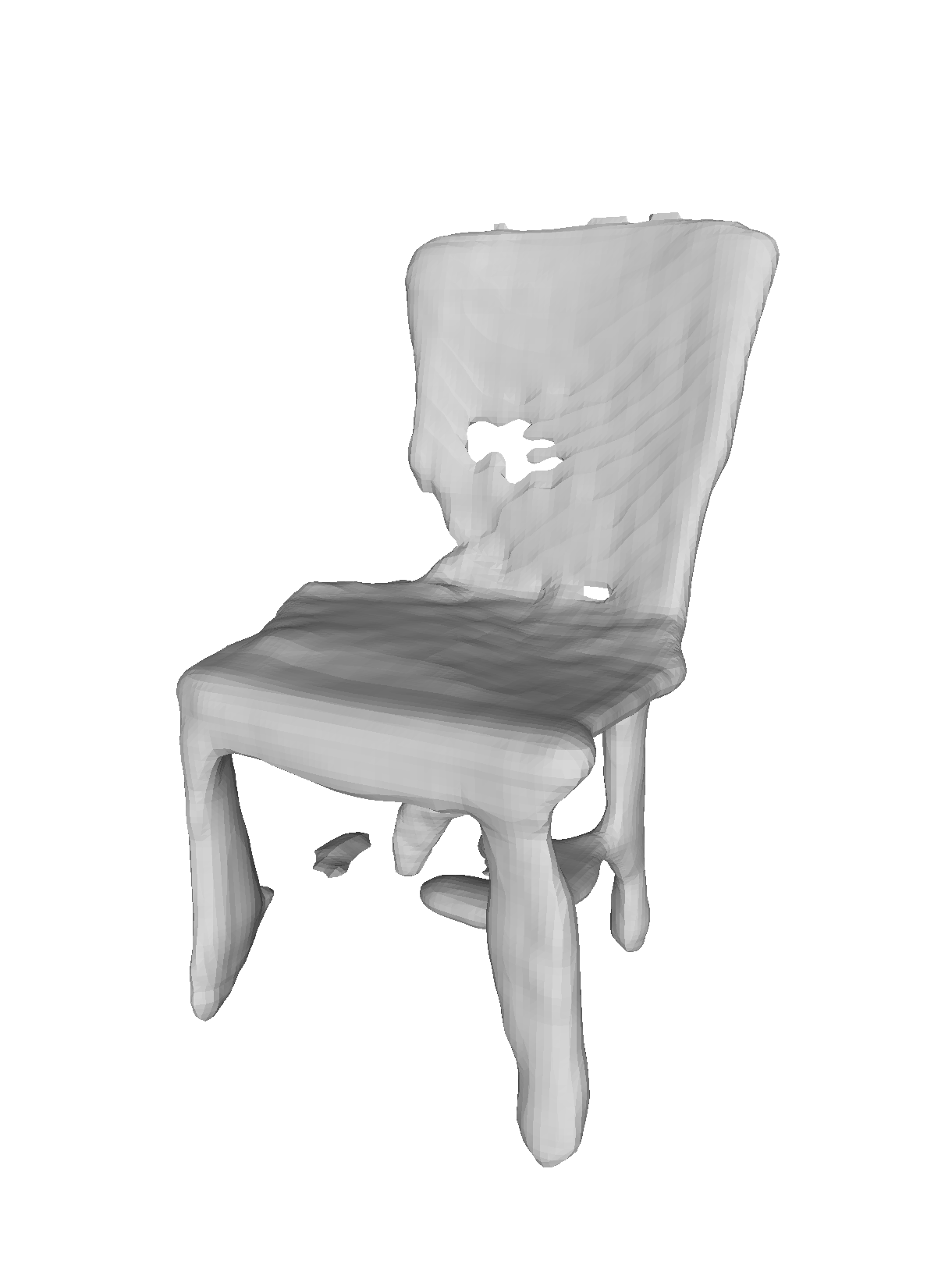} \\
		\insertb{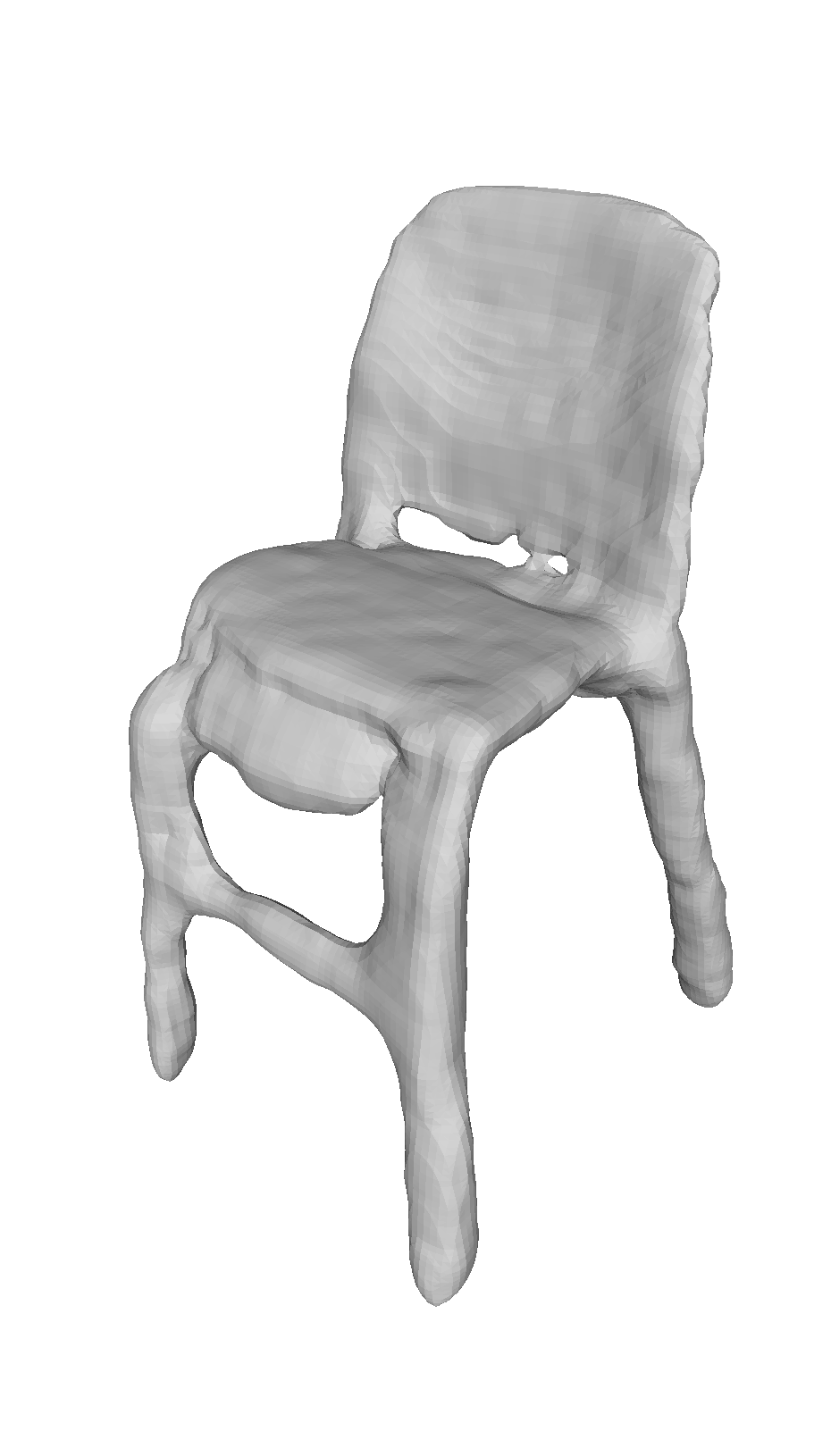} \\
		\insertc{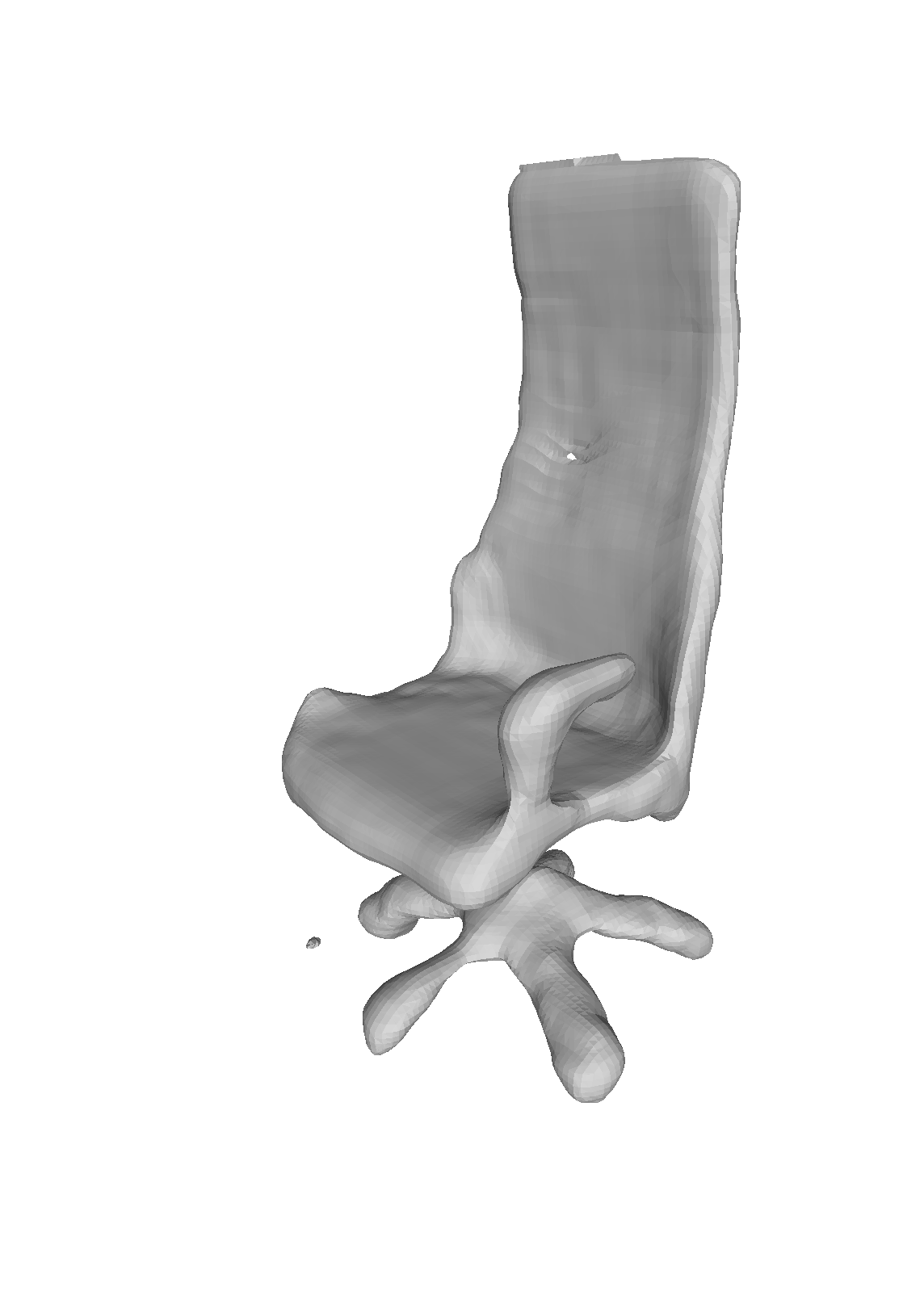} \\
		(5)
	\end{minipage}
	\begin{minipage}{0.135\linewidth}
		\centering
		\inserta{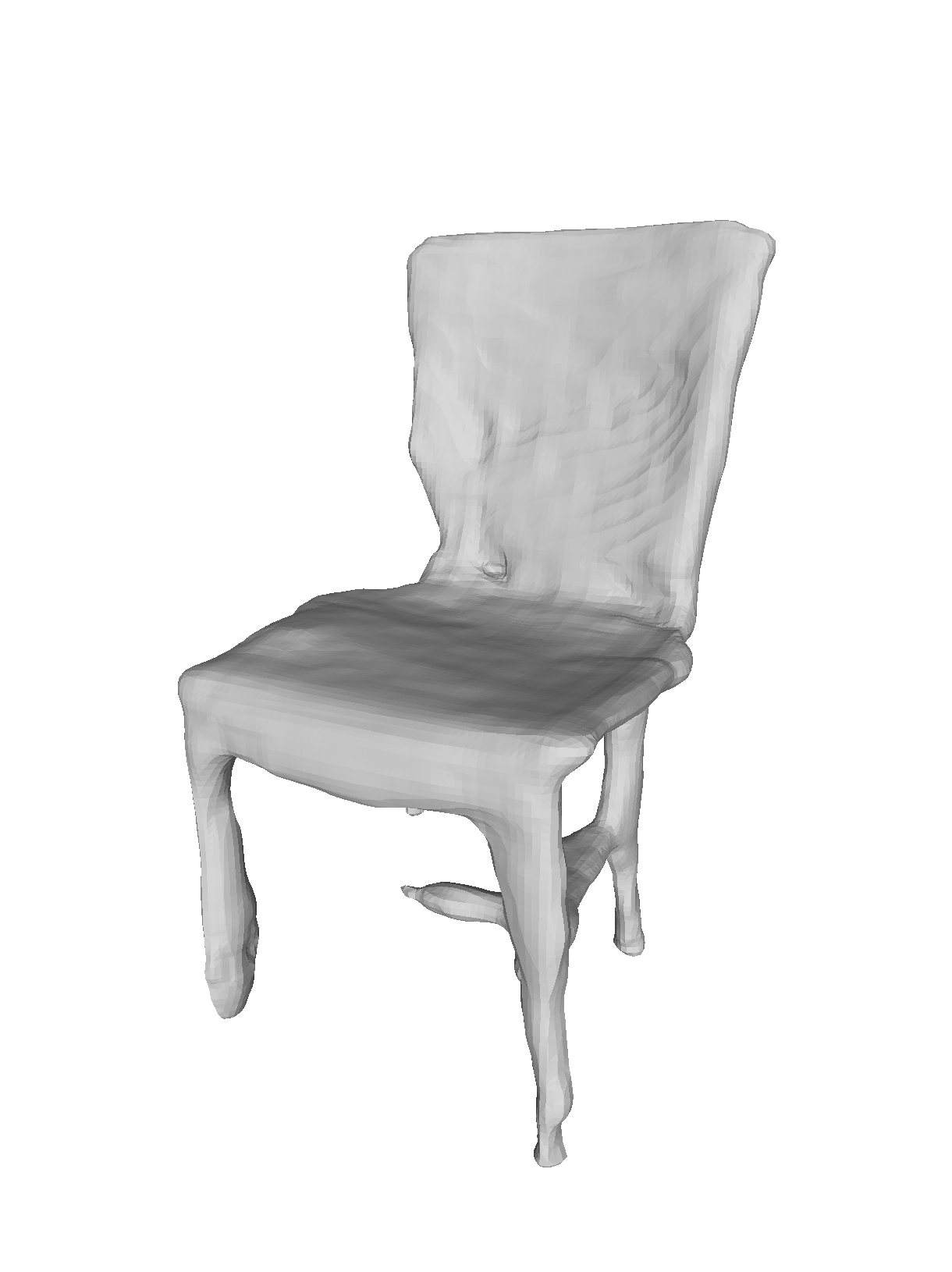} \\
		\insertb{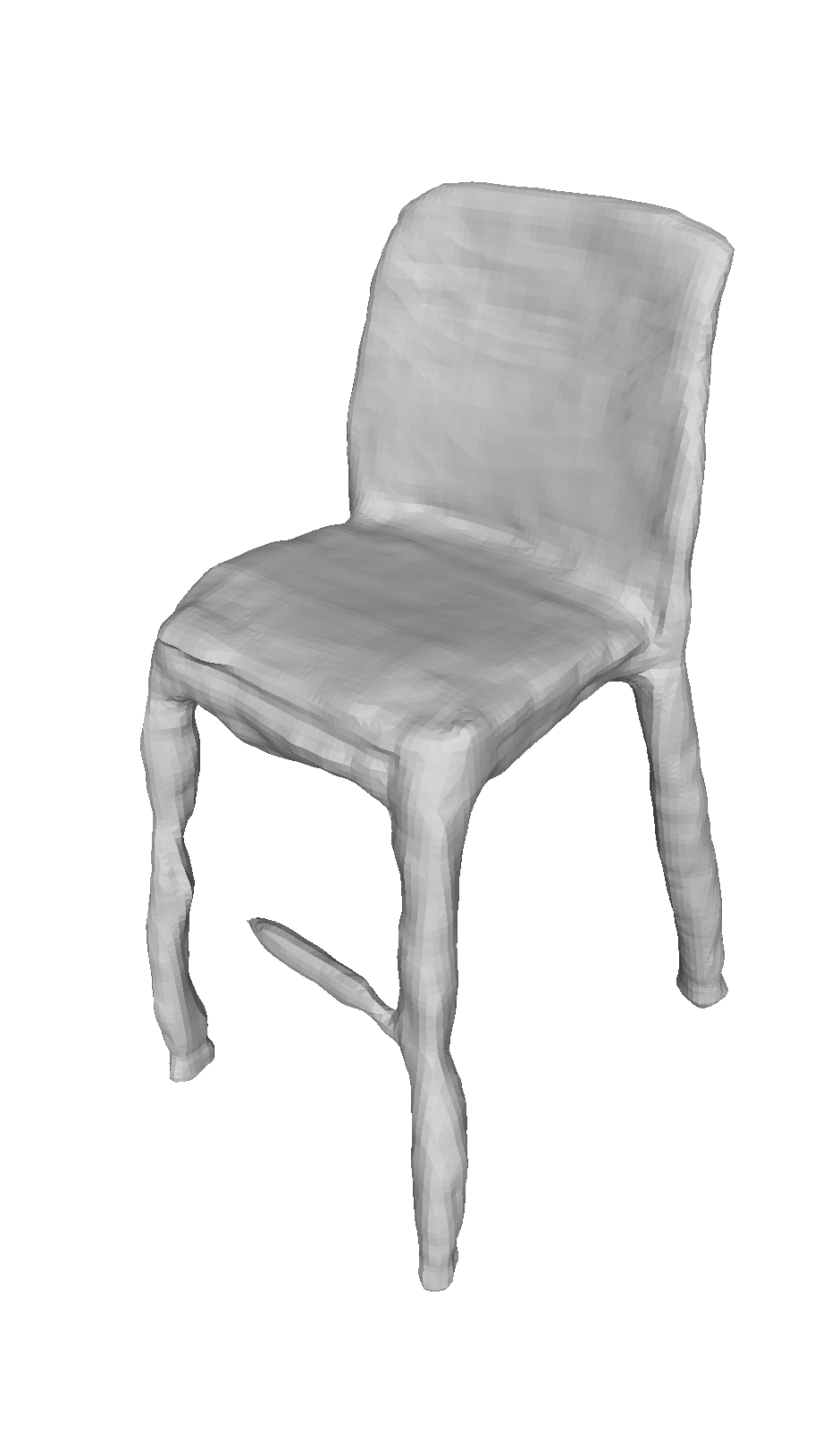} \\
		\insertc{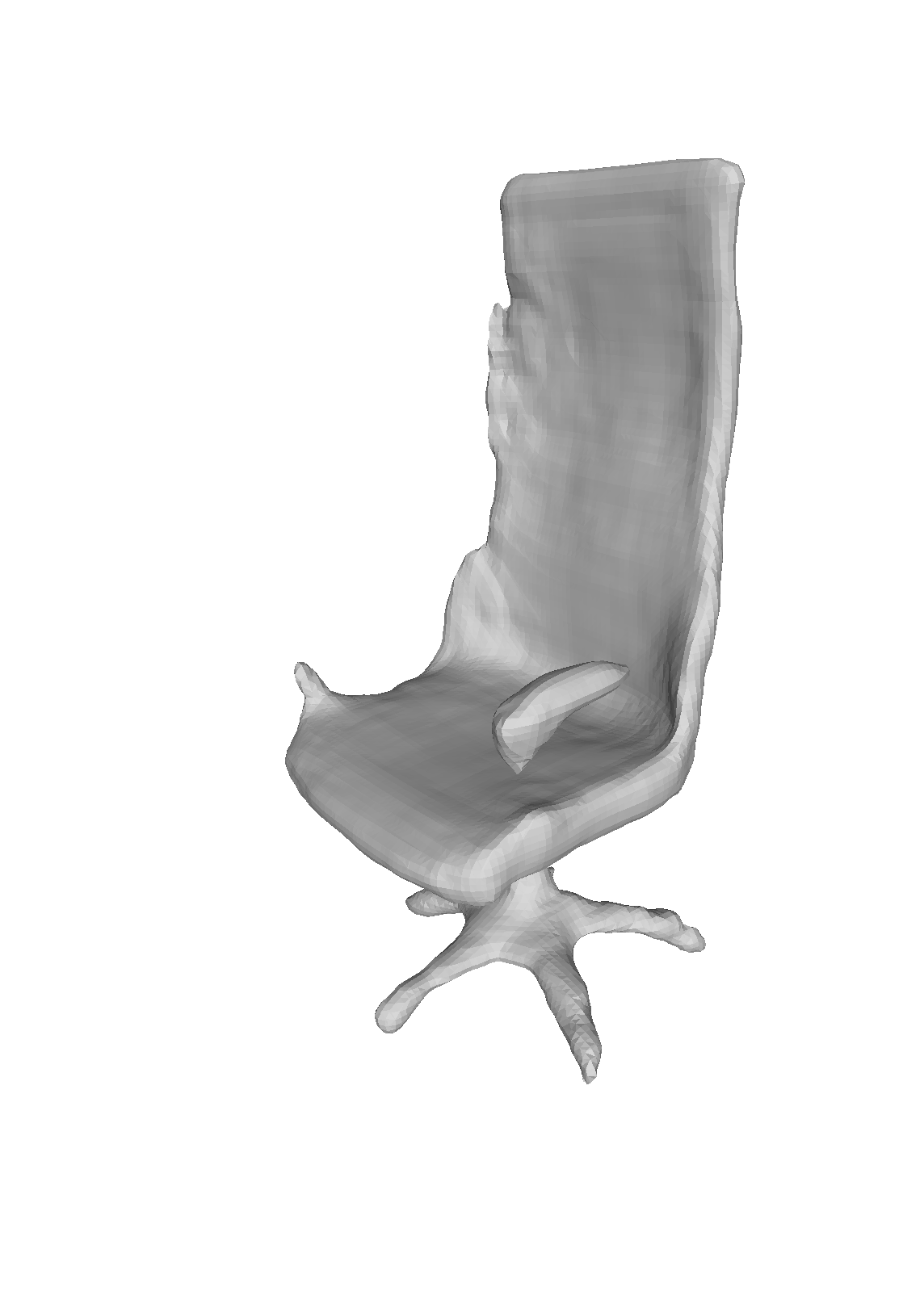} \\
		(6)
	\end{minipage}
	\begin{minipage}{0.135\linewidth}
		\centering
		\inserta{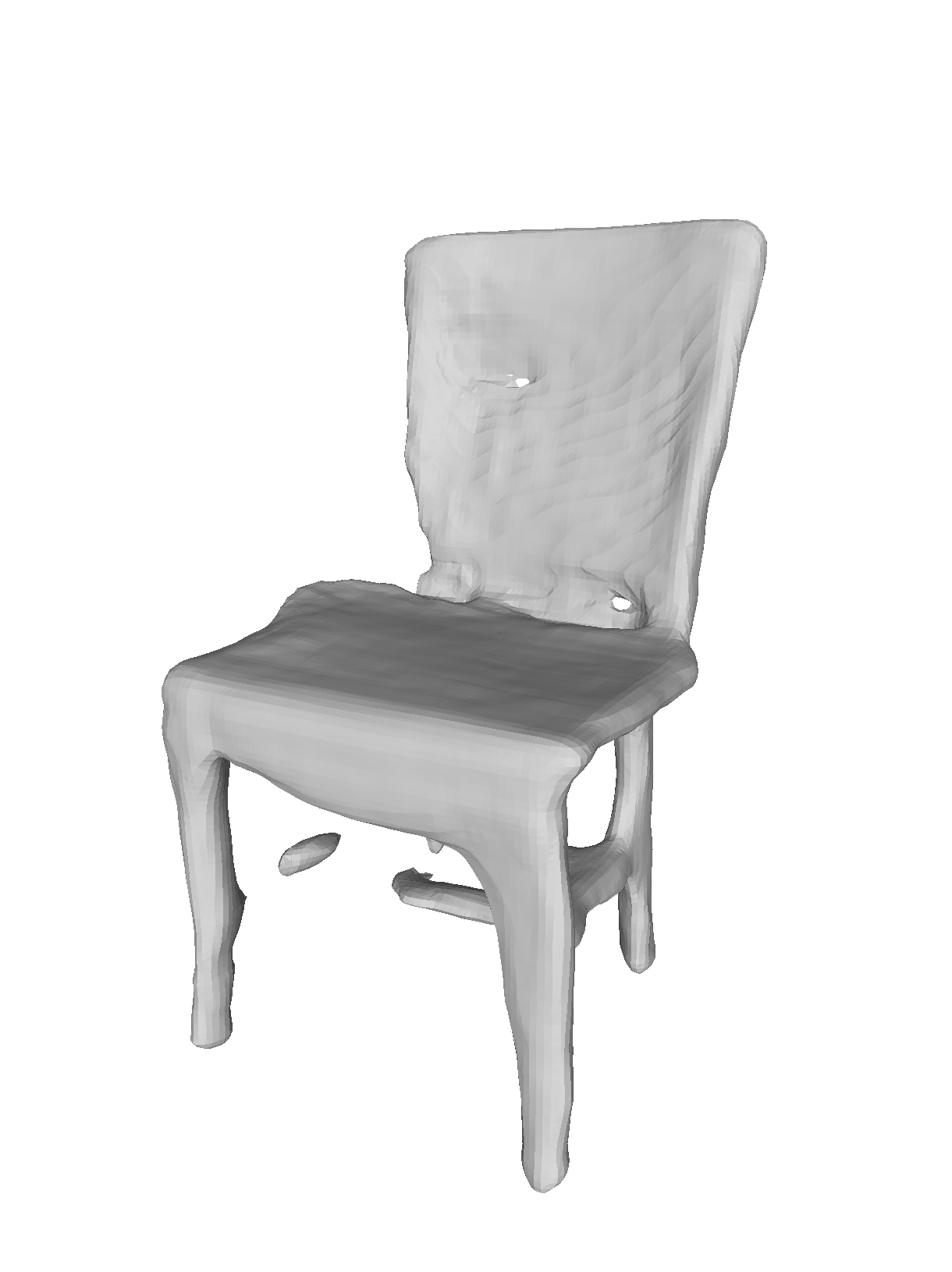} \\
		\insertb{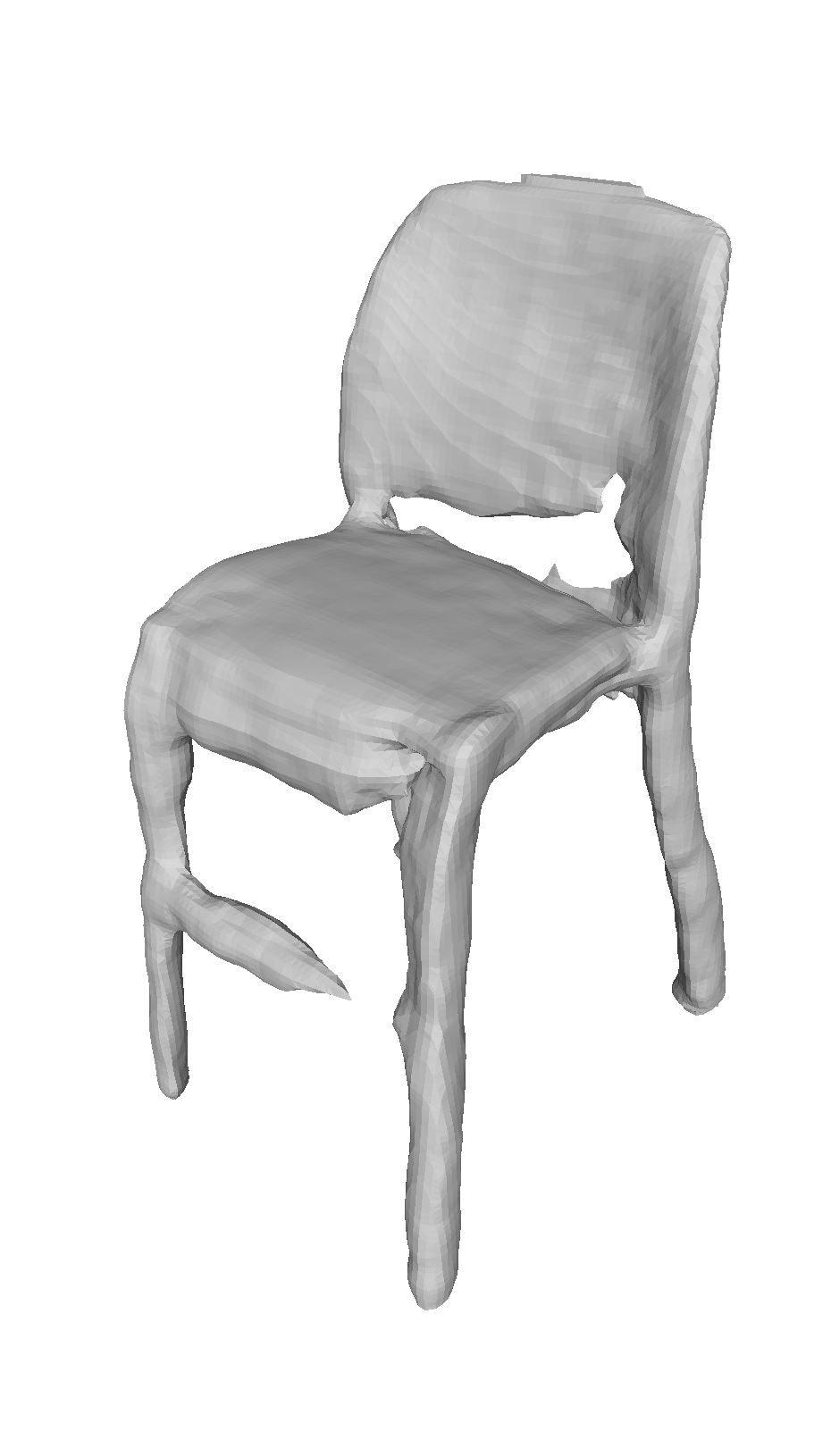} \\
		\insertc{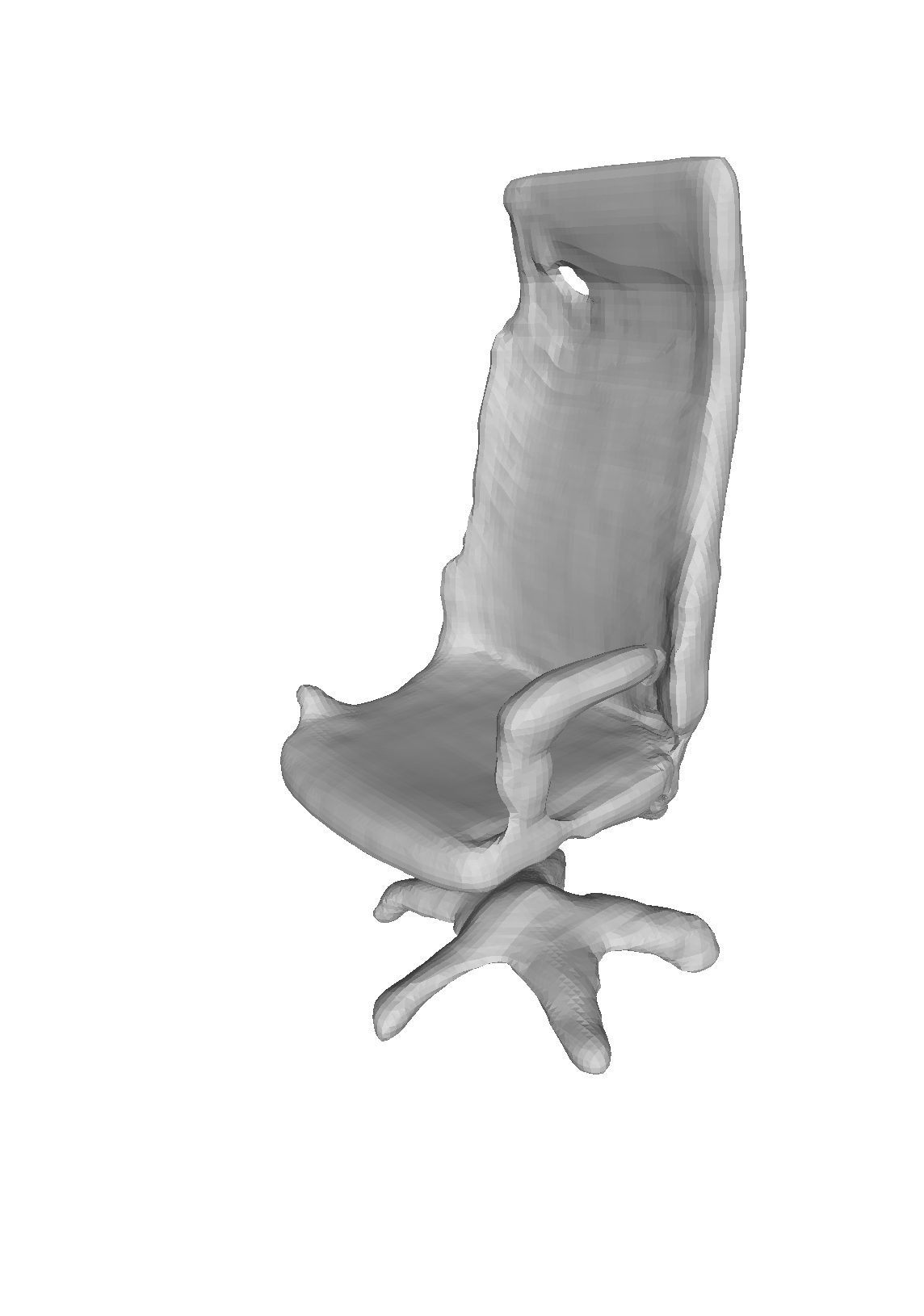} \\
		(7)
	\end{minipage}
    \caption{Visualization of the ablation experiments in Tab. 4 of the main paper. (1) Real Scan. (2) Ground Truth. \Frst{(3) Ours}. (4) $\mathbf{F} = \mathbf{F}_s$. (5) $\mathbf{F} = \mathbf{F}_t$. (6) Non-Adap. (7) Contrad.} 
	\label{fig:abltab4}
	\vspace{-2mm}
\end{figure*}

\begin{figure*}[htbp]
	\centering
	\begin{minipage}{0.105\linewidth}
		\centering
		\insertd{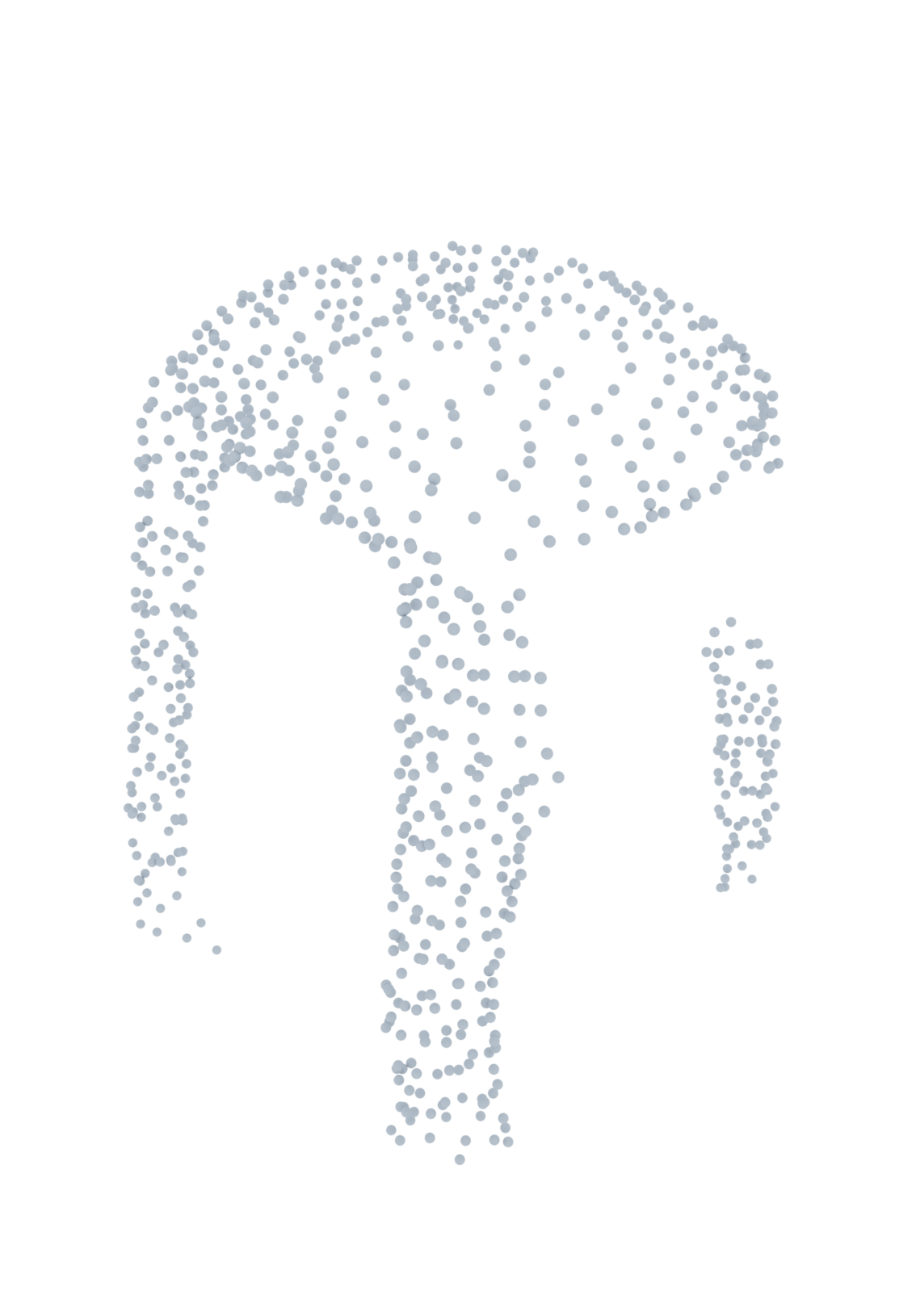} \\
		\inserte{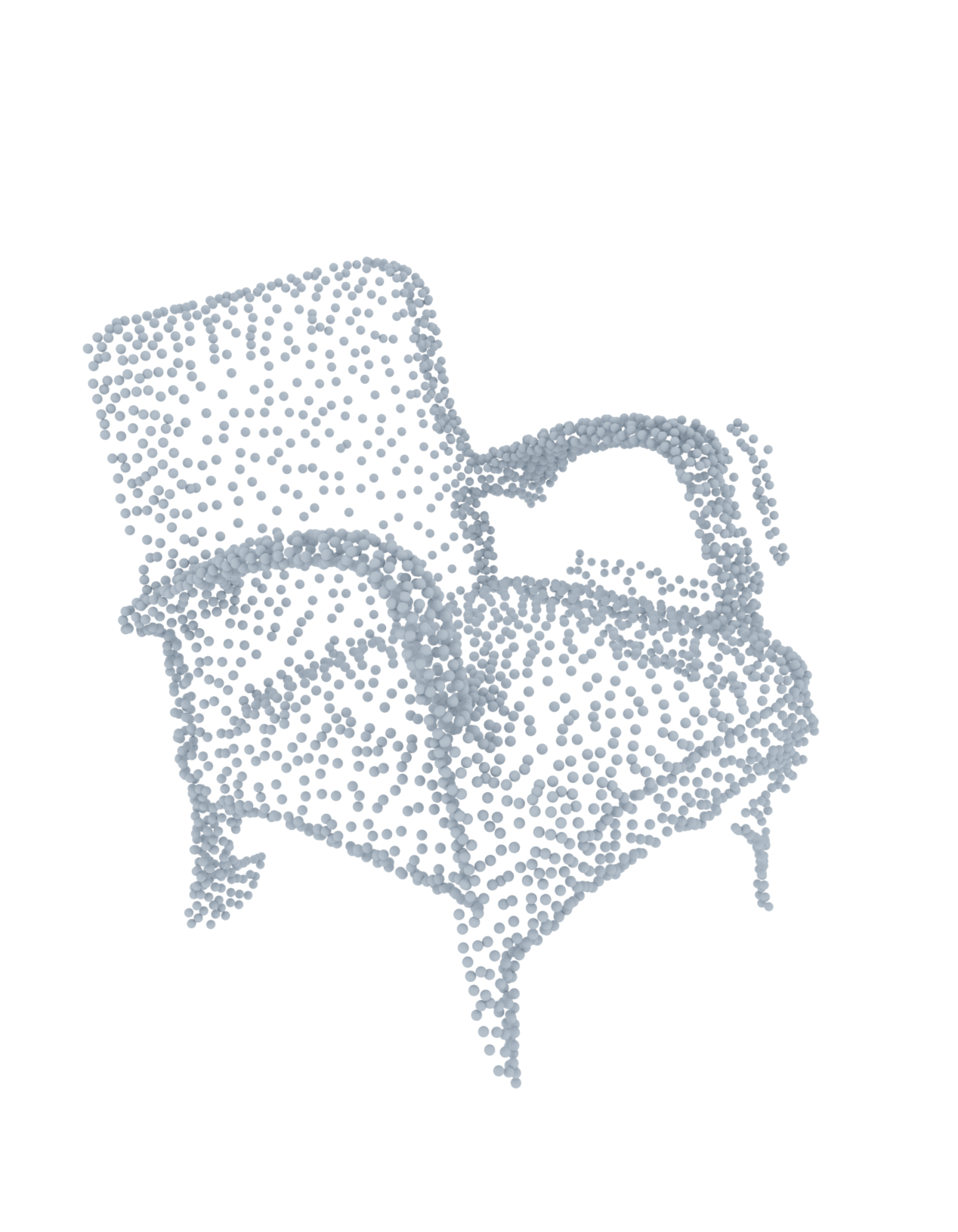} \\
		\insertf{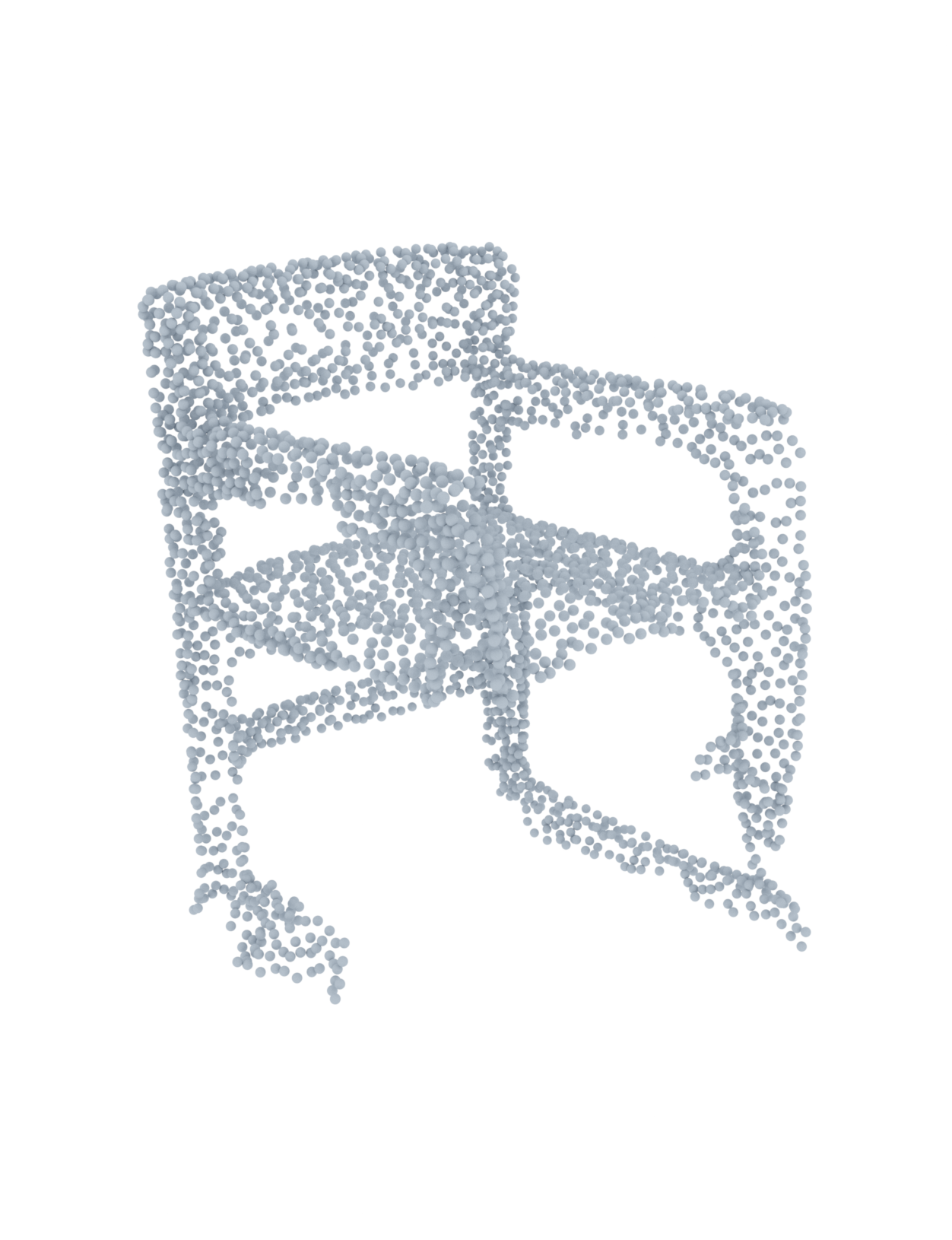} \\
		(1)
	\end{minipage}
	\begin{minipage}{0.105\linewidth}
		\centering
		\insertd{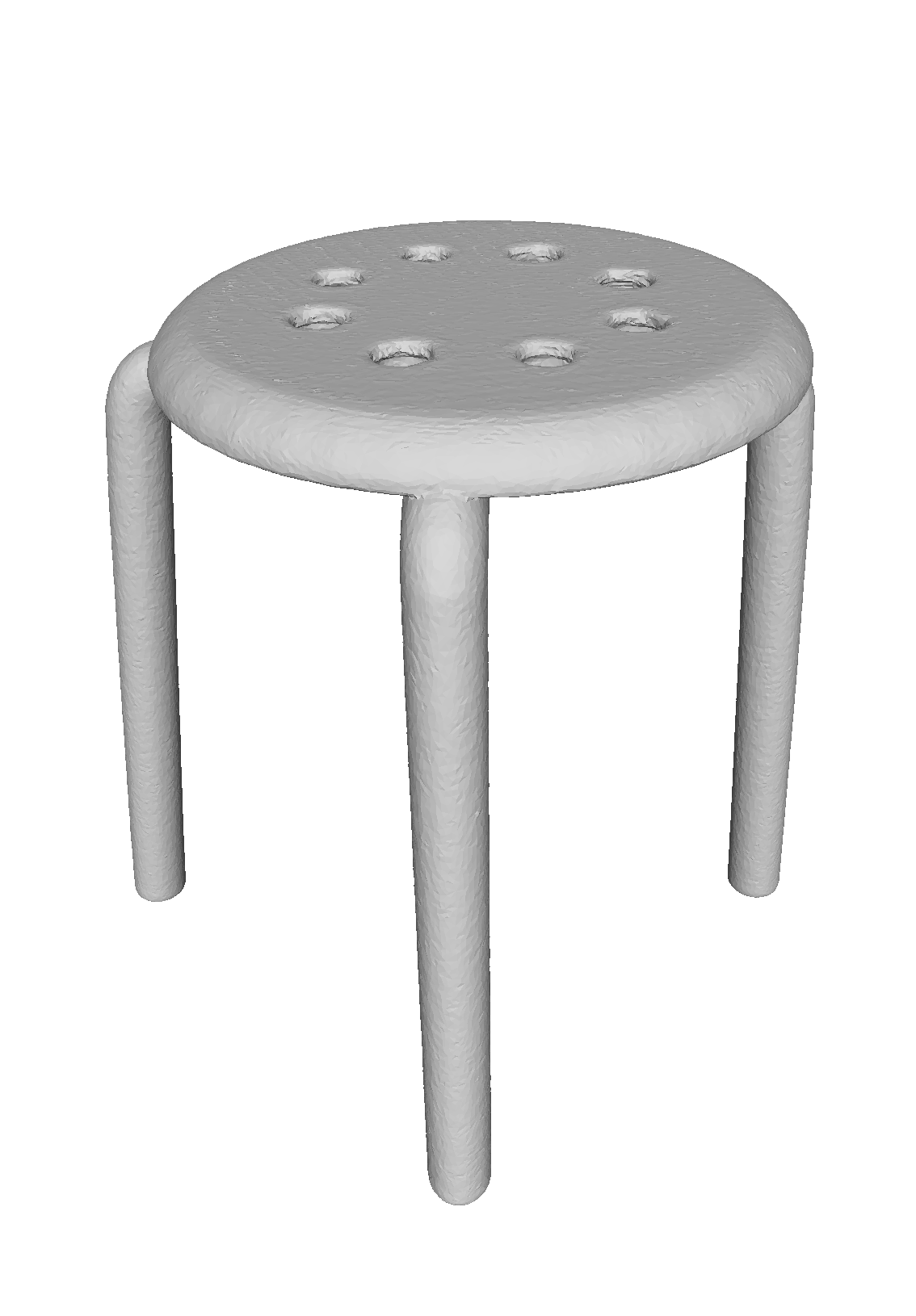} \\
		\inserte{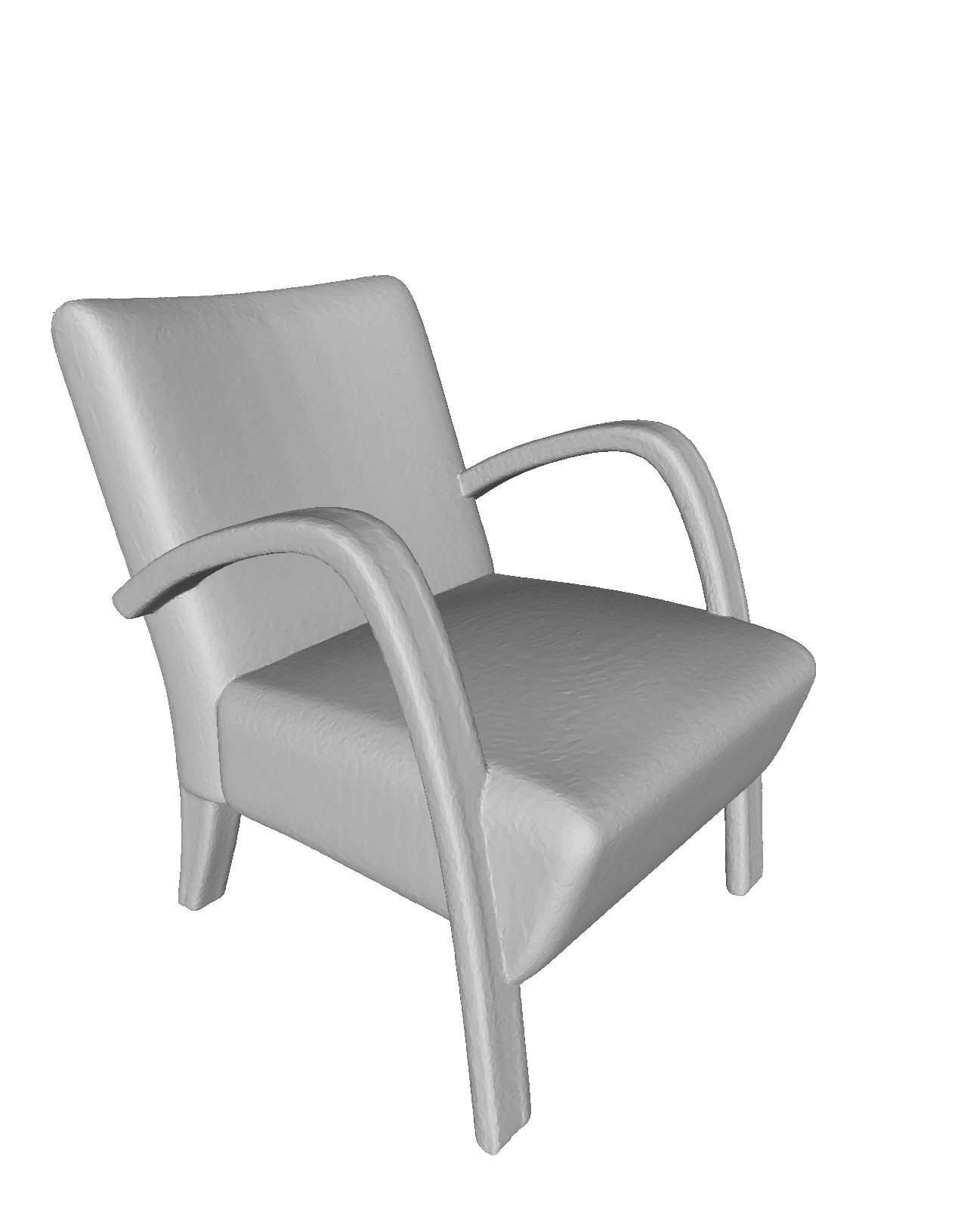} \\
		\insertf{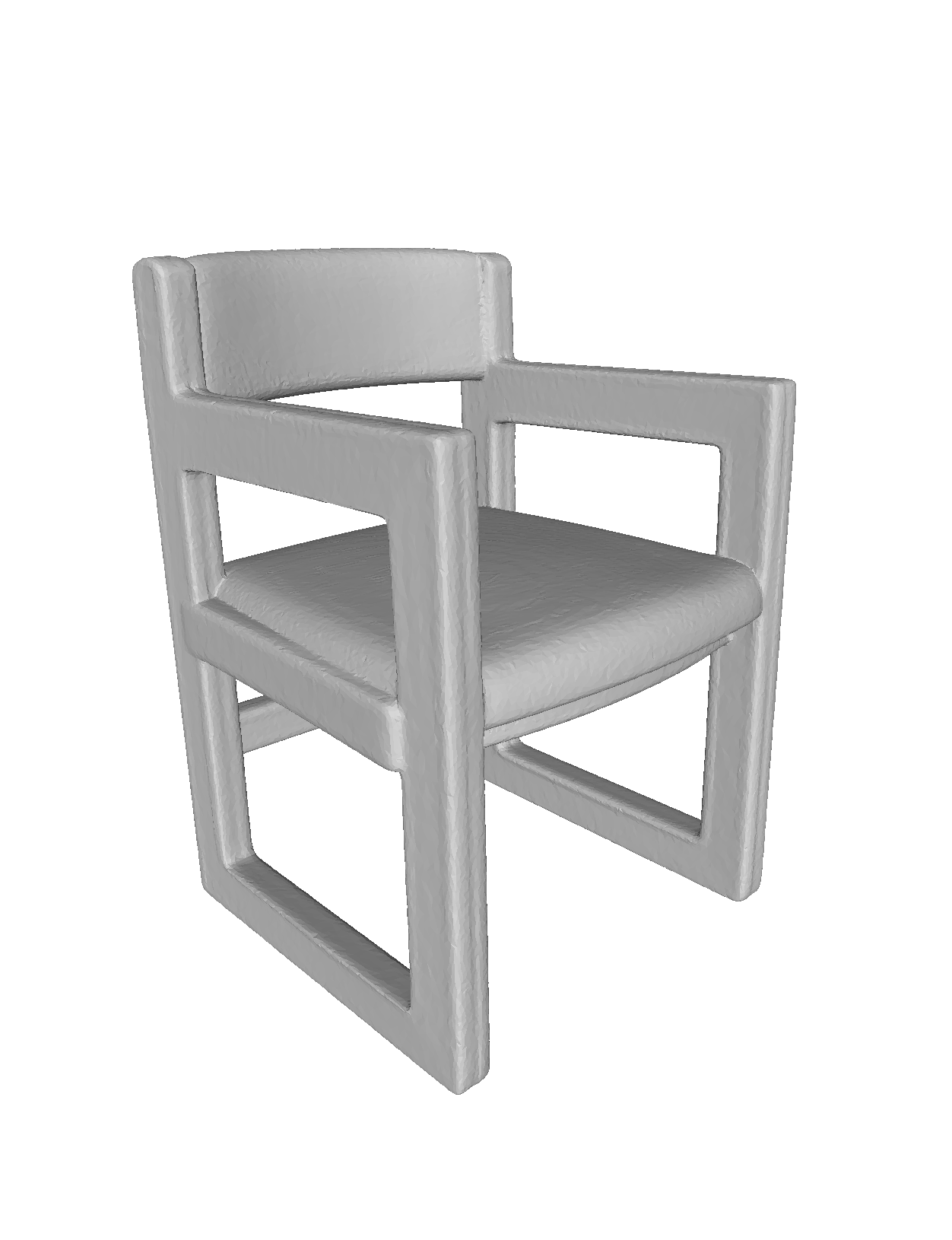} \\
		(2)
	\end{minipage}
    \begin{minipage}{0.105\linewidth}
		\centering
		\insertd{figures/fig_abl5/6_Random-Surface(Ours)_scene0091_00_20.png} \\
		\inserte{figures/fig_abl5/6_Random-Surface(Ours)_scene0496_00_9.png} \\
		\insertf{figures/fig_abl5/6_Random-Surface(Ours)_scene0665_00_4.png} \\
		\Frst{(3)}
	\end{minipage}
	\begin{minipage}{0.105\linewidth}
		\centering
		\insertd{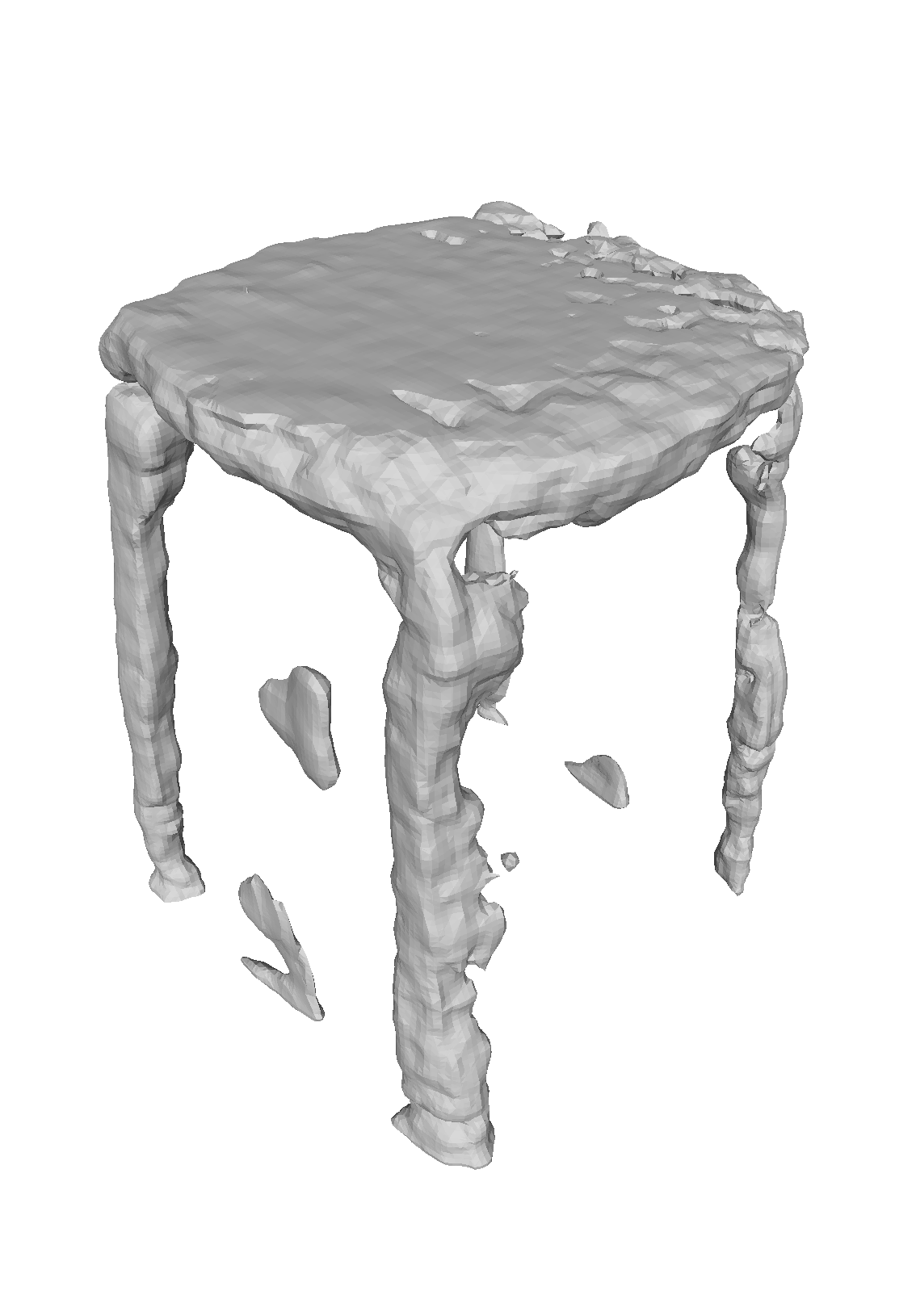} \\
		\inserte{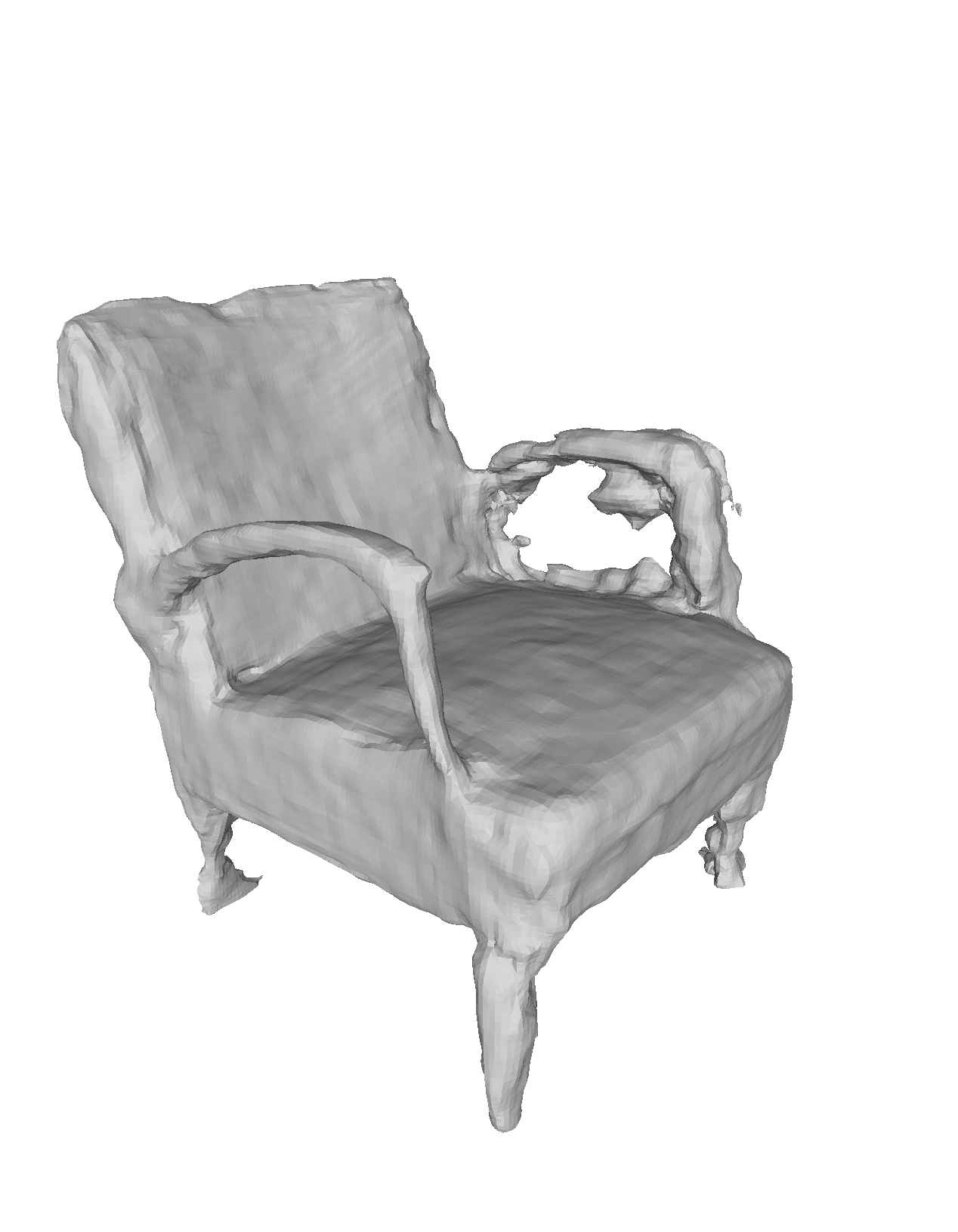} \\
		\insertf{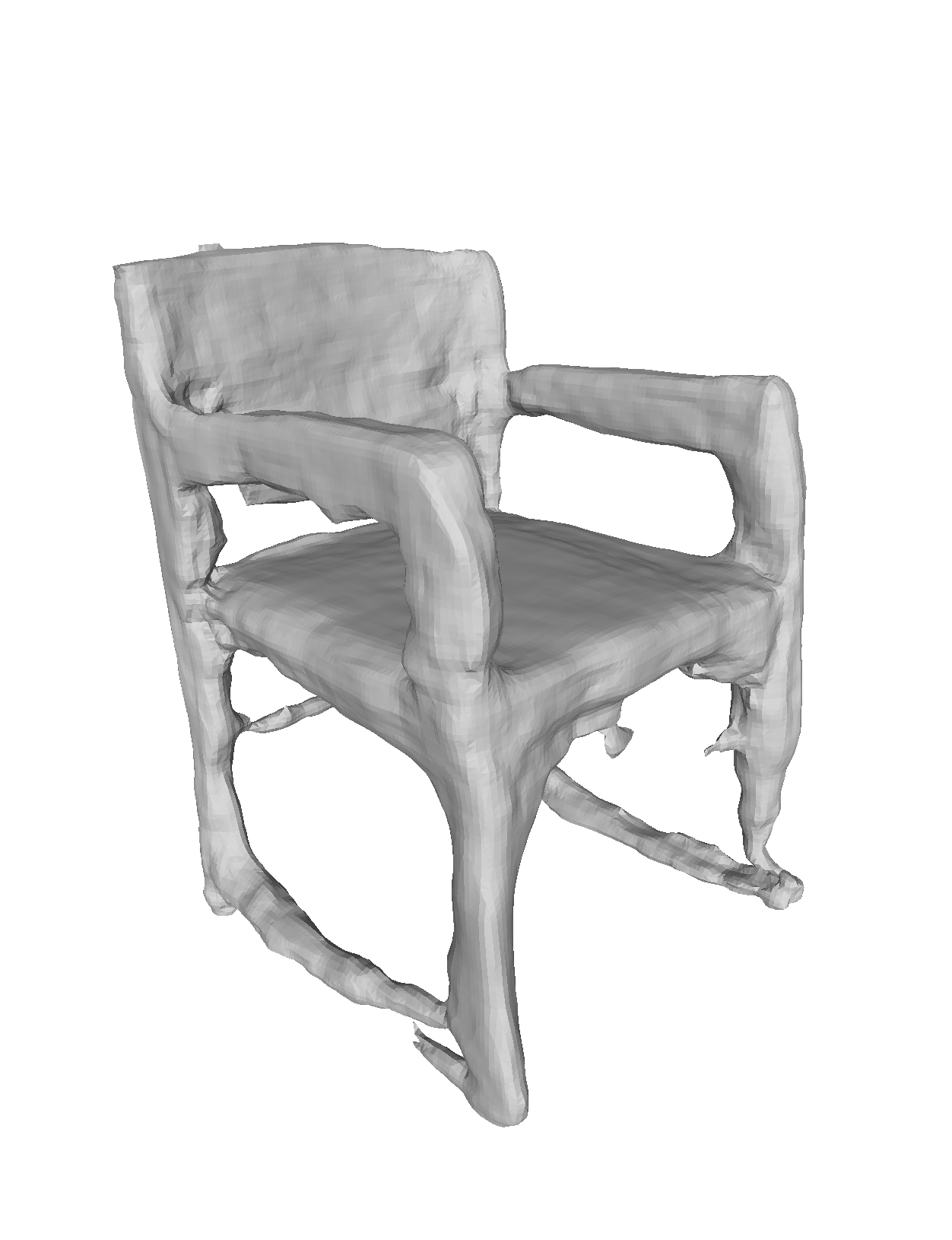} \\
		(4)
	\end{minipage}
	\begin{minipage}{0.105\linewidth}
		\centering
		\insertd{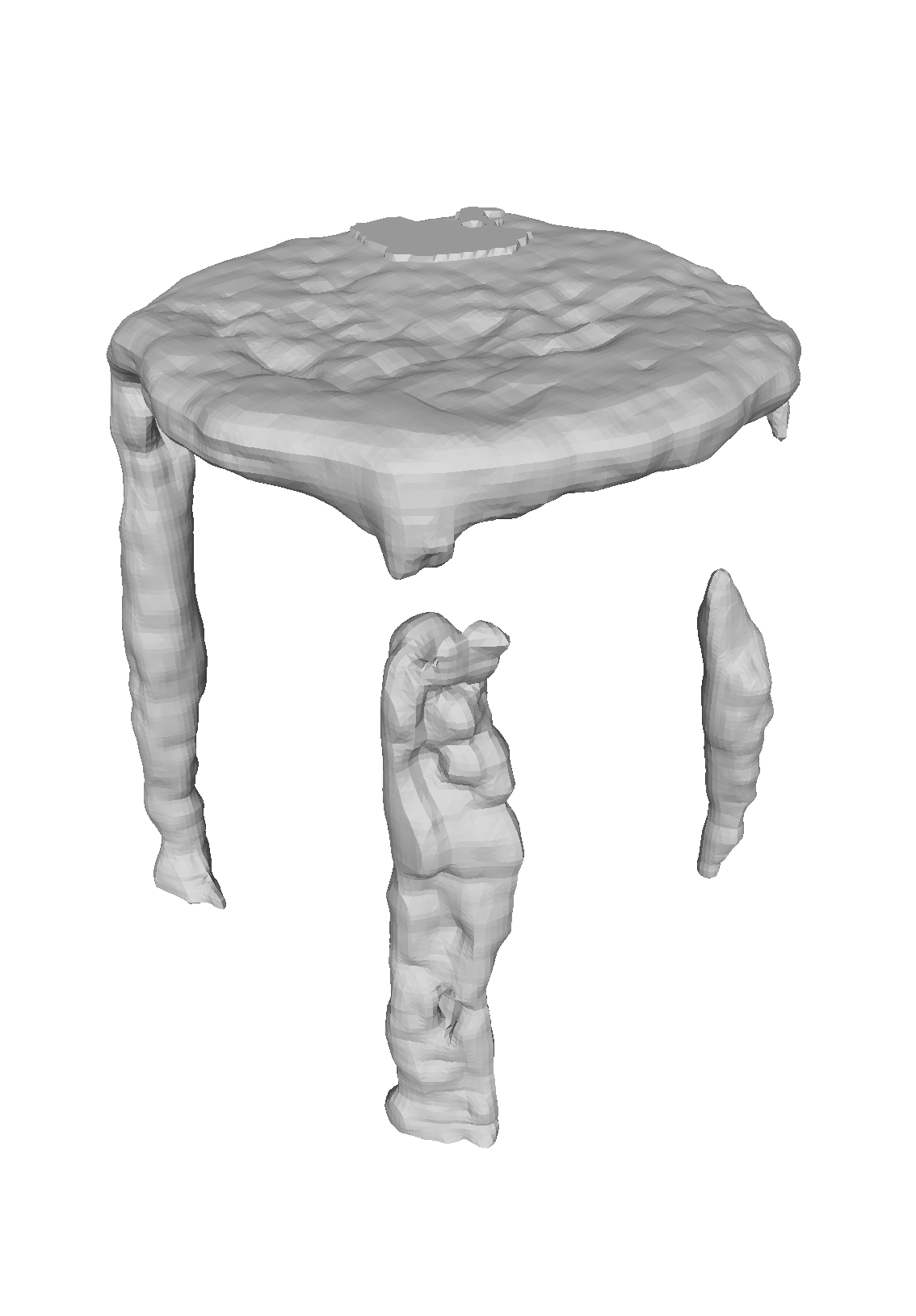} \\
		\inserte{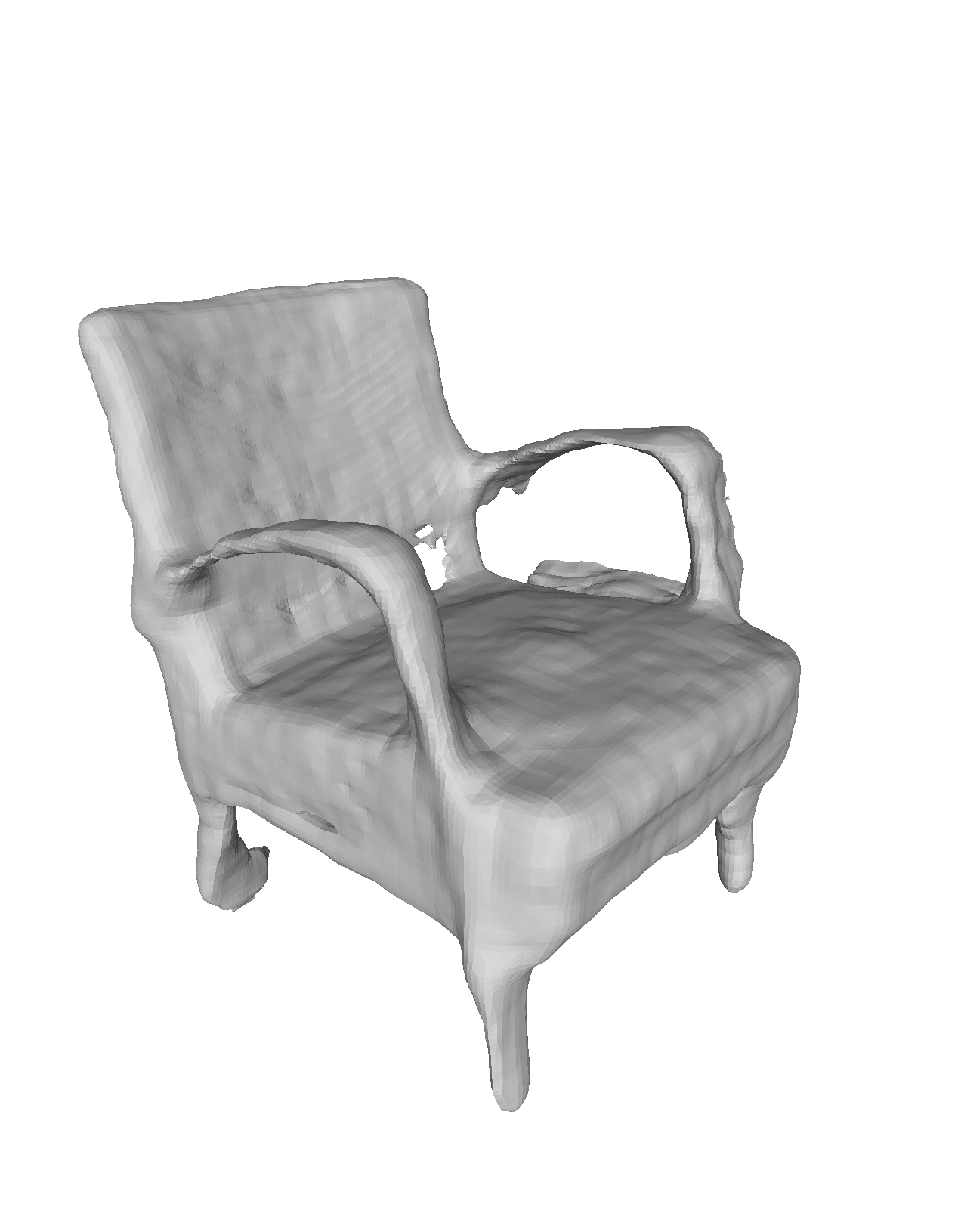} \\
		\insertf{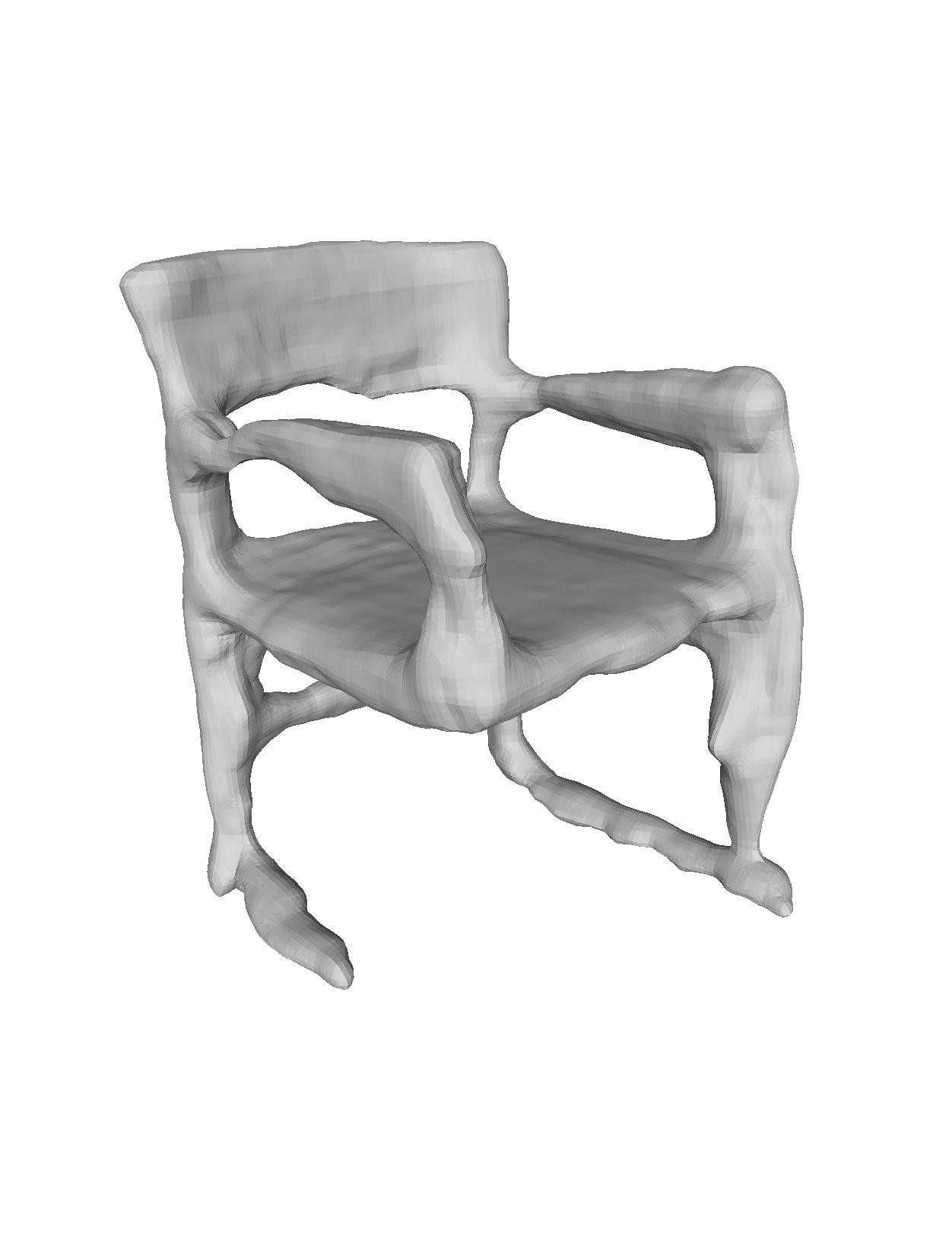} \\
		(5)
	\end{minipage}
	\begin{minipage}{0.105\linewidth}
		\centering
		\insertd{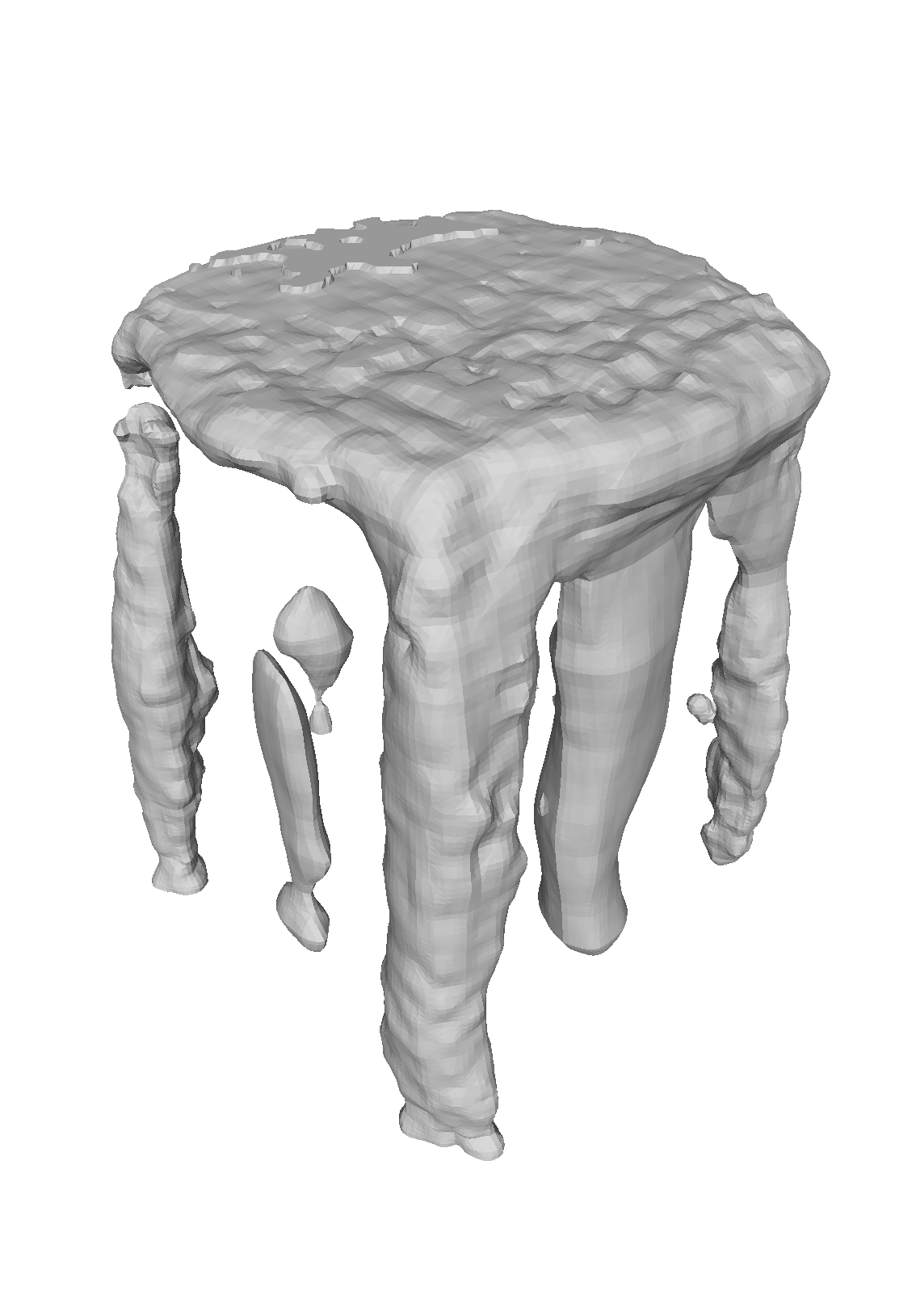} \\
		\inserte{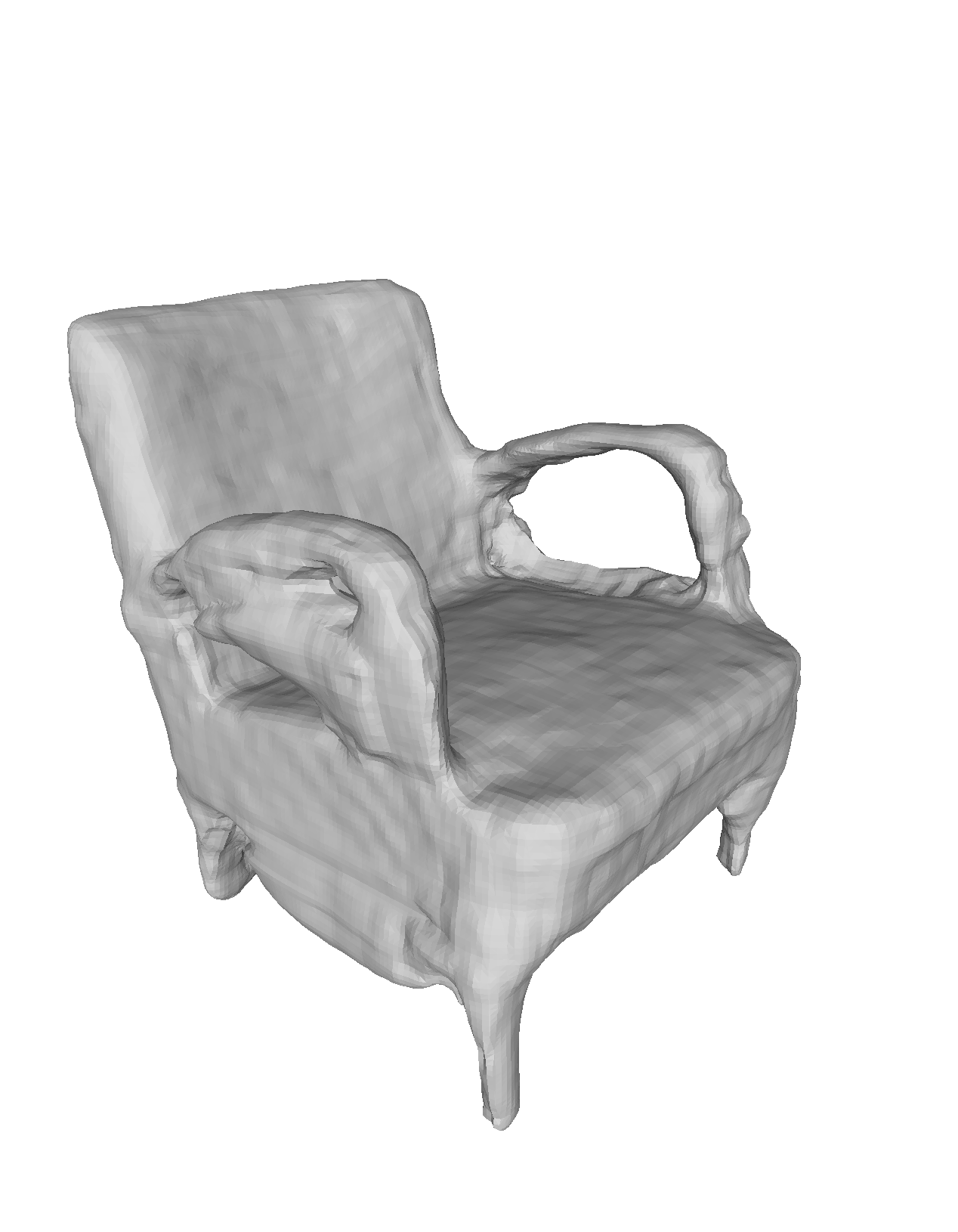} \\
		\insertf{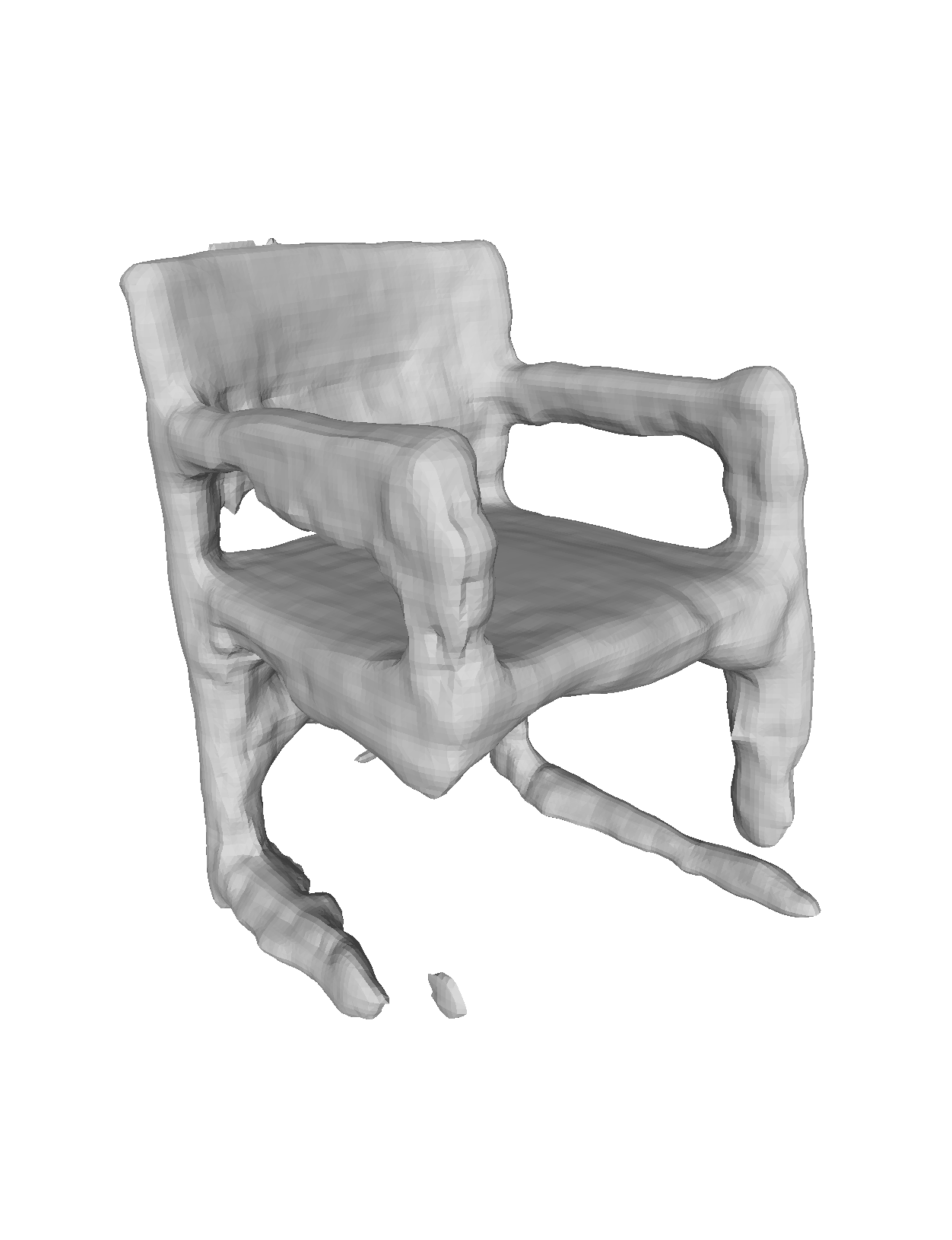} \\
		(6)
	\end{minipage}
	\begin{minipage}{0.105\linewidth}
		\centering
		\insertd{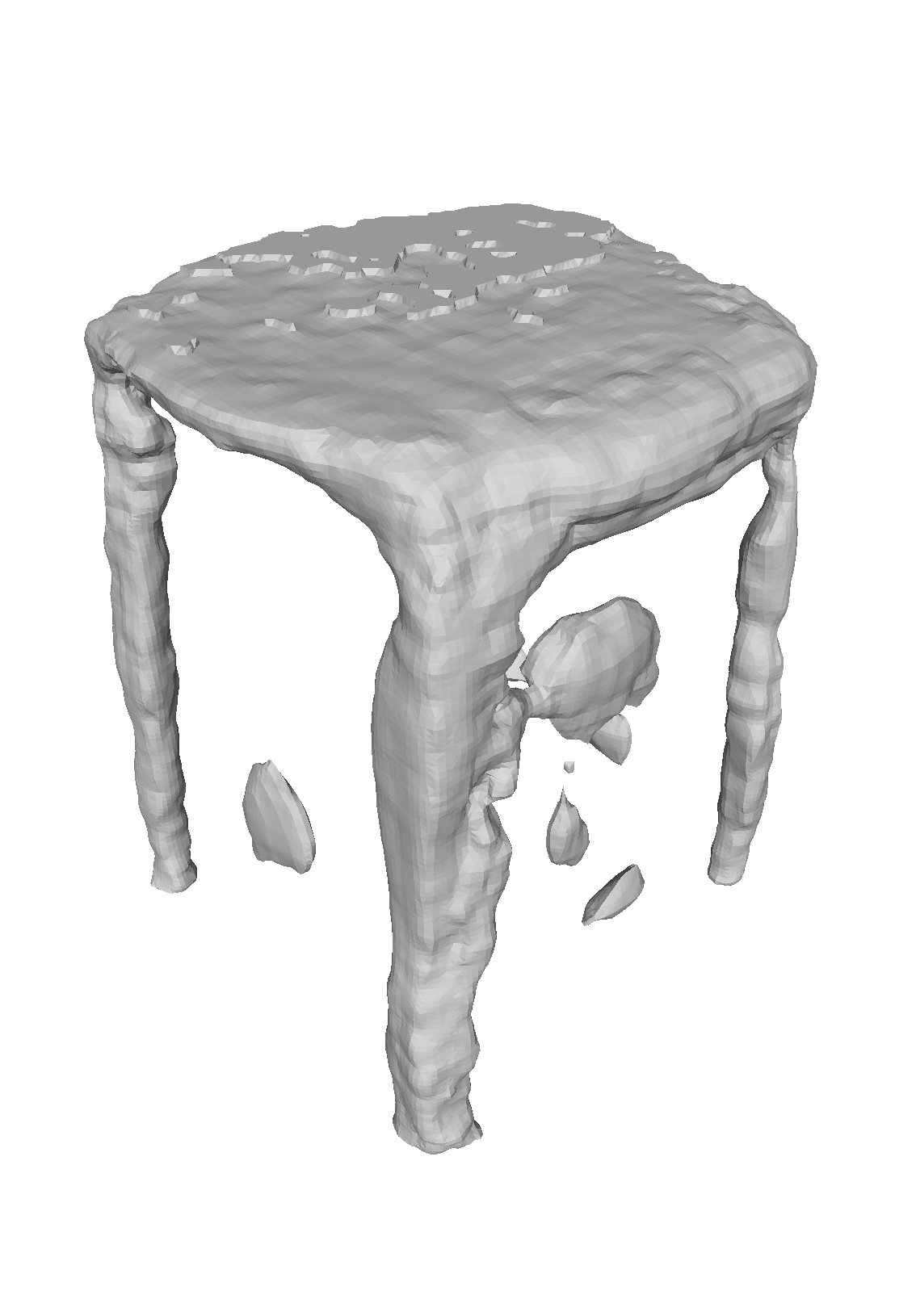} \\
		\inserte{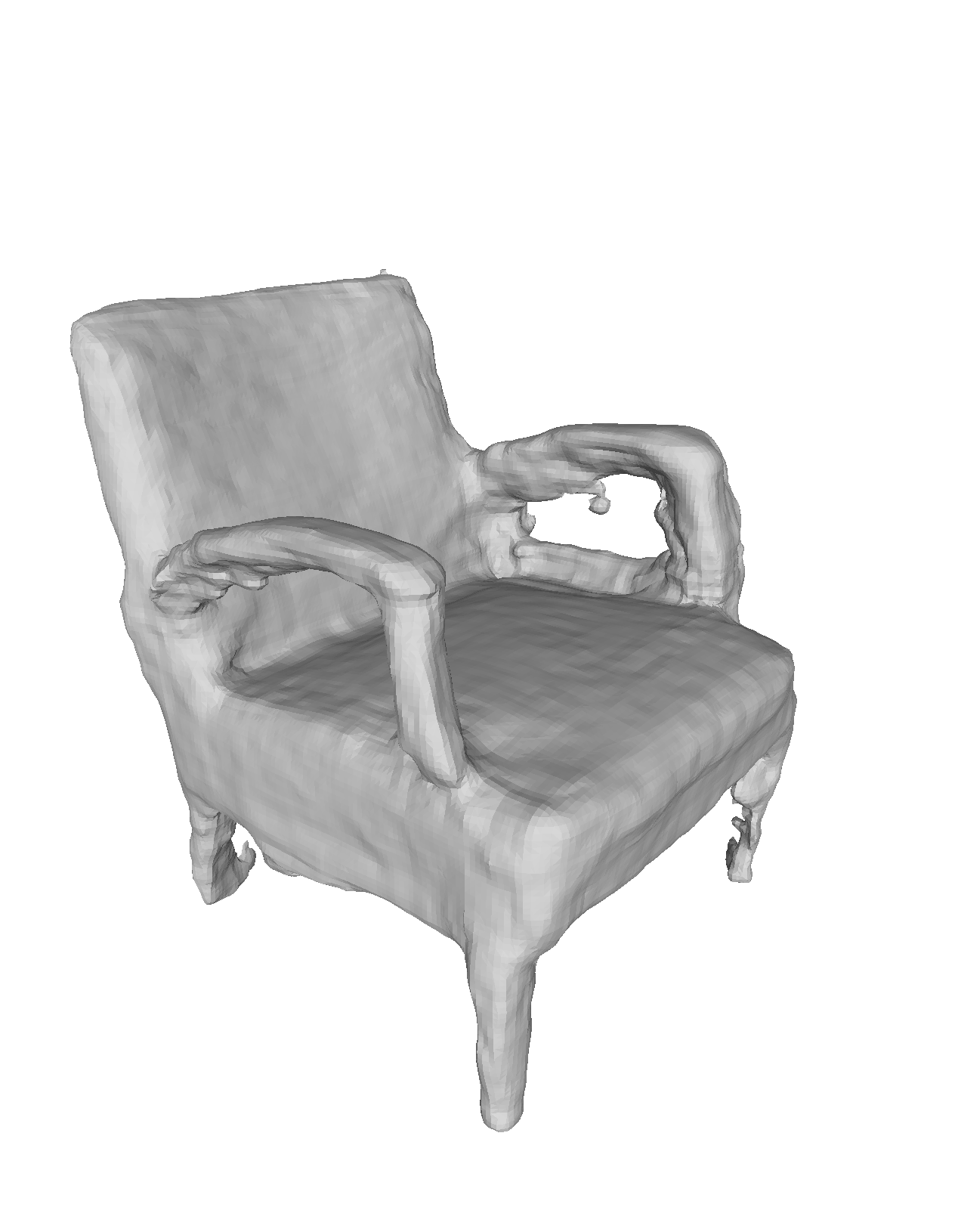} \\
		\insertf{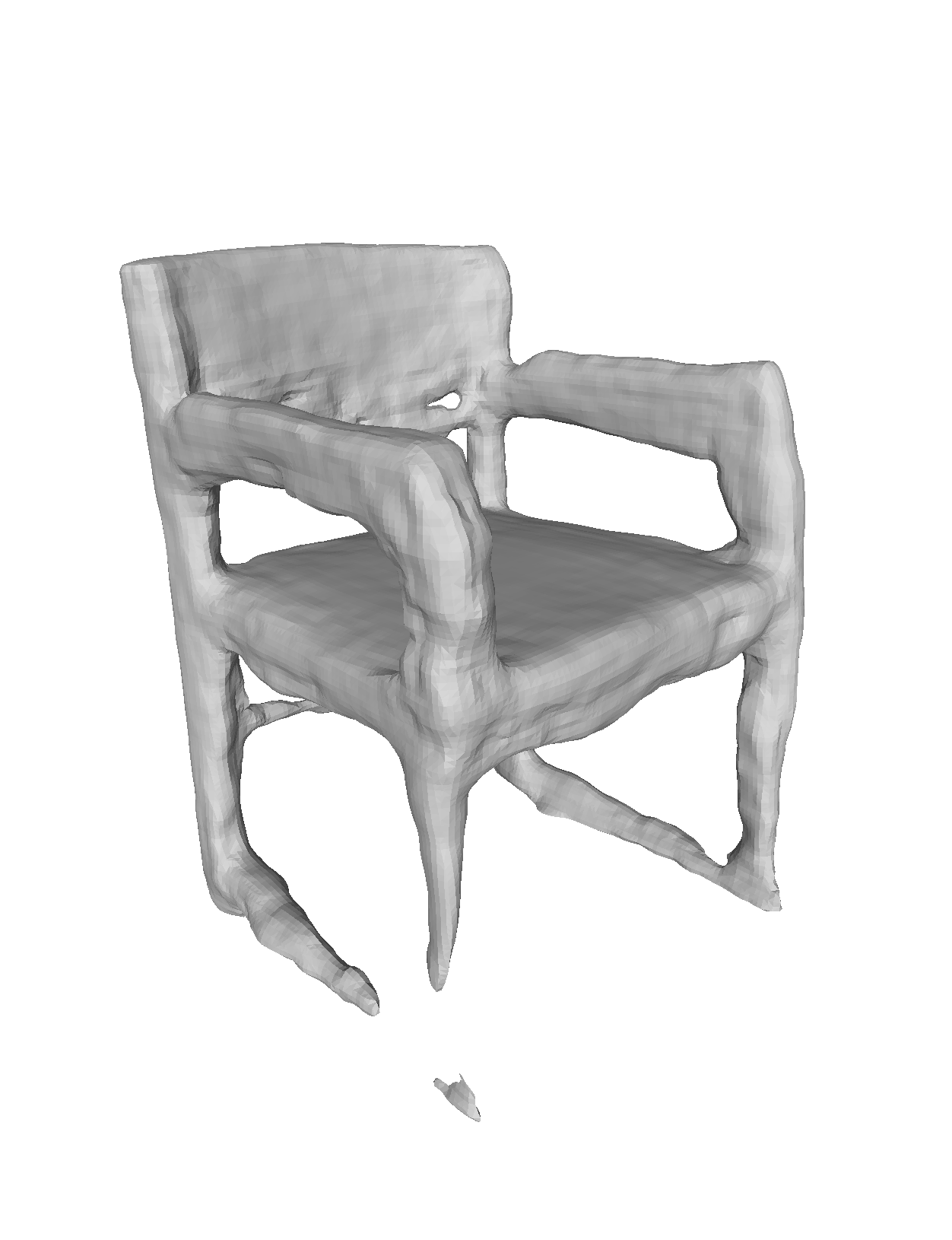} \\
		(7)
	\end{minipage}
	\begin{minipage}{0.105\linewidth}
		\centering
		\insertd{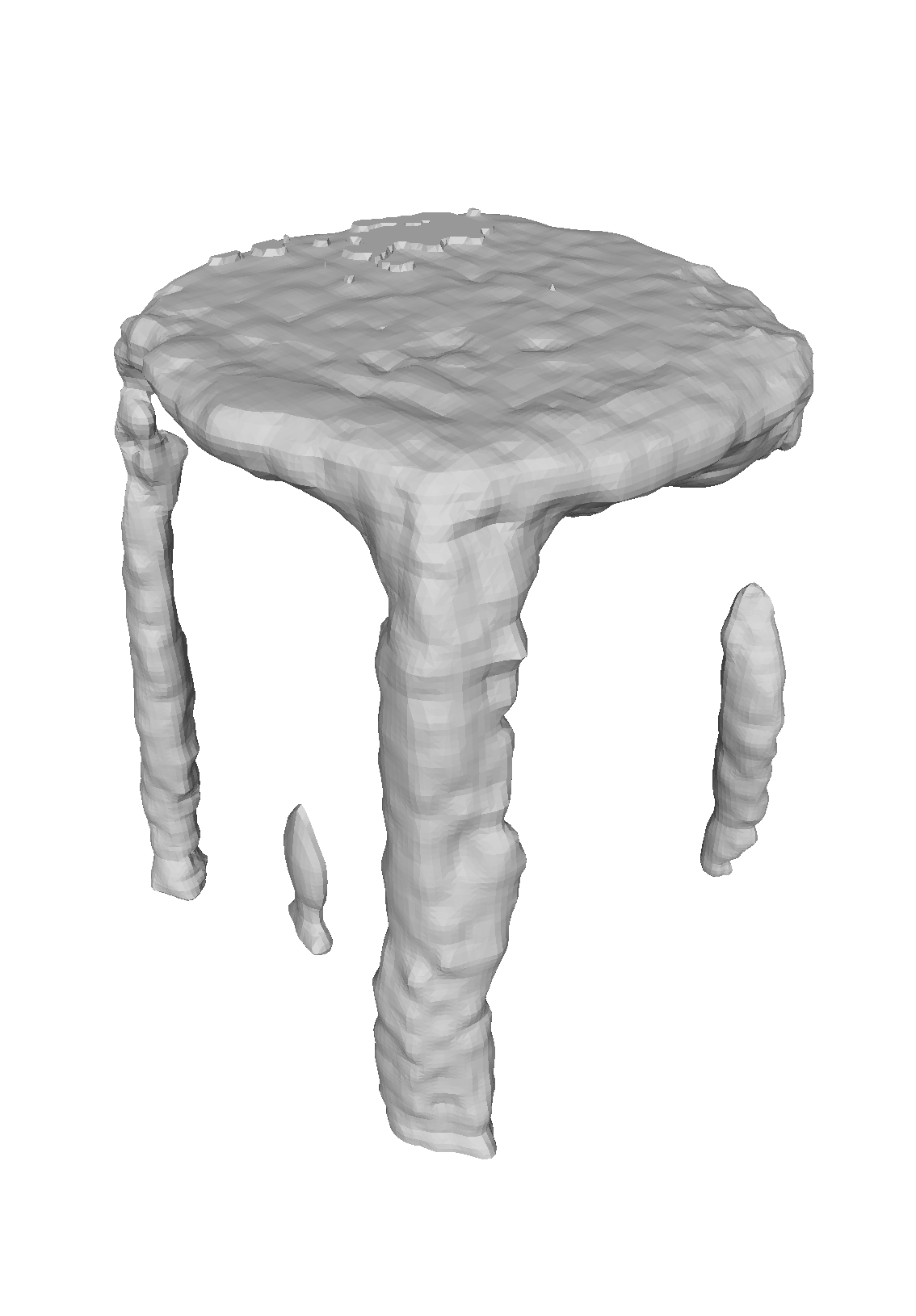} \\
		\inserte{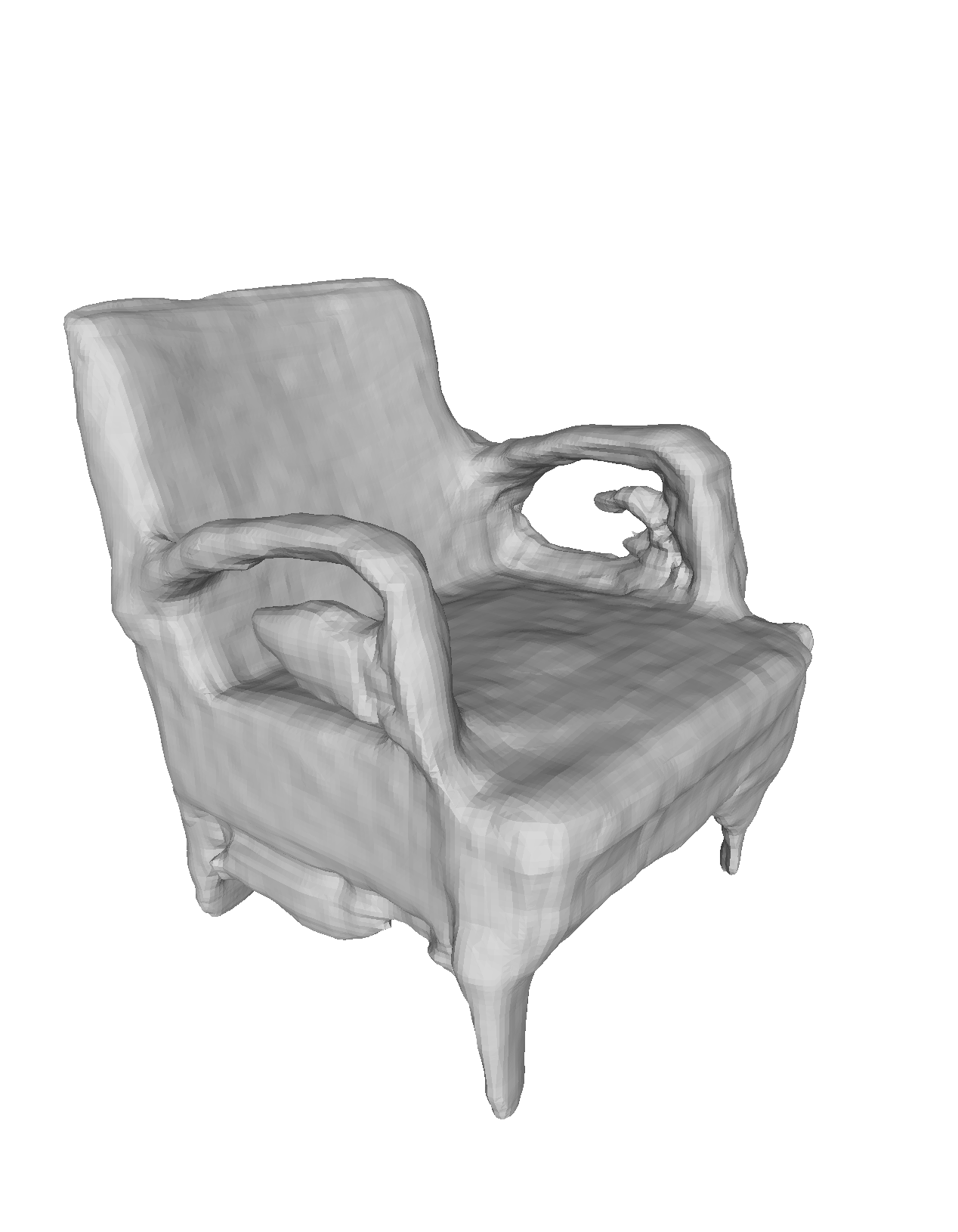} \\
		\insertf{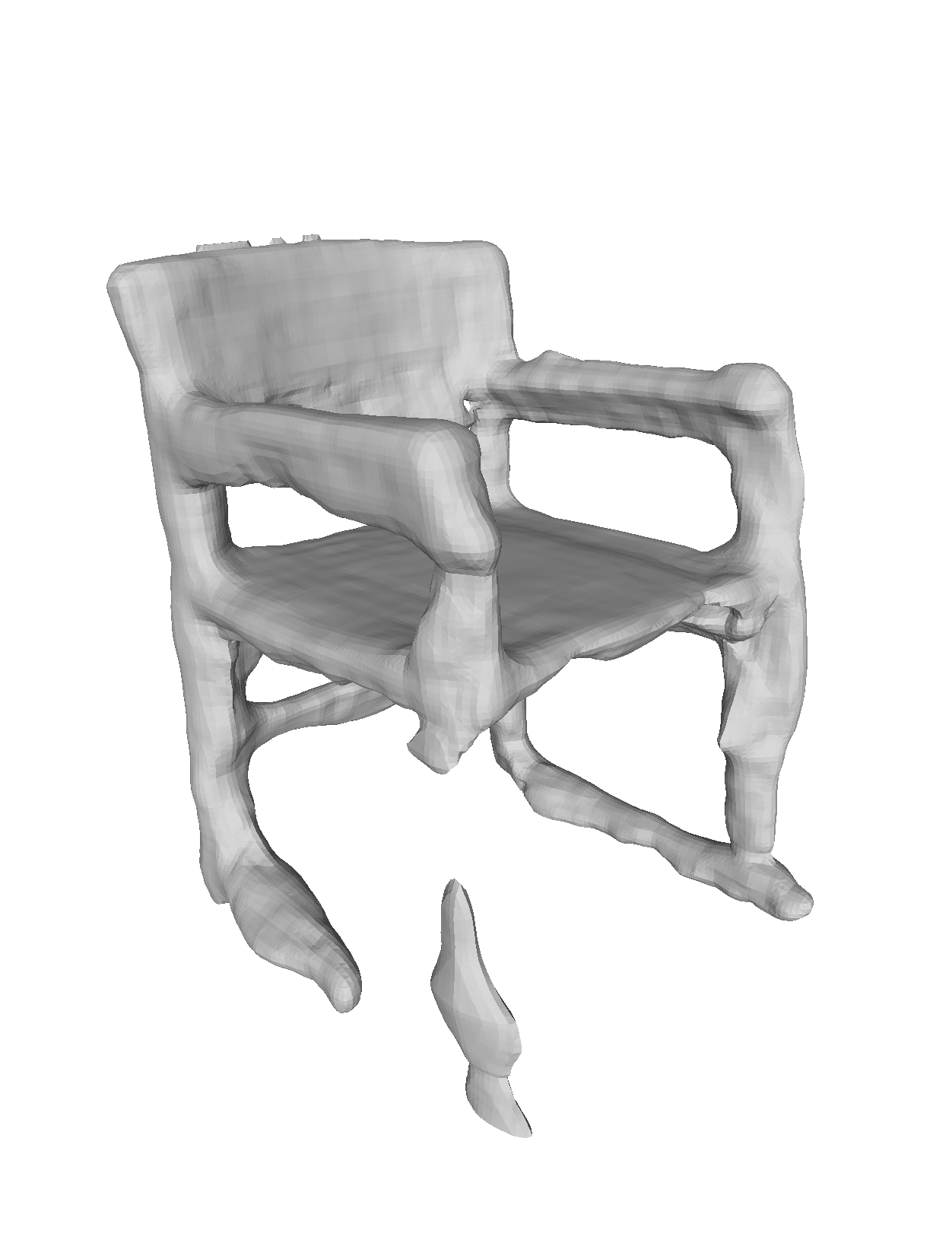} \\
		(8)
	\end{minipage}
	\begin{minipage}{0.105\linewidth}
		\centering
		\insertd{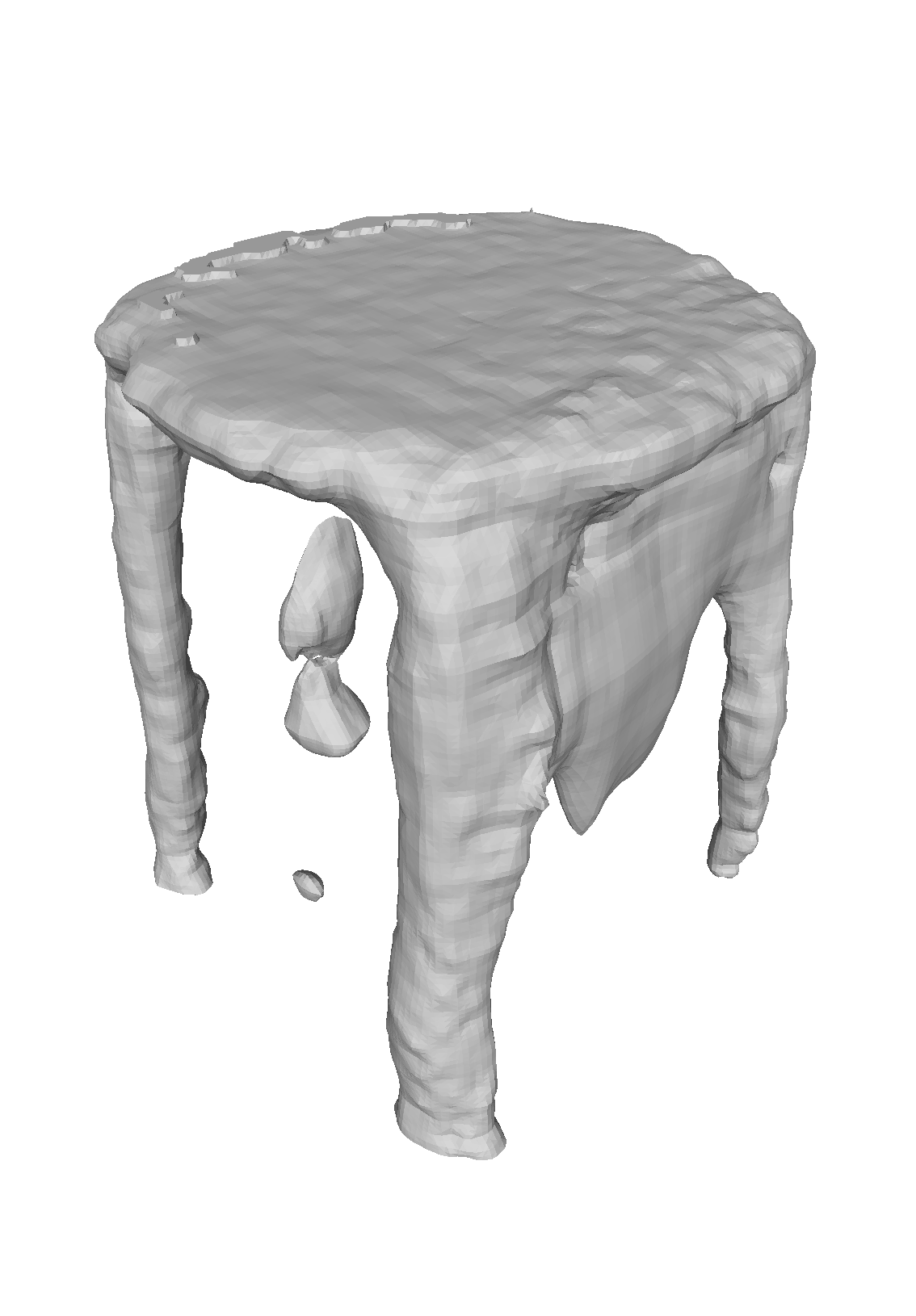} \\
		\inserte{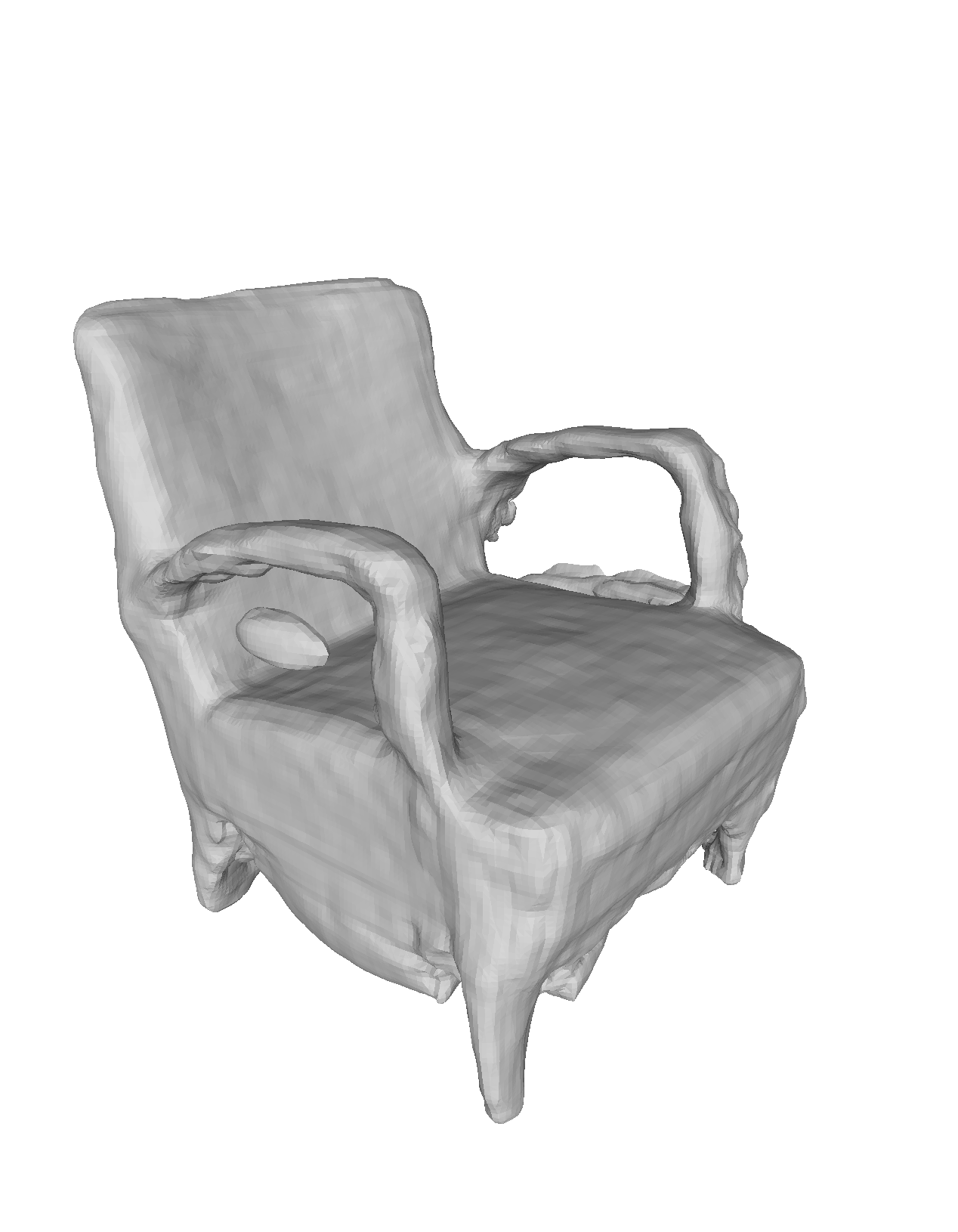} \\
		\insertf{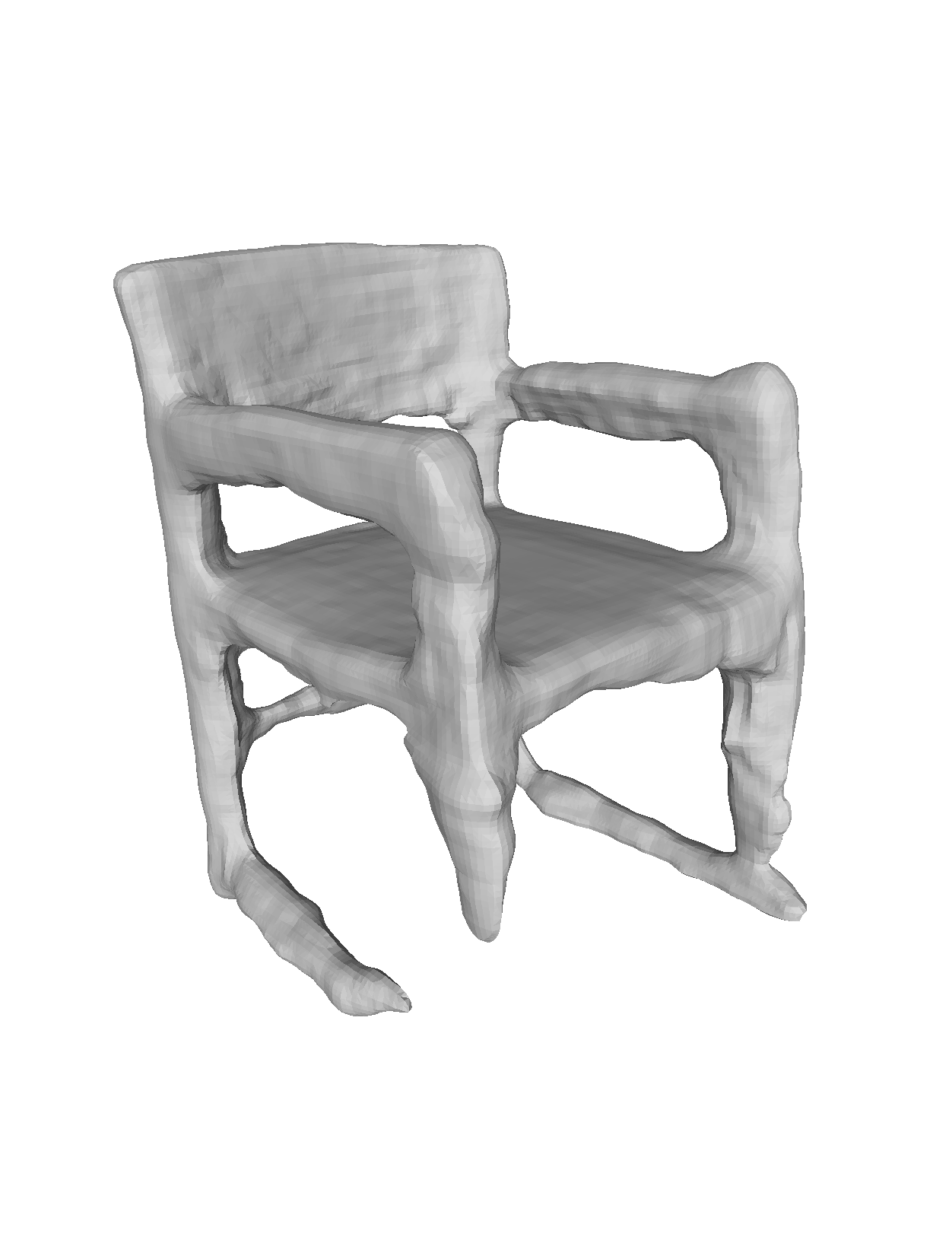} \\
		(9)
	\end{minipage}
	\caption{Visualization of the ablation experiments in Tab. 5 of the main paper. (1) Real Scan. (2) Ground Truth. \Frst{(3) Ours (Random \& Surface)}. (4) w/o consistency training. (5) Random Only. (6) Volume Only. (7) Surface Only. (8) Random \& Volume. (9) Random , Volume \& Surface.} 
	\label{fig:abltab5}
	\vspace{-2mm}
\end{figure*}

%% file: figures/Supp_failure_case.tex
\begin{figure}[htbp] \centering
    \includegraphics[width=0.48\textwidth]{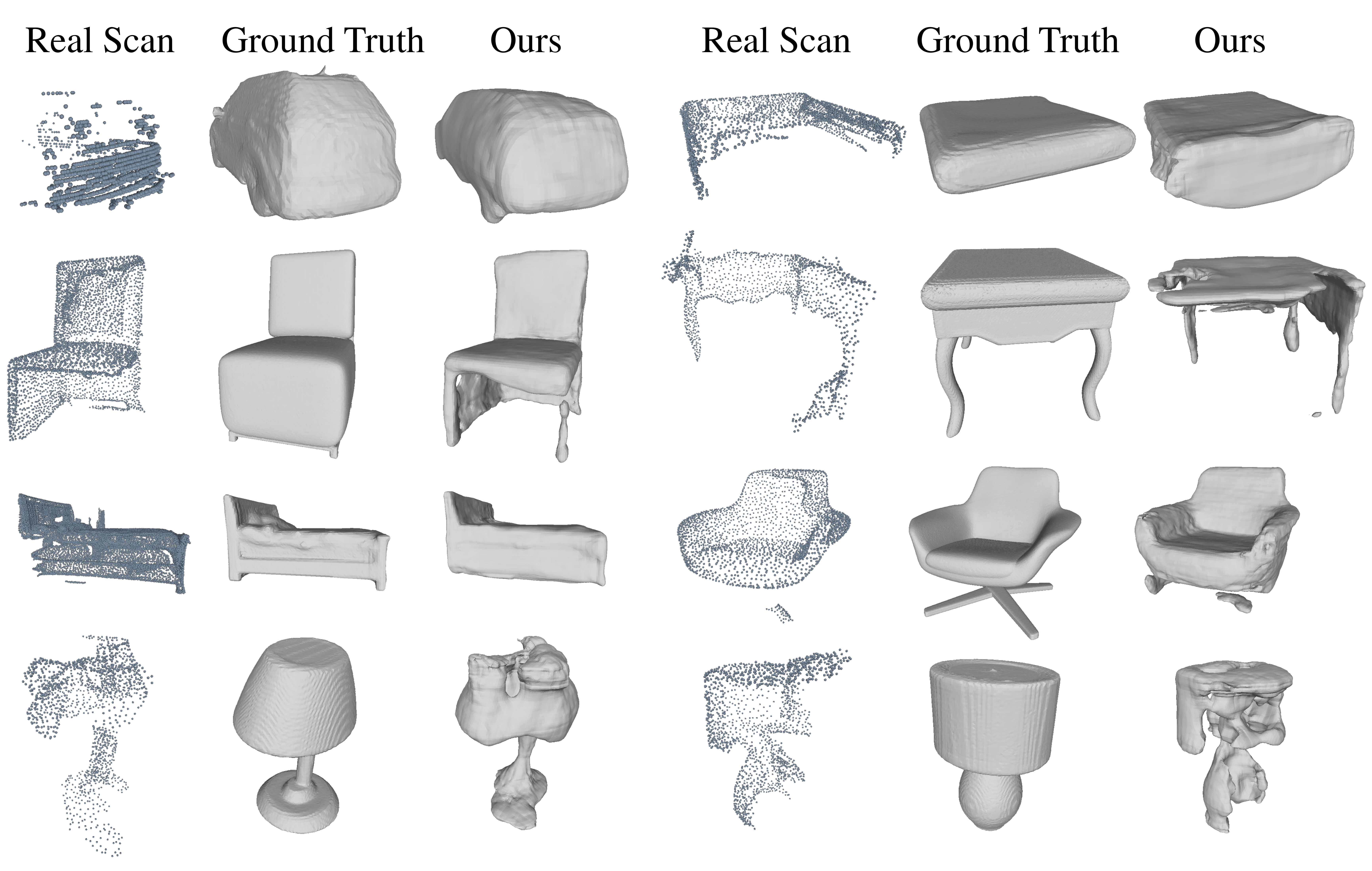}
    \caption{Three kinds of failure cases of reconstruction. The first two rows give 4 examples for case (i), and the third and the forth row gives 2 examples for case (ii) and (iii), respectively. } \label{fig:failure_case}
    % \vspace{-5mm}
\end{figure}

%% file: 0_main.bbl
\begin{thebibliography}{100}\itemsep=-1pt

\bibitem{achituve2021self}
Idan Achituve, Haggai Maron, and Gal Chechik.
\newblock Self-supervised learning for domain adaptation on point clouds.
\newblock In {\em WACV}, 2021.

\bibitem{afham2022crosspoint}
Mohamed Afham, Isuru Dissanayake, Dinithi Dissanayake, Amaya Dharmasiri,
  Kanchana Thilakarathna, and Ranga Rodrigo.
\newblock Crosspoint: Self-supervised cross-modal contrastive learning for 3d
  point cloud understanding.
\newblock In {\em CVPR}, 2022.

\bibitem{avetisyan2019scan2cad}
Armen Avetisyan, Manuel Dahnert, Angela Dai, Manolis Savva, Angel~X Chang, and
  Matthias Nie{\ss}ner.
\newblock Scan2cad: Learning cad model alignment in rgb-d scans.
\newblock In {\em CVPR}, 2019.

\bibitem{badki2020meshlet}
Abhishek Badki, Orazio Gallo, Jan Kautz, and Pradeep Sen.
\newblock Meshlet priors for 3d mesh reconstruction.
\newblock In {\em CVPR}, 2020.

\bibitem{berthelot2021adamatch}
David Berthelot, Rebecca Roelofs, Kihyuk Sohn, Nicholas Carlini, and Alex
  Kurakin.
\newblock Adamatch: A unified approach to semi-supervised learning and domain
  adaptation.
\newblock In {\em ICLR}, 2022.

\bibitem{bian2022unsupervised2}
Yikai Bian, Le Hui, Jianjun Qian, and Jin Xie.
\newblock Unsupervised domain adaptation for point cloud semantic segmentation
  via graph matching.
\newblock {\em arXiv preprint arXiv:2208.04510}, 2022.

\bibitem{bian2022unsupervised1}
Yikai Bian, Jin Xie, and Jianjun Qian.
\newblock Unsupervised domain adaptive point cloud semantic segmentation.
\newblock In {\em ACPR}, 2022.

\bibitem{cai2022learning}
Yingjie Cai, Kwan-Yee Lin, Chao Zhang, Qiang Wang, Xiaogang Wang, and Hongsheng
  Li.
\newblock Learning a structured latent space for unsupervised point cloud
  completion.
\newblock In {\em CVPR}, 2022.

\bibitem{cardace2021refrec}
Adriano Cardace, Riccardo Spezialetti, Pierluigi~Zama Ramirez, Samuele Salti,
  and Luigi Di~Stefano.
\newblock Refrec: Pseudo-labels refinement via shape reconstruction for
  unsupervised 3d domain adaptation.
\newblock In {\em 3DV}, 2021.

\bibitem{chang2015shapenet}
Angel~X Chang, Thomas Funkhouser, Leonidas Guibas, Pat Hanrahan, Qixing Huang,
  Zimo Li, Silvio Savarese, Manolis Savva, Shuran Song, Hao Su, et~al.
\newblock Shapenet: An information-rich 3d model repository.
\newblock {\em arXiv preprint arXiv:1512.03012}, 2015.

\bibitem{chen2020homm}
Chao Chen, Zhihang Fu, Zhihong Chen, Sheng Jin, Zhaowei Cheng, Xinyu Jin, and
  Xian-Sheng Hua.
\newblock Homm: Higher-order moment matching for unsupervised domain
  adaptation.
\newblock In {\em AAAI}, 2020.

\bibitem{chen2020harmonizing}
Chaoqi Chen, Zebiao Zheng, Xinghao Ding, Yue Huang, and Qi Dou.
\newblock Harmonizing transferability and discriminability for adapting object
  detectors.
\newblock In {\em CVPR}, 2020.

\bibitem{chen2022reusing}
Lin Chen, Huaian Chen, Zhixiang Wei, Xin Jin, Xiao Tan, Yi Jin, and Enhong
  Chen.
\newblock Reusing the task-specific classifier as a discriminator:
  Discriminator-free adversarial domain adaptation.
\newblock In {\em CVPR}, 2022.

\bibitem{chen2019unpaired}
Xuelin Chen, Baoquan Chen, and Niloy~J Mitra.
\newblock Unpaired point cloud completion on real scans using adversarial
  training.
\newblock In {\em ICLR}, 2019.

\bibitem{chen2019transferability}
Xinyang Chen, Sinan Wang, Mingsheng Long, and Jianmin Wang.
\newblock Transferability vs. discriminability: Batch spectral penalization for
  adversarial domain adaptation.
\newblock In {\em ICML}, 2019.

\bibitem{chen2019crdoco}
Yun-Chun Chen, Yen-Yu Lin, Ming-Hsuan Yang, and Jia-Bin Huang.
\newblock Crdoco: Pixel-level domain transfer with cross-domain consistency.
\newblock In {\em CVPR}, 2019.

\bibitem{chen2019learning}
Zhiqin Chen and Hao Zhang.
\newblock Learning implicit fields for generative shape modeling.
\newblock In {\em CVPR}, 2019.

\bibitem{chibane2020implicit}
Julian Chibane, Thiemo Alldieck, and Gerard Pons-Moll.
\newblock Implicit functions in feature space for 3d shape reconstruction and
  completion.
\newblock In {\em CVPR}, 2020.

\bibitem{chibane2020implicit2}
Julian Chibane and Gerard Pons-Moll.
\newblock Implicit feature networks for texture completion from partial 3d
  data.
\newblock In {\em ECCV}, 2020.

\bibitem{choy20163d}
Christopher~B Choy, Danfei Xu, JunYoung Gwak, Kevin Chen, and Silvio Savarese.
\newblock 3d-r2n2: A unified approach for single and multi-view 3d object
  reconstruction.
\newblock In {\em ECCV}, 2016.

\bibitem{cubes1987high}
Marching Cubes.
\newblock A high resolution 3d surface construction algorithm.
\newblock In {\em Proceedings of the ACM on Computer Graphics and Interactive
  Techniques}, 1987.

\bibitem{cui2020gvb}
Shuhao Cui, Shuhui Wang, Junbao Zhuo, Chi Su, Qingming Huang, and Qi Tian.
\newblock Gradually vanishing bridge for adversarial domain adaptation.
\newblock In {\em CVPR}, 2020.

\bibitem{dai2017scannet}
Angela Dai, Angel~X Chang, Manolis Savva, Maciej Halber, Thomas Funkhouser, and
  Matthias Nie{\ss}ner.
\newblock Scannet: Richly-annotated 3d reconstructions of indoor scenes.
\newblock In {\em CVPR}, 2017.

\bibitem{damodaran2018deepjdot}
Bharath~Bhushan Damodaran, Benjamin Kellenberger, R{\'e}mi Flamary, Devis Tuia,
  and Nicolas Courty.
\newblock Deepjdot: Deep joint distribution optimal transport for unsupervised
  domain adaptation.
\newblock In {\em ECCV}, 2018.

\bibitem{deprelle2019learning}
Theo Deprelle, Thibault Groueix, Matthew Fisher, Vladimir Kim, Bryan Russell,
  and Mathieu Aubry.
\newblock Learning elementary structures for 3d shape generation and matching.
\newblock {\em NeurIPS}, 2019.

\bibitem{erler2020points2surf}
Philipp Erler, Paul Guerrero, Stefan Ohrhallinger, Niloy~J Mitra, and Michael
  Wimmer.
\newblock Points2surf learning implicit surfaces from point clouds.
\newblock In {\em ECCV}, 2020.

\bibitem{gadelha2021deep}
Matheus Gadelha, Rui Wang, and Subhransu Maji.
\newblock Deep manifold prior.
\newblock In {\em ICCV}, 2021.

\bibitem{gao2020learning}
Jun Gao, Wenzheng Chen, Tommy Xiang, Alec Jacobson, Morgan McGuire, and Sanja
  Fidler.
\newblock Learning deformable tetrahedral meshes for 3d reconstruction.
\newblock {\em NeurIPS}, 2020.

\bibitem{gao2021gradient}
Zhiqiang Gao, Shufei Zhang, Kaizhu Huang, Qiufeng Wang, and Chaoliang Zhong.
\newblock Gradient distribution alignment certificates better adversarial
  domain adaptation.
\newblock In {\em ICCV}, 2021.

\bibitem{geiger2012we}
Andreas Geiger, Philip Lenz, and Raquel Urtasun.
\newblock Are we ready for autonomous driving? the kitti vision benchmark
  suite.
\newblock In {\em CVPR}, 2012.

\bibitem{ghosh2020data}
Atin Ghosh and Alexandre~H Thiery.
\newblock On data-augmentation and consistency-based semi-supervised learning.
\newblock In {\em ICLR}, 2020.

\bibitem{groueix2018papier}
Thibault Groueix, Matthew Fisher, Vladimir~G Kim, Bryan~C Russell, and Mathieu
  Aubry.
\newblock A papier-m{\^a}ch{\'e} approach to learning 3d surface generation.
\newblock In {\em CVPR}, 2018.

\bibitem{gu2020weakly}
Jiayuan Gu, Wei-Chiu Ma, Sivabalan Manivasagam, Wenyuan Zeng, Zihao Wang, Yuwen
  Xiong, Hao Su, and Raquel Urtasun.
\newblock Weakly-supervised 3d shape completion in the wild.
\newblock In {\em ECCV}, 2020.

\bibitem{hane2017hierarchical}
Christian H{\"a}ne, Shubham Tulsiani, and Jitendra Malik.
\newblock Hierarchical surface prediction for 3d object reconstruction.
\newblock In {\em 3DV}, 2017.

\bibitem{hanocka2020point2mesh}
Rana Hanocka, Gal Metzer, Raja Giryes, and Daniel Cohen-Or.
\newblock Point2mesh: A self-prior for deformable meshes.
\newblock {\em arXiv preprint arXiv:2005.11084}, 2020.

\bibitem{hoyer2022daformer}
Lukas Hoyer, Dengxin Dai, and Luc Van~Gool.
\newblock Daformer: Improving network architectures and training strategies for
  domain-adaptive semantic segmentation.
\newblock In {\em CVPR}, 2022.

\bibitem{huang2022generation}
Junxuan Huang, Junsong Yuan, and Chunming Qiao.
\newblock Generation for unsupervised domain adaptation: A gan-based approach
  for object classification with 3d point cloud data.
\newblock In {\em ICASSP}, 2022.

\bibitem{pfnet20}
Zitian Huang, Yikuan Yu, Jiawen Xu, Feng Ni, and Xinyi Le.
\newblock Pf-net: Point fractal network for 3d point cloud completion.
\newblock In {\em CVPR}, 2020.

\bibitem{jaritz2020xmuda}
Maximilian Jaritz, Tuan-Hung Vu, Raoul~de Charette, Emilie Wirbel, and Patrick
  P{\'e}rez.
\newblock xmuda: Cross-modal unsupervised domain adaptation for 3d semantic
  segmentation.
\newblock In {\em CVPR}, 2020.

\bibitem{jiang2020local}
Chiyu Jiang, Avneesh Sud, Ameesh Makadia, Jingwei Huang, Matthias Nie{\ss}ner,
  Thomas Funkhouser, et~al.
\newblock Local implicit grid representations for 3d scenes.
\newblock In {\em CVPR}, 2020.

\bibitem{jiang2021lidarnet}
Peng Jiang and Srikanth Saripalli.
\newblock Lidarnet: A boundary-aware domain adaptation model for point cloud
  semantic segmentation.
\newblock In {\em ICRA}, 2021.

\bibitem{khodabandeh2019robust}
Mehran Khodabandeh, Arash Vahdat, Mani Ranjbar, and William~G Macready.
\newblock A robust learning approach to domain adaptive object detection.
\newblock In {\em ICCV}, 2019.

\bibitem{leung2021domain}
Brandon Leung, Siddharth Singh, and Arik Horodniceanu.
\newblock Domain adaptation for real-world single view 3d reconstruction.
\newblock {\em arXiv preprint arXiv:2108.10972}, 2021.

\bibitem{li2021self}
Qing Li, Xiaojiang Peng, and Qi Hao.
\newblock Self-ensemling for 3d point cloud domain adaption.
\newblock {\em arXiv preprint arXiv:2112.05301}, 2021.

\bibitem{li2021sp}
Ruihui Li, Xianzhi Li, Ka-Hei Hui, and Chi-Wing Fu.
\newblock Sp-gan: Sphere-guided 3d shape generation and manipulation.
\newblock {\em TOG}, 2021.

\bibitem{lin2018learning}
Chen-Hsuan Lin, Chen Kong, and Simon Lucey.
\newblock Learning efficient point cloud generation for dense 3d object
  reconstruction.
\newblock In {\em AAAI}, 2018.

\bibitem{lin2020sdf}
Chen-Hsuan Lin, Chaoyang Wang, and Simon Lucey.
\newblock Sdf-srn: Learning signed distance 3d object reconstruction from
  static images.
\newblock {\em NeurIPS}, 2020.

\bibitem{litany2018deformable}
Or Litany, Alex Bronstein, Michael Bronstein, and Ameesh Makadia.
\newblock Deformable shape completion with graph convolutional autoencoders.
\newblock In {\em CVPR}, 2018.

\bibitem{morphing20}
Minghua Liu, Lu Sheng, Sheng Yang, Jing Shao, and Shi-Min Hu.
\newblock Morphing and sampling network for dense point cloud completion.
\newblock In {\em AAAI}, 2020.

\bibitem{long2017deep}
Mingsheng Long, Han Zhu, Jianmin Wang, and Michael~I Jordan.
\newblock Deep transfer learning with joint adaptation networks.
\newblock In {\em ICML}, 2017.

\bibitem{luo2021learnable}
Xiaoyuan Luo, Shaolei Liu, Kexue Fu, Manning Wang, and Zhijian Song.
\newblock A learnable self-supervised task for unsupervised domain adaptation
  on point clouds.
\newblock {\em arXiv preprint arXiv:2104.05164}, 2021.

\bibitem{mei2020instance}
Ke Mei, Chuang Zhu, Jiaqi Zou, and Shanghang Zhang.
\newblock Instance adaptive self-training for unsupervised domain adaptation.
\newblock In {\em ECCV}, 2020.

\bibitem{melas2021pixmatch}
Luke Melas-Kyriazi and Arjun~K Manrai.
\newblock Pixmatch: Unsupervised domain adaptation via pixelwise consistency
  training.
\newblock In {\em CVPR}, 2021.

\bibitem{mescheder2019occupancy}
Lars Mescheder, Michael Oechsle, Michael Niemeyer, Sebastian Nowozin, and
  Andreas Geiger.
\newblock Occupancy networks: Learning 3d reconstruction in function space.
\newblock In {\em CVPR}, 2019.

\bibitem{munir2021ssal}
Muhammad~Akhtar Munir, Muhammad~Haris Khan, M Sarfraz, and Mohsen Ali.
\newblock Ssal: Synergizing between self-training and adversarial learning for
  domain adaptive object detection.
\newblock {\em NeurIPS}, 2021.

\bibitem{ouasfi2022few}
Amine Ouasfi and Adnane Boukhayma.
\newblock Few'zero level set'-shot learning of shape signed distance functions
  in feature space.
\newblock {\em arXiv preprint arXiv:2207.04161}, 2022.

\bibitem{vrnet21}
Liang Pan, Xinyi Chen, Zhongang Cai, Junzhe Zhang, Haiyu Zhao, Shuai Yi, and
  Ziwei Liu.
\newblock Variational relational point completion network.
\newblock In {\em CVPR}, 2021.

\bibitem{park2019deepsdf}
Jeong~Joon Park, Peter Florence, Julian Straub, Richard Newcombe, and Steven
  Lovegrove.
\newblock Deepsdf: Learning continuous signed distance functions for shape
  representation.
\newblock In {\em CVPR}, 2019.

\bibitem{peng2021sparse}
Duo Peng, Yinjie Lei, Wen Li, Pingping Zhang, and Yulan Guo.
\newblock Sparse-to-dense feature matching: Intra and inter domain cross-modal
  learning in domain adaptation for 3d semantic segmentation.
\newblock In {\em CVPR}, 2021.

\bibitem{peng2021shape}
Songyou Peng, Chiyu Jiang, Yiyi Liao, Michael Niemeyer, Marc Pollefeys, and
  Andreas Geiger.
\newblock Shape as points: A differentiable poisson solver.
\newblock {\em NeurIPS}, 2021.

\bibitem{peng2020convolutional}
Songyou Peng, Michael Niemeyer, Lars Mescheder, Marc Pollefeys, and Andreas
  Geiger.
\newblock Convolutional occupancy networks.
\newblock In {\em ECCV}, 2020.

\bibitem{pinheiro2019domain}
Pedro~O Pinheiro, Negar Rostamzadeh, and Sungjin Ahn.
\newblock Domain-adaptive single-view 3d reconstruction.
\newblock In {\em ICCV}, 2019.

\bibitem{qin2019pointdan}
Can Qin, Haoxuan You, Lichen Wang, C-C~Jay Kuo, and Yun Fu.
\newblock Pointdan: A multi-scale 3d domain adaption network for point cloud
  representation.
\newblock {\em NeurIPS}, 2019.

\bibitem{sulzer2022Deep}
Sulzer Raphael, Landrieu Lo{\"{\i}}c, Boulch Alexandre, Marlet Renaud, and
  Vallet Bruno.
\newblock Deep surface reconstruction from point clouds with visibility
  information.
\newblock {\em arXiv preprint arXiv:2203.09440}, 2020.

\bibitem{rezaeianaran2021seeking}
Farzaneh Rezaeianaran, Rakshith Shetty, Rahaf Aljundi, Daniel~Olmeda Reino,
  Shanshan Zhang, and Bernt Schiele.
\newblock Seeking similarities over differences: Similarity-based domain
  alignment for adaptive object detection.
\newblock In {\em ICCV}, 2021.

\bibitem{rochan2022unsupervised}
Mrigank Rochan, Shubhra Aich, Eduardo~R Corral-Soto, Amir Nabatchian, and
  Bingbing Liu.
\newblock Unsupervised domain adaptation in lidar semantic segmentation with
  self-supervision and gated adapters.
\newblock In {\em ICRA}, 2022.

\bibitem{saito2018mcd}
Kuniaki Saito, Kohei Watanabe, Yoshitaka Ushiku, and Tatsuya Harada.
\newblock Maximum classifier discrepancy for unsupervised domain adaptation.
\newblock In {\em CVPR}, 2018.

\bibitem{saito2019pifu}
Shunsuke Saito, Zeng Huang, Ryota Natsume, Shigeo Morishima, Angjoo Kanazawa,
  and Hao Li.
\newblock Pifu: Pixel-aligned implicit function for high-resolution clothed
  human digitization.
\newblock In {\em ICCV}, 2019.

\bibitem{shen2022domain}
Yuefan Shen, Yanchao Yang, Mi Yan, He Wang, Youyi Zheng, and Leonidas~J Guibas.
\newblock Domain adaptation on point clouds via geometry-aware implicits.
\newblock In {\em CVPR}, 2022.

\bibitem{sitzmann2020metasdf}
Vincent Sitzmann, Eric Chan, Richard Tucker, Noah Snavely, and Gordon
  Wetzstein.
\newblock Metasdf: Meta-learning signed distance functions.
\newblock {\em NeurIPS}, 2020.

\bibitem{sohn2020fixmatch}
Kihyuk Sohn, David Berthelot, Nicholas Carlini, Zizhao Zhang, Han Zhang,
  Colin~A Raffel, Ekin~Dogus Cubuk, Alexey Kurakin, and Chun-Liang Li.
\newblock Fixmatch: Simplifying semi-supervised learning with consistency and
  confidence.
\newblock {\em NeurIPS}, 2020.

\bibitem{tatarchenko2019single}
Maxim Tatarchenko, Stephan~R Richter, Ren{\'e} Ranftl, Zhuwen Li, Vladlen
  Koltun, and Thomas Brox.
\newblock What do single-view 3d reconstruction networks learn?
\newblock In {\em CVPR}, 2019.

\bibitem{topnet19}
Lyne~P Tchapmi, Vineet Kosaraju, Hamid Rezatofighi, Ian Reid, and Silvio
  Savarese.
\newblock Topnet: Structural point cloud decoder.
\newblock In {\em CVPR}, 2019.

\bibitem{tzeng2017adversarial}
Eric Tzeng, Judy Hoffman, Kate Saenko, and Trevor Darrell.
\newblock Adversarial discriminative domain adaptation.
\newblock In {\em CVPR}, 2017.

\bibitem{uy2021joint}
Mikaela~Angelina Uy, Vladimir~G Kim, Minhyuk Sung, Noam Aigerman, Siddhartha
  Chaudhuri, and Leonidas~J Guibas.
\newblock Joint learning of 3d shape retrieval and deformation.
\newblock In {\em CVPR}, 2021.

\bibitem{cascaded20}
Xiaogang Wang, Marcelo~H Ang~Jr, and Gim~Hee Lee.
\newblock Cascaded refinement network for point cloud completion.
\newblock In {\em CVPR}, 2020.

\bibitem{softpoolnet20}
Yida Wang, David~Joseph Tan, Nassir Navab, and Federico Tombari.
\newblock Softpoolnet: Shape descriptor for point cloud completion and
  classification.
\newblock In {\em ECCV}, 2020.

\bibitem{wei2021deep}
Xingkui Wei, Zhengqing Chen, Yanwei Fu, Zhaopeng Cui, and Yinda Zhang.
\newblock Deep hybrid self-prior for full 3d mesh generation.
\newblock In {\em ICCV}, 2021.

\bibitem{wen2021cycle4completion}
Xin Wen, Zhizhong Han, Yan-Pei Cao, Pengfei Wan, Wen Zheng, and Yu-Shen Liu.
\newblock Cycle4completion: Unpaired point cloud completion using cycle
  transformation with missing region coding.
\newblock In {\em CVPR}, 2021.

\bibitem{skip-att20}
Xin Wen, Tianyang Li, Zhizhong Han, and Yu-Shen Liu.
\newblock Point cloud completion by skip-attention network with hierarchical
  folding.
\newblock In {\em CVPR}, 2020.

\bibitem{wen2021pmp}
Xin Wen, Peng Xiang, Zhizhong Han, Yan-Pei Cao, Pengfei Wan, Wen Zheng, and
  Yu-Shen Liu.
\newblock Pmp-net: Point cloud completion by learning multi-step point moving
  paths.
\newblock In {\em CVPR}, 2021.

\bibitem{williams2022neural}
Francis Williams, Zan Gojcic, Sameh Khamis, Denis Zorin, Joan Bruna, Sanja
  Fidler, and Or Litany.
\newblock Neural fields as learnable kernels for 3d reconstruction.
\newblock In {\em CVPR}, 2022.

\bibitem{williams2019deep}
Francis Williams, Teseo Schneider, Claudio Silva, Denis Zorin, Joan Bruna, and
  Daniele Panozzo.
\newblock Deep geometric prior for surface reconstruction.
\newblock In {\em CVPR}, 2019.

\bibitem{wu2019squeezesegv2}
Bichen Wu, Xuanyu Zhou, Sicheng Zhao, Xiangyu Yue, and Kurt Keutzer.
\newblock Squeezesegv2: Improved model structure and unsupervised domain
  adaptation for road-object segmentation from a lidar point cloud.
\newblock In {\em ICRA}, 2019.

\bibitem{wu2017marrnet}
Jiajun Wu, Yifan Wang, Tianfan Xue, Xingyuan Sun, Bill Freeman, and Josh
  Tenenbaum.
\newblock Marrnet: 3d shape reconstruction via 2.5 d sketches.
\newblock {\em NeurIPS}, 2017.

\bibitem{wu20213d}
Mengxi Wu, Hao Huang, and Yi Fang.
\newblock 3d point cloud completion with geometric-aware adversarial
  augmentation.
\newblock {\em arXiv preprint arXiv:2109.10161}, 2021.

\bibitem{wu2020multimodal}
Rundi Wu, Xuelin Chen, Yixin Zhuang, and Baoquan Chen.
\newblock Multimodal shape completion via conditional generative adversarial
  networks.
\newblock In {\em ECCV}, 2020.

\bibitem{wu20153d}
Zhirong Wu, Shuran Song, Aditya Khosla, Fisher Yu, Linguang Zhang, Xiaoou Tang,
  and Jianxiong Xiao.
\newblock 3d shapenets: A deep representation for volumetric shapes.
\newblock In {\em CVPR}, 2015.

\bibitem{snowflakenet21}
Peng Xiang, Xin Wen, Yu-Shen Liu, Yan-Pei Cao, Pengfei Wan, Wen Zheng, and
  Zhizhong Han.
\newblock Snowflakenet: Point cloud completion by snowflake point deconvolution
  with skip-transformer.
\newblock In {\em ICCV}, 2021.

\bibitem{xiao2022transfer}
Aoran Xiao, Jiaxing Huang, Dayan Guan, Fangneng Zhan, and Shijian Lu.
\newblock Transfer learning from synthetic to real lidar point cloud for
  semantic segmentation.
\newblock In {\em AAAI}, 2022.

\bibitem{xie2019pix2vox}
Haozhe Xie, Hongxun Yao, Xiaoshuai Sun, Shangchen Zhou, and Shengping Zhang.
\newblock Pix2vox: Context-aware 3d reconstruction from single and multi-view
  images.
\newblock In {\em ICCV}, 2019.

\bibitem{grnet20}
Haozhe Xie, Hongxun Yao, Shangchen Zhou, Jiageng Mao, Shengping Zhang, and
  Wenxiu Sun.
\newblock Grnet: Gridding residual network for dense point cloud completion.
\newblock In {\em ECCV}, 2020.

\bibitem{xie2020unsupervised}
Qizhe Xie, Zihang Dai, Eduard Hovy, Thang Luong, and Quoc Le.
\newblock Unsupervised data augmentation for consistency training.
\newblock {\em NeurIPS}, 2020.

\bibitem{xu2019disn}
Qiangeng Xu, Weiyue Wang, Duygu Ceylan, Radomir Mech, and Ulrich Neumann.
\newblock Disn: Deep implicit surface network for high-quality single-view 3d
  reconstruction.
\newblock {\em NeurIPS}, 2019.

\bibitem{fbnet22}
Xuejun Yan, Hongyu Yan, Jingjing Wang, Hang Du, Zhihong Wu, Di Xie, Shiliang
  Pu, and Li Lu.
\newblock Fbnet: Feedback network for point cloud completion.
\newblock In {\em ECCV}, 2022.

\bibitem{yi2021complete}
Li Yi, Boqing Gong, and Thomas Funkhouser.
\newblock Complete \& label: A domain adaptation approach to semantic
  segmentation of lidar point clouds.
\newblock In {\em CVPR}, 2021.

\bibitem{yin20213dstylenet}
Kangxue Yin, Jun Gao, Maria Shugrina, Sameh Khamis, and Sanja Fidler.
\newblock 3dstylenet: Creating 3d shapes with geometric and texture style
  variations.
\newblock In {\em ICCV}, 2021.

\bibitem{yu2021pointr}
Xumin Yu, Yongming Rao, Ziyi Wang, Zuyan Liu, Jiwen Lu, and Jie Zhou.
\newblock Pointr: Diverse point cloud completion with geometry-aware
  transformers.
\newblock In {\em ICCV}, 2021.

\bibitem{pcn18}
Wentao Yuan, Tejas Khot, David Held, Christoph Mertz, and Martial Hebert.
\newblock Pcn: Point completion network.
\newblock In {\em 3DV}, 2018.

\bibitem{zhang2021unsupervised}
Junzhe Zhang, Xinyi Chen, Zhongang Cai, Liang Pan, Haiyu Zhao, Shuai Yi,
  Chai~Kiat Yeo, Bo Dai, and Chen~Change Loy.
\newblock Unsupervised 3d shape completion through gan inversion.
\newblock In {\em CVPR}, 2021.

\bibitem{zhang2022point}
Wenxiao Zhang, Zhen Dong, Jun Liu, Qingan Yan, Chunxia Xiao, et~al.
\newblock Point cloud completion via skeleton-detail transformer.
\newblock {\em TVCG}, 2022.

\bibitem{zhao2021epointda}
Sicheng Zhao, Yezhen Wang, Bo Li, Bichen Wu, Yang Gao, Pengfei Xu, Trevor
  Darrell, and Kurt Keutzer.
\newblock epointda: An end-to-end simulation-to-real domain adaptation
  framework for lidar point cloud segmentation.
\newblock In {\em AAAI}, 2021.

\bibitem{zou2021geometry}
Longkun Zou, Hui Tang, Ke Chen, and Kui Jia.
\newblock Geometry-aware self-training for unsupervised domain adaptation on
  object point clouds.
\newblock In {\em ICCV}, 2021.

\end{thebibliography}
